\documentclass[Afour,sageh,times]{sagej}
\usepackage{moreverb,url}
\usepackage{xcolor}
\definecolor{MyLinkColor}{rgb}{0,0.07843,0.45098}
\usepackage[colorlinks,bookmarksopen,bookmarksnumbered,colorlinks=true, allcolors=MyLinkColor]{hyperref}

\newcommand\BibTeX{{\rmfamily B\kern-.05em \textsc{i\kern-.025em b}\kern-.08em
T\kern-.1667em\lower.7ex\hbox{E}\kern-.125emX}}

\usepackage{natbib}
\usepackage{amsmath}
\usepackage{xspace}
\usepackage{multirow}
\usepackage{pdfpages}
\usepackage{longtable} 
\usepackage{tabularx}

\usepackage{makecell}

\usepackage{listings}
\usepackage{xcolor}
\usepackage{dsfont}
\definecolor{backcolour}{rgb}{0.95,0.95,0.92}
\definecolor{codegreen}{rgb}{0,0.6,0}
\definecolor{codegray}{rgb}{0.5,0.5,0.5}
\definecolor{codepurple}{rgb}{0.58,0,0.82}

\lstdefinestyle{mystyle}{
    backgroundcolor=\color{backcolour},
    commentstyle=\color{codegreen},
    keywordstyle=\color{magenta},
    numberstyle=\tiny\color{codegray},
    stringstyle=\color{codepurple},
    basicstyle=\ttfamily\footnotesize,
    breakatwhitespace=false,         
    breaklines=true,                 
    captionpos=b,                    
    keepspaces=true,                 
    numbers=left,                    
    numbersep=5pt,                  
    showspaces=false,                
    showstringspaces=false,
    showtabs=false,                  
    tabsize=2
}

\usepackage{amsmath,amsfonts,bm}

\def\eqref#1{equation~\ref{#1}}

\def\1{\bm{1}}

\def\vg{{\bm{g}}}

\def\vp{{\bm{p}}}

\DeclareMathAlphabet{\mathsfit}{\encodingdefault}{\sfdefault}{m}{sl}
\SetMathAlphabet{\mathsfit}{bold}{\encodingdefault}{\sfdefault}{bx}{n}

\def\gD{{\mathcal{D}}}

\def\gG{{\mathcal{G}}}

\def\gO{{\mathcal{O}}}
\def\gP{{\mathcal{P}}}

\def\gR{{\mathcal{R}}}

\newcommand{\myparagraph}[1]{\vspace{5pt}\noindent\textbf{#1}}

\newcolumntype{P}[1]{>{\centering\arraybackslash}p{#1}}
\newcolumntype{M}[1]{>{\centering\arraybackslash}m{#1}}

\newcommand{\model}{SetItUp\xspace}
\newcommand{\taskname}{functional\xspace}
\newcommand{\tasknameabbr}{FORM\xspace}

\newcommand{\objectSet}{\ensuremath{\mathcal{O}}}
\newcommand{\noise}{\ensuremath{\epsilon}}

\newcommand{\loss}{\ensuremath{\mathcal{L}}}

\newcommand{\tfMatrix}{\mathbf{H}}
\newcommand{\goalPose}{\tfMatrix^{\mathrm{goal}}}
\newcommand{\initPose}{\tfMatrix^{\mathrm{init}}}
\newcommand{\graspPose}{\tfMatrix^{\mathrm{grasp}}}
\newcommand{\setTfMatrix}{\mathcal{H}}
\newcommand{\setTfMatrixPick}{\setTfMatrix^{\mathrm{pick}}}
\newcommand{\setTfMatrixPrePlace}{\setTfMatrix^{\mathrm{pre-place}}}
\newcommand{\setTfMatrixPostPlace}{\setTfMatrix^{\mathrm{post-place}}}

\makeatletter
\newcommand{\onedot}{\ifx\@let@token.\else.\null\fi}
\makeatother

\def\eg{\emph{e.g}\onedot} 
\def\ie{\emph{i.e}\onedot}

\makeatother

\def\volumeyear{2024}

\begin{document}

\runninghead{Xu et al.}

\title{``Set It Up": Functional Object Arrangement with Compositional Generative Models}

\author{Yiqing Xu\affilnum{1}, Jiayuan Mao\affilnum{2}, Linfeng Li\affilnum{1}, Yilun Du\affilnum{2}, Tomas Loz\'{a}no-P\'{e}rez\affilnum{2}, Leslie Pack Kaelbling \affilnum{2} and David Hsu \affilnum{1}}

\affiliation{\affilnum{1}School of Computing, National University of Singapore\\
\affilnum{2}CSAIL, Massachusetts Institute of Technology}
\corrauth{Yiqing XU, National University of Singapore}
\email{xuyiqing@comp.nus.edu.sg}

\begin{abstract}
\textit{Functional object arrangement} (\tasknameabbr) is the task of arranging objects to fulfill a function, \eg, ``set up a dining table for two''. One key challenge here is that the instructions for \tasknameabbr are often under-specified and do not explicitly specify the desired object goal poses. This paper presents \textit{\model}, a neuro-symbolic framework that learns to specify the goal poses of objects from a few training examples and a structured natural-language task specification. \model uses a \textit{grounding graph}, which is composed of abstract spatial relations among objects (\eg, {\it left-of}), as its intermediate representation. This decomposes the \tasknameabbr problem into two stages: (i)~predicting this graph among objects and (ii)~predicting object poses given the grounding graph. For (i), \model leverages large language models (LLMs) to induce Python programs from a task specification and a few training examples. This program can be executed to generate grounding graphs in novel scenarios. For (ii), \model pre-trains a collection of diffusion models to capture primitive spatial relations and online composes these models to predict object poses based on the grounding graph.
We evaluated \model on a dataset spanning three distinct task families: arranging tableware on a dining table, organizing items on a bookshelf, and laying out furniture in a bedroom. Experiments show that \model outperforms existing models in generating functional, physically feasible, and aesthetically pleasing object arrangements.
\end{abstract}


\maketitle

\footnotetext{This article extends our paper presented at the Robotics: Science and Systems (RSS) conference 2024, titled “Set It Up: Functional Object Arrangement with Compositional Generative Models” \citep{Xu-RSS-24}.}

\section{Introduction}

\begin{figure*}[tp]
    \centering
\includegraphics[width=\linewidth]{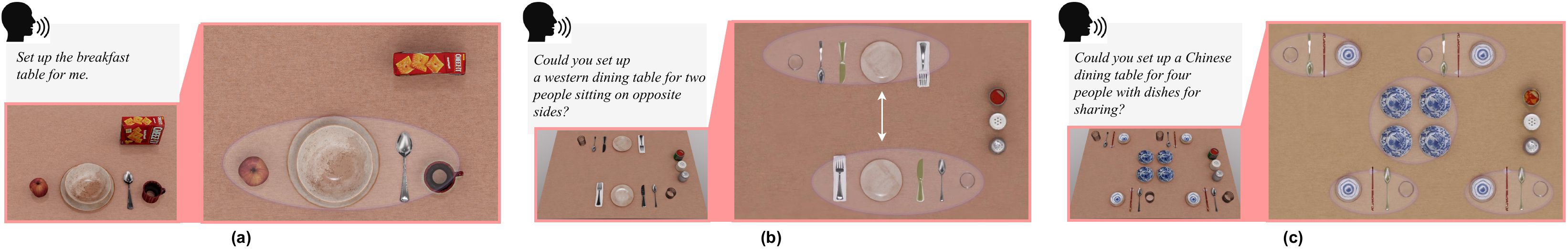}
    \vspace{-2em}
    \caption{Given highly varied, under-specified instructions-including different numbers of diners, dining styles, and specific requirements, as illustrated in (a-c)-we generate precise object poses to create diverse arrangements that fulfill the intended functionality, are physically feasible, and aesthetically appealing.}
    \label{fig:intro}
    \vspace{-1em}
\end{figure*}

How can robots learn to carry out user instructions such as ``Set up the table for a Chinese dinner of four people.''? Given this instruction, humans can usually place the tableware---bowls, plates, chopsticks---piece by piece in a physically feasible and aesthetically pleasing arrangement, sometimes with the help of a few demonstrations from an expert (Figure~\ref{fig:intro}). To achieve the same result with robots today, we very likely need to program the exact geometric pose of each piece of dinnerware. Further, we need to develop different programs, depending on the dinner style, the number of people, user preferences, etc. How can robot learning enable human-level performance and greatly reduce the burden on users? The key challenge here is for robots to understand high-level human instructions. In our example, the instruction specifies only the function of the tableware arrangement, while making no explicit mention of physical feasibility or aesthetics. 


Recently, there is a large body of work that attempts to ground natural language instructions for object arrangement at tabletop or even house scales. However, most of it focuses on grounding spatial relations and creating robot plans based on explicit and unambiguous instructions, such as ``put the plate next to the fork”~\citep{danielczuk2021object, driess2021learning, goodwin2022semantically, manuelli2019kpam, qureshi2021nerp, simeonov2021long, simeonov2023se, yuan2022sornet, zeng2021transporter}. In contrast, in this paper, we focus on goal specification for object arrangement tasks based on under-specified natural language descriptions, such as ``{\em set a Chinese dinner table for two},” ``{\em arrange objects on my bookshelf, ensuring the greenery receives sufficient lighting},” and ``{\em generate a bedroom furniture layout to maximize open space}.'' Given these ambiguous instructions, our goal is to generate configurations of objects that are functional, physically feasible, aesthetically appealing, and aligned with user preferences. These high-level object arrangements are intended as goal specifications for downstream execution, and we intentionally isolate this reasoning step from low-level motion planning to focus on the challenges of interpreting and grounding abstract, user-driven intent. We call this task {\it \taskname object arrangement} (\tasknameabbr), and it presents three challenges.


\begin{figure}[tp]
    \centering
\includegraphics[width=\linewidth]{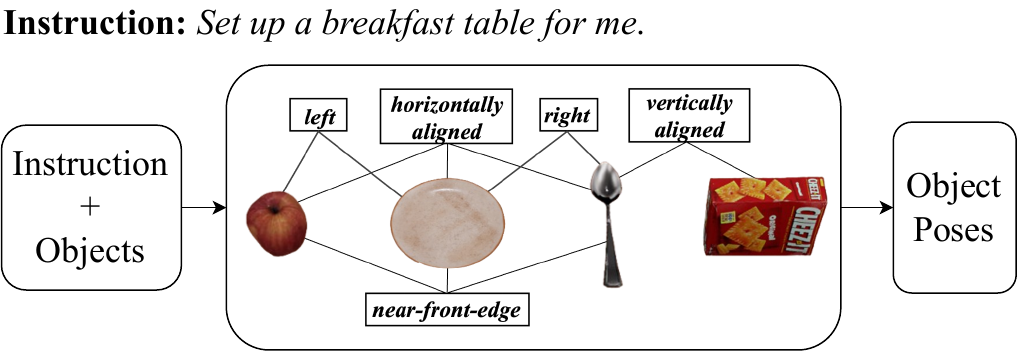}
    \vspace{-2em}
    \caption{Given a human instruction and a set of objects, our framework \model first generates a {\em grounding graph}---a multi-ary relational structure over object subsets based on a library of abstract spatial relations. This graph captures the desired layout semantics and is grounded into concrete object poses using compositional diffusion models.}
    \label{fig:teaser}
    \vspace{-1em}
\end{figure}
First, unlike low-level spatial relations---which are well-annotated~\citep{krishna2017visual} or easily synthesized from human-written rules---global scene annotations for {\it \taskname} arrangements under under-specified instructions are extremely limited. This scarcity is further compounded by the need to capture subtle layout shifts from slight changes in natural language. Second, beyond data scarcity, \taskname tasks require generalization over a vast combinatorial space---unstructured instructions, diverse object sets, and varying environments---many unseen during training. Finally, valid arrangements must simultaneously satisfy functional intent, physical feasibility, and aesthetic coherence. Balancing these often conflicting goals makes \taskname a complex optimization problem.

We formalize the \tasknameabbr problem in a learning-based setting structured around task families (e.g., dining tables, bookshelves). Given object categories and shapes, the goal is to generate object poses that are functional, physically feasible, and aesthetically coherent, conditioned on natural language instructions. To enable learning under limited supervision, we assume access to: (i) a structured natural language task specification per family, and (ii) five instruction-arrangement pairs. The task specification---shared across tasks---encodes domain-specific prior knowledge and outlines a general strategy for abstract arrangements using natural language, allowing non-programmers to express organizational principles without formal syntax~\citep{lieberman2005feasibility, mihalcea2006nlp}. This is a one-time effort per family, reusable across all instances. The training examples serve to validate generalization of the logic, rather than directly guide generation. Finally, we introduce a benchmark suite with rule-based metrics and human evaluation protocols to assess physical feasibility, functionality, and visual quality.

We introduce a neuro-symbolic generative framework (Figure~\ref{fig:teaser}) for {\it \taskname} object arrangements. The key idea is to use a symbolic intermediate---an abstract {\em grounding graph}---built from a predefined library of spatial relations (e.g., \emph{left-of}, \emph{horizontally-aligned}). Given an instruction like “set up a breakfast table,” \model first generates an abstract grounding graph capturing abstract layout semantics, then grounds it into real-valued object poses. This decouples the task into (i) \textit{semantic inference} from instruction and object types to grounding graph, and (ii) \textit{geometric grounding} from graph to object poses.

Each component is further decomposed to enhance data efficiency and generalization. For \emph{semantic inference}, we adopt a hierarchical, verifier-in-the-loop approach: the structured task specification is translated into a Python-style program sketch with function stubs; an LLM-based coder fills in implementations, and a verifier checks correctness against five demonstrations. This yields modular, reliable programs from natural language. For \emph{geometric grounding}, we pre-train a library of diffusion models---one per primitive spatial relation---using synthetic data. At inference, these are composed according to the predicted graph. For example, to place a spoon and fork relative to a plate, models for \emph{left-of} and \emph{right-of} are composed. This strategy avoids retraining and scales across diverse scenes.

Our decomposition leads to strong gains across all metrics, especially in generalizing to unseen object sets and instructions. On dining table, bookshelf, and bedroom benchmarks, \model outperforms diffusion-only methods~\citep{liu2022structdiffusion}, LLM-based approaches~\citep{wu2023tidybot}, and our own ablations. Both qualitative examples and human studies confirm that \model produces functionally coherent, physically feasible, and aesthetically preferred arrangements.

This paper extends our prior work presented at the Robotics: Science and Systems (RSS) conference, titled “Set It Up: Functional Object Arrangement with Compositional Generative Models”\citep{Xu-RSS-24}, in two key ways: (i) we introduce a new compositional program induction pipeline that takes structured natural language task specifications as input, improving usability and reliability (Section\ref{ssec:llm}); and (ii) we expand experimental evaluations beyond tabletop settings to new task families, including personalized bookshelf arrangement (Section~\ref{sssec:bookshelf}) and compact bedroom furniture layout (Section~\ref{sssec:bedroom}), demonstrating broader generality and robustness. Appendix~\ref{append:open_discussion} provides a detailed comparison with the conference version.
\section{Related Work}

\subsection{Object Rearrangement}
Robotic object rearrangement has been widely studied, often assuming access to a fully specified goal configuration with exact object poses~\citep{danielczuk2021object, driess2021learning, goodwin2022semantically, manuelli2019kpam, qureshi2021nerp, simeonov2021long, simeonov2023se, yuan2022sornet, zeng2021transporter}. Many works focus on task and motion planning (TAMP) given such well-defined targets or goal images. Others interpret language instructions to infer object placements based on explicit spatial relations~\citep{gkanatsios2023energy}. ALFRED~\citep{blukis2022persistent, shridhar2020alfred} introduced multi-step language-guided rearrangement, inspiring models that integrate high-level skills---but these typically reduce to placing objects into the correct receptacles, without reasoning about inter-object spatial or functional relations needed for structured tabletop setups. In contrast, our method does not rely on detailed goal states or concrete instructions. We handle under-specified human input by inferring abstract spatial relations among objects and computing precise arrangements that fulfill them.

\subsection{Functional Object Arrangement}

Arranging objects into functionally coherent tabletop layouts goes beyond geometric placement---it requires that items not only appear neat but also fulfill intended purposes, often by grouping functionally related objects.

Existing benchmarks~\citep{szot2021habitat, weihs2021visual, kant2022housekeep} address basic room tidying, focusing on placing items in designated containers to meet semantic or aesthetic goals~\citep{wu2023tidybot, sarch2022tidee}, but overlook spatial relations critical for functional and visual coherence on tabletops. Prior data-driven methods like Structformer~\citep{liu2022structformer} and StructDiffusion~\citep{liu2022structdiffusion} predict arrangements from vague instructions using a transformer or diffusion model. Others estimate tidiness scores~\citep{kapelyukh2022my, wei2023lego} or gradients toward neat configurations~\citep{wu2022targf, kapelyukh2022my}, but rely heavily on large datasets and generalize poorly to new object combinations or user-specific intent. DALL-E-BOT~\citep{kapelyukh2023dall} improves generalization using a pre-trained VLM to generate commonsense layouts for open-category objects. However, its single-model, zero-shot approach cannot account for user preferences or functional goals, and it ignores object geometries---key to physical feasibility.

In contrast, our method enables zero-shot generalization to novel objects and contexts by combining symbolic reasoning via an LLM with compositional diffusion-based grounding, eliminating the need for expert demonstrations.

\subsection{Knowledge Extraction from LLM}

Recent advances in large language models (LLMs) have enabled their use in robot decision-making, particularly for task planning in household settings~\citep{liu2023pre, ahn2022can, huang2022language, wu2023tidybot, zeng2023learning, jiang2024llms}. Huang et al.\citep{huang2022language} showed that iterative prompt augmentation improves task plan generation, while SayCan\citep{ahn2022can} integrates affordance functions to assess action feasibility in tasks like “make breakfast.” LLM-GROP~\citep{ding2023task} uses LLMs to suggest object placements via prompting. However, LLMs struggle with spatial reasoning in complex environments, especially as object count increases. Some methods attempt to bridge this gap by combining LLM-suggested layouts with vision models trained on neat configurations~\citep{xu2023tidy}, but performance drops with novel arrangements due to reliance on a single visual model.
In contrast, our method uses compositional diffusion models to ground symbolic object relations, enabling more accurate and physically feasible placements. Their inherent compositionality allows for principled relational optimization, improving generalization to complex geometric layouts.
\section{Overview}

\subsection{Problem Formulation}

We study the problem of learning to perform functional object arrangement (\tasknameabbr) from high-level specifications and a small number of examples. At training time, each object arrangement {\em task family} is denoted as a tuple $\langle \emph{spec}, \gD \rangle$, where $\emph{spec}$ is a structured natural language specification describing the general strategy for arranging objects, and $\gD$ is a small dataset of examples tied to a specific scene type and user preference. The goal is to learn from $\langle \emph{spec}, \gD \rangle$ and generalize to novel \tasknameabbr tasks within the same family.

\begin{figure*}[tp]
    \centering
    \includegraphics[width=\linewidth, page=1]{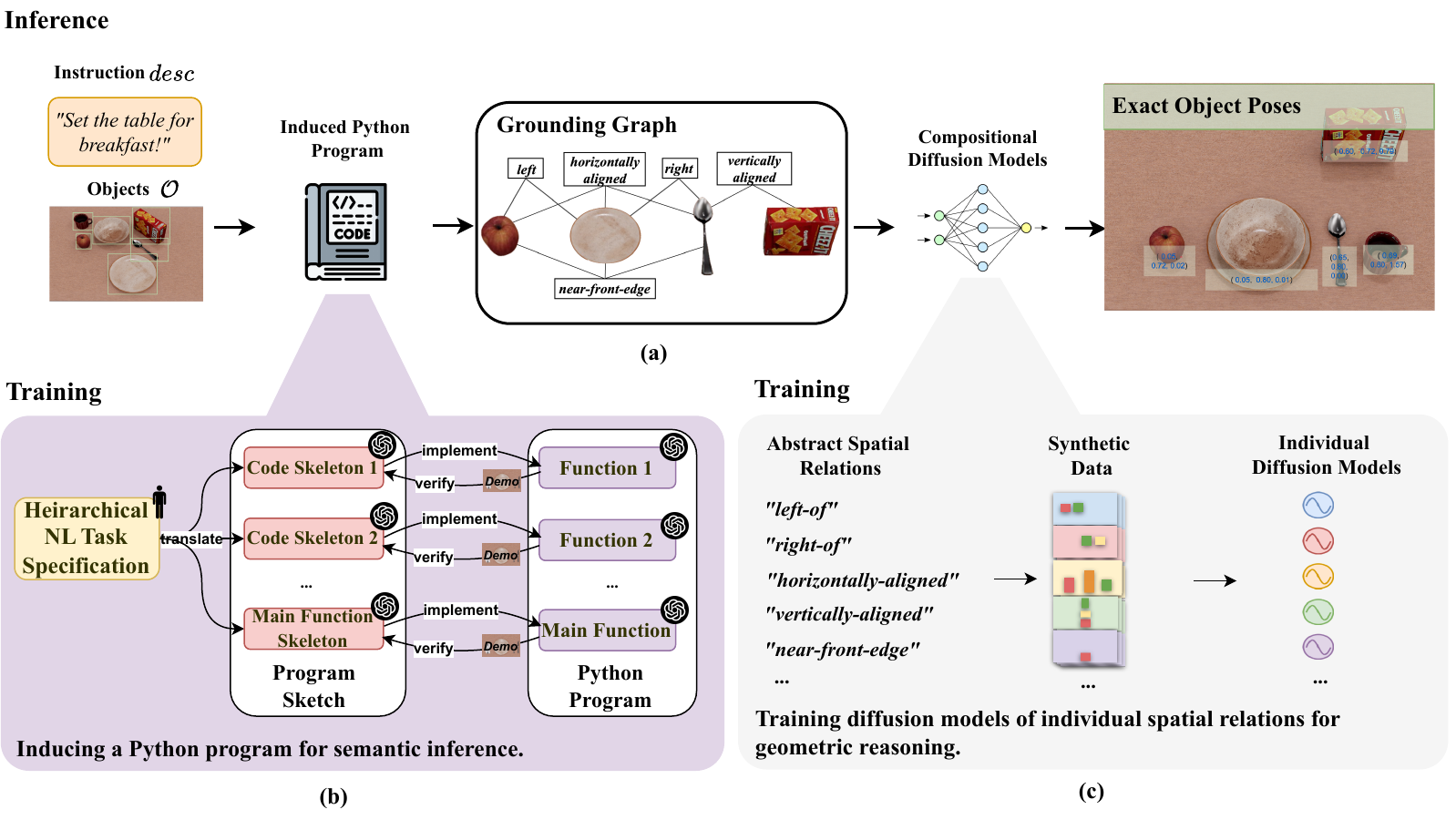}
    \caption{Overall architecture of \model: An intermediate abstract grounding graph decomposes tasks into semantic inference-interpreting instructions as symbolic object relations-and geometric reasoning-grounding this graph into exact object poses. At inference time (a), given an instruction $\emph{desc}$ and objects $\objectSet$, we first generate an abstract spatial relation description using a Python program, then we ground these relations into object poses by composing individual diffusion models selected from the library to propose the poses satisfying all specified relations. At training time, as illustrated in (b) and (c), we induce arrangement Python programs from the natural language specifications via LLMs and pretrain diffusion models, each predicting poses for a single abstract relation, to prepare for the abstract relation library.}
    \label{fig:model-overview}
\end{figure*}

The specification \emph{spec} organizes the natural language description into two components: a {\em domain specification} that defines reusable arrangement patterns (e.g., “western-style setup for a single diner”), and a {\em procedural description} that outlines how to interpret instructions, decompose them into sub-components, and generate arrangements---detailed further in Section~\ref{sec:hierarchical_task_specification}. Together, they define a hierarchical generative model for symbolic object arrangements by decomposing the natural language description into a hierarchy of simpler sub-tasks with well-defined input-output specifications. 

Each training example $d \in \gD$ is a tuple $\langle \emph{desc}_d, \objectSet_d, \gP_d \rangle$, where $\emph{desc}_d$ is a natural language instruction (e.g., “set up a dining table for two”), $\objectSet_d = \{o_0, \ldots, o{N-1}\}$ is a set of input objects with static properties: types ($n_i$) and shapes ($g_i$, represented as 2D bounding boxes), and $\gP_d = \{p_0, \ldots, p_{N-1}\}$ is the desired arrangement---planar poses $(x, y, \theta)$ for each object in a canonical table frame.\footnote{This can be extended to 3D objects and poses.} Only a few examples (e.g., five) are needed, keeping data collection lightweight.

At test time, the system is given a new instance $\langle \emph{desc}, \objectSet \rangle$ from the same task family, with a novel instruction and object set $\objectSet = \{o_0, \ldots, o_{N-1}\}$ (potentially differing in category and shape). It must output a set of poses $\gP$ such that the resulting arrangement is physically feasible, functional, aesthetic, and aligned with patterns from $\gD$ as well as user intent in $\emph{desc}$.

\subsection{SetItUp}

To enable data-efficient learning and robust generalization in \emph{\taskname} object arrangement (\tasknameabbr), we introduce a neuro-symbolic generative framework called \emph{\model}. The core idea is to use a library of abstract spatial relations (e.g., \emph{left-of}, \emph{horizontally-aligned}) to build a grounding graph---an intermediate representation that guides the arrangement process. These relations are geometrically well-defined, allowing us to decompose the task into two sub-problems: (i) generating a symbolic grounding graph that encodes the arrangement plan, and (ii) grounding this graph into concrete object poses.

Figure~\ref{fig:model-overview} illustrates our overall framework. Central to our approach is the use of abstract spatial relations as an intermediate representation, which decomposes \tasknameabbr tasks into two generative sub-tasks. The first sub-task focuses on semantic inference: we induce a hierarchical symbolic generative model in the form of Python programs that yield functional abstract relations for the test instruction and objects. The second sub-task addresses geometric reasoning: we employ a compositional generative model that takes these abstract relations as input and proposes the corresponding object poses. By integrating the symbolic generative Python program for semantic inference with compositional diffusion models for geometric grounding through intermediate grounding graphs, we ensure that the final arrangements are functional, physically feasible, and aesthetically appealing.

 During training, \textit{semantic inference} is performed by inducing a Python program from the structured task specification $\emph{spec}$ and demonstrations $\mathcal{D}$, using a modular LLM-based program induction pipeline. An LLM first acts as a syntactic translator, converting $\emph{spec}$ into a Python program sketch with function names, signatures, and descriptions. A second LLM serves as a code generator, filling in each function’s implementation. A third LLM acts as a verifier, checking outputs against $\mathcal{D}$ and revising the sketch if inconsistencies arise. For \textit{geometric reasoning}, we pre-train a library of diffusion models on synthetic data to predict object poses given abstract relations and object geometries.

At inference time, given a new instruction $\emph{desc}$ and object set $\mathcal{O}$, the induced Python program generates a symbolic grounding graph---essentially a factor-graph-like representation of the desired arrangement. The corresponding diffusion models are then composed to predict object poses that satisfy the structure specified by this graph.

The following sections elaborate on each component: Section~\ref{ssec:dsl} introduces the abstract spatial relation library, each paired with a geometric-rule classifier and a generative model; Section~\ref{ssec:diffusion} presents the compositional inference algorithm that grounds these graphs into object poses; and Section~\ref{ssec:llm} details our modular program induction process, where LLMs operate as translator, generator, and verifier to induce symbolic relation programs from $\emph{spec}$ and a few examples.
\section{Pose Generation via Compositional Diffusion Models}
\label{sec:ccsp}

The grounding graph $\gG$ consists of symbolic relations $\{R_1, \ldots, R_N\}$, each associated with a subset of objects. To ground this graph to exact object poses $\gP$ that satisfy all the relations, we compose individual diffusion models $\{f_{R_1}, \ldots, f_{R_N}\}$. These models are indexed by a single spatial relation type and are trained to generate object poses satisfying the relation given object shape features. We use diffusion models to generate the object poses for each spatial relation because of their compositionality. Individually trained score functions can be combined at inference time to generate object poses $\gP$ that satisfy a novel combination of relations, specified by $\gG$.

In Section~\ref{ssec:dsl}, we describe our spatial relation library $\gR$, where each spatial relation $R$ corresponds to a diffusion-based generative model $f_{R}$. Grounding graphs $\gG = \{R_1, \ldots, R_M\}$ can be constructed using any combination of the symbolic relations from this library. To generate the object poses $\gP$, we compose the corresponding diffusion models $\{f_{R_1}, \ldots, f_{R_M}\}$ based on the grounding graph and perform a variant of the Unadjusted Langevin Algorithm (ULA) sampler (Section~\ref{ssec:diffusion}).

\subsection{A Library of Spatial Relations}
\label{ssec:dsl}
  \begin{table}[tp]
   \centering
   \caption{The set of abstract spatial relations among objects. These relations are formally defined by rules based on 2D object shapes and poses in the canonical frame.}
   \label{tab:relationships}
   \footnotesize
   \setlength{\tabcolsep}{2pt} 
   \begin{tabularx}{\columnwidth}{@{}X X@{}}
   \toprule
   \multicolumn{2}{l}{\textbf{Unary Relations}}  \\ \midrule
   central-column & central-row \\
   at-center & left-half \\
   right-half & front-half \\
   back-half & near-left-edge \\
   near-right-edge & near-front-edge \\
   near-back-edge & facing-front \\
   facing-back & against-right-wall \\
   against-left-wall & against-front-wall \\
   against-back-wall & right-of-wall \\
   left-of-wall & center-of-wall \\ 
   at-front-left-corner & at-front-right-corner \\
   at-back-left-corner & at-back-right-corner \\ \midrule
   \multicolumn{2}{l}{\textbf{Binary Relations}} \\ \midrule
   horizontally-aligned-bottom & horizontally-aligned-centroid \\
   vertically-aligned-centroid & horizontal-symmetry-on-table \\ 
   vertical-symmetry-on-table & left-of \\
   right-of & on-top-of \\
   centered & left-touching \\
   right-touching & in-front-of \\
   under-window & \\ \midrule
   \multicolumn{2}{l}{\textbf{Ternary Relations}} \\ \midrule
   horizontal-symmetry-about-axis-obj & vertical-symmetry-about-axis-obj \\ \midrule
   \multicolumn{2}{l}{\textbf{Variable-Arity Relations}} \\ \midrule
   aligned-in-horizontal-line-bottom & aligned-in-horizontal-line-centroid \\ 
   aligned-in-vertical-line-centroid & regular-grid \\
    contiguously-aligned & height-sorted-ascending \\  height-sorted-descending &
   width-sorted-ascending \\ width-sorted-descending \\ \bottomrule
   \end{tabularx}
   \end{table}

Our spatial relation library, $\gR$, encompasses 48 basic relations listed in Table \ref{tab:relationships}. Each abstract relation takes a (possibly variable-sized) set of objects as arguments --- and describes a desired spatial relation among them. These abstract relations provide one level of abstraction over 2D poses and, therefore, can serve as a natural input-output format for LLMs. On the other hand, they are sufficiently detailed to be directly interpreted as geometric concepts.  To extend the system to task families requiring novel relations, it would be straightforward to augment this set.
Furthermore, these spatial constraints are defined unambiguously based on simple geometric transformations, which enables us to effectively classify and generate random object arrangements that satisfy a particular relation. Finally and most importantly, this finite set of relations can be {\it composed} to describe an expansive set of possible scene-level arrangements with an indefinite number of objects and relations.

Formally, each abstract spatial relation $R$ is associated with two models: a classifier model $h_R$ and a generative model $f_R$. Let $k_R$ be the arity of the relation. The classifier $h_R$ is a function, denoted as $h_R\left( g_1, \ldots, g_{k_R}, p_1, \ldots, p_{k_R} \right)$, that takes the static properties $\{ g_i \}$ of $k_R$ objects (shapes represented by 2D bounding-boxes in our case) and their poses $\{ p_i \}$ as input, and outputs a Boolean value indicating whether the input objects satisfy the relation $R$. Similarly, the generative model $f_R$ is a function that takes the static properties of $k_R$ objects $\{ g_i \}$ and produces samples of $\{ p_i \}$ from the distribution\footnote{In practice, we always constrain the value range of $p_i$ to be within $[0, 1]^*$; therefore, this distribution is properly defined.} $q_R\left( p_1, \ldots, p_{k_R} \mid g_1, \ldots, g_{k_R} \right) \propto \mathds{1}\left[ h_R\left( g_1, \ldots, g_{k_R}, p_1, \ldots, p_{k_R} \right) \right]$, where $\mathds{1}[\cdot]$ is the indicator function. In general, $R$ (and therefore the associated $h_R$ and $f_R$) can be a set-based function and, therefore, may have a variable arity, in which case $k_R = \star$.

\myparagraph{Grounding graph.} Given the library of abstract spatial relations $\gR$, we can encode a desired spatial arrangement of an object set $\gO$ as a graph of ground spatial relations, $\gG = \{r_i(o^i_1, \ldots, o^i_{k_i})\}_i$ where each $r_i \in \mathcal{R}$ is a relation (such as {\it horizontally-aligned}), $k_i$ is the arity of $r_i$, and $o^i_1, \ldots, o^i_{k_i}$ are elements of $\mathcal{O}$. The objective is to produce a set of poses $\gP = \{ p_i \}$, such that, for all elements $(R, (o_1, \ldots, o_{k_R})) \in \mathcal{G}$, $h_R(g_1, \ldots g_{k_R}, p_1, \ldots, p_{k_R})$ is true, where $g_i$ is the corresponding shape of $o_i$.

We can interpret $(\gP, \gG)$ as a graph of constraints. Based on the probabilistic distribution specified with all relations in $\gR$ (\ie, uniform distributions over allowable assignments), $(\gP, \gG)$ can also be interpreted as a factor graph, specifying a joint distribution over values of $\gP$. This will be our inference-time objective.

\myparagraph{Classifying spatial relations} Since all relations used in our examples are unambiguously defined based on simple geometric transformations (\eg, by comparing the 2D coordinate of objects in a canonical table frame), we use a small set of rules to construct the classifier function $h_R$. We include the details of the rules in the appendix \ref{apped:abstract_rule}.  They could instead be learned in conjunction with the generative models for relations that are defined through examples only. 

\begin{figure}[tp]
    \centering
    \includegraphics[width=\linewidth]{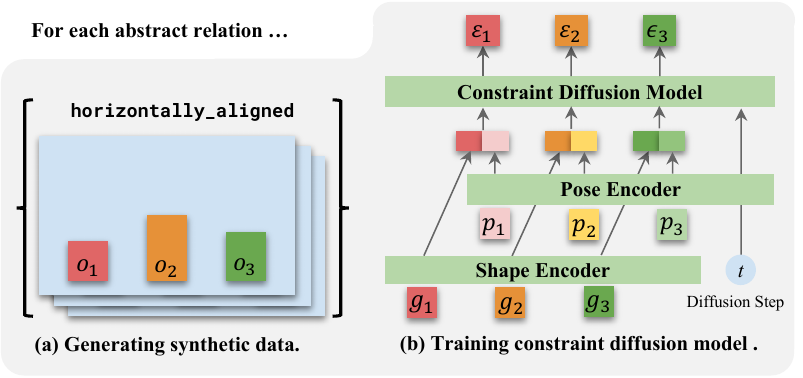}
    \caption{Training a single constraint diffusion model involves a two-stage process. First, for every abstract relation listed in Table \ref{tab:relationships}, we generate a synthetic dataset based on predefined rules. Then, we train a relation-specific diffusion model that can draw samples of object poses that satisfy the relation.}
    \label{fig:single-diffusion-model}
\end{figure}

\myparagraph{Generating spatial relations.}
One straightforward approach to the overall problem would be to hand-specify $f_R$ for each relation type and use standard non-linear optimization methods to search for $\gP$.
There are two substantial difficulties with this approach.
First, we may want to extend to relation types for which we do not know an analytical form for $f$, and so would want to acquire it via learning.
Second, the optimization problem for finding assignments to $P$ given a ground graph representation $(\gP, \gG)$ is highly non-convex and hence very difficult, and the standard method would typically require a great deal of tuning (e.g., because of the steepness of the objective near the constraint boundary).

For these reasons, we adopt a strategy that is inspired by \citet{yang2023diffusion}, which is (i) to pre-train an individual diffusion-based generative model for each relation {\em type} $R \in \mathcal{R}$ and (ii) to combine the resulting ``denoising'' gradients to generate samples of $\gP$ from the high-scoring region.
One significant deviation from the method of \citet{yang2023diffusion} is that we train diffusion models for each relation type completely independently and combine them only at inference time, using a novel inference mechanism that is both easy to implement and theoretically sound.

Our model architecture and the training paradigm are illustrated in Figure~\ref{fig:single-diffusion-model}. Specifically, for each relation type $R \in \mathcal{R}$ of arity $k$, we require a training dataset of {\em positive examples} $\gD_R = \{(g_1, \ldots, g_k, p_1, \ldots, p_k)\}$ specifying satisfactory poses $\{p_i\}$ for the given object shapes $\{g_i\}$. Note that for set-based relations, the examples in the training set may have differing arity. For all relations in our library, we generate these datasets synthetically, as described in the appendix \ref{apped:abstract_rule}.

We construct a denoising diffusion model~\citep{ho2020denoising} for each relation $R$ where the distribution $q_R(p_1, \ldots, p_k \mid g_1, \ldots, g_k)$ maximizes the likelihood $\{(g_1, \ldots, g_k, p_1, \ldots, p_k)\}$. We denote this distribution as $q_R(\vp \mid \vg)$ for brevity, where $\vp$ and $\vg$ are vector representations of the poses and the shapes, respectively. We learn a denoising function $\noise_R(\vp, \vg, t)$ which learns to denoise poses across a set of timesteps of $t$ in $\{1, \ldots, T\}$:
\begin{align*}  
\loss_{\emph{MSE}} = &\mathds{E}_{(\vp, \vg) \sim \mathcal{D}_R, \mathbf{\noise} \sim \mathcal{N}(\mathbf{0}, \mathbf{I}), t \sim U(0, T)} \\
&\left [ \left\|\mathbf{\noise} - \noise_{R}(\sqrt{\bar{\alpha}_t} \vp + \sqrt{1-\bar{\alpha}_t}\mathbf{\noise}, \vg, t) \right\|^2 \right ],
\end{align*}
where $t$ is a uniformly sampled diffusion step, $\mathbf{\noise}$ is a sample of  Gaussian noise, and $\bar{\alpha}_t$ is the diffusion denoising schedule.  In the case that $R$ has fixed arity, the network $\noise_{R}$ is a multi-layer perceptron (MLP); however, when it is set-based, we use a transformer to handle arbitrary input set sizes.  Details of these networks are provided in the appendix \ref{append:diffusion}.

\begin{figure}[tp]
    \centering
    \includegraphics[width=\columnwidth]{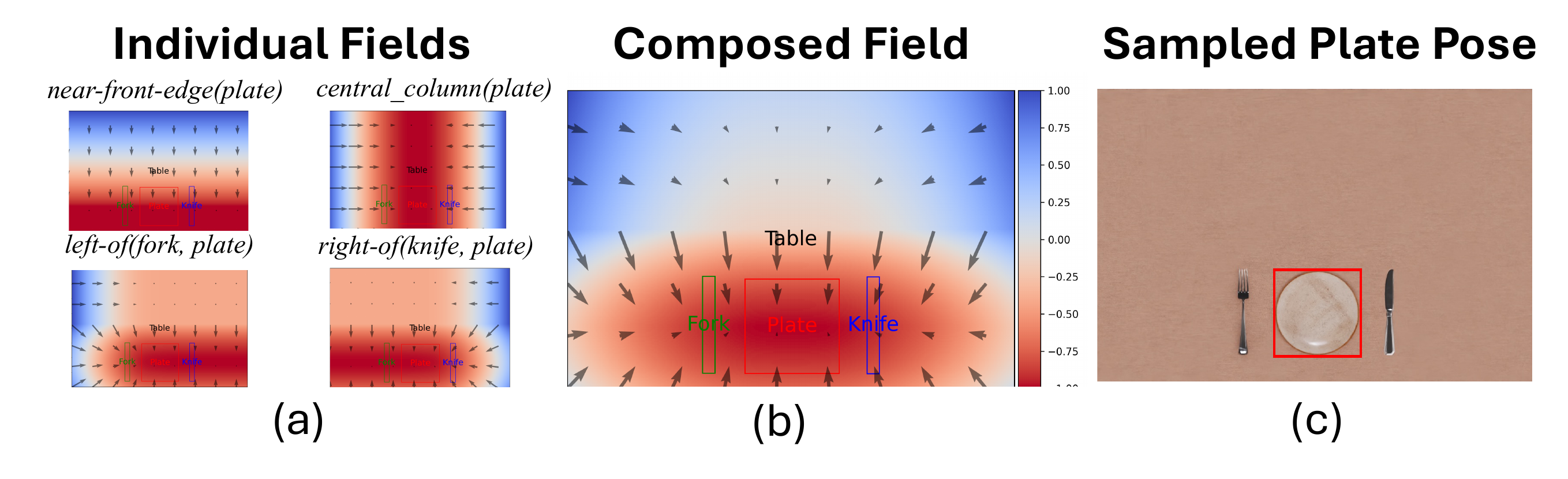}
    \vspace{-2em}
    \caption{Example illustrating the composition of diffusion models to determine object placement from multiple spatial relations. (a) Individual energy fields for four relations involving the plate, with the poses of the table, fork, and knife fixed. (b) Normalized composite energy field. (c) Sampled plate pose obtained using annealed MCMC over the composed field.}
    \vspace{-1em}
    \label{fig:composed_energy}
\end{figure} 

The denoising functions $\{\noise_R(\vp, \vg, t)\}_{t=0:T}$ represent the score of a sequence of $T$ individual distributions, $\{q_R^t(\vp \mid \vg)\}_{t=0:T}$, transitioning from $q_R^0(\vp \mid \vg) = q_R(\vp \mid \vg)$ to $q_R^T(\vp \mid \vg) = \mathcal{N}(\mathbf{0}, \mathbf{I})$. 
Therefore, to draw samples with the diffusion process, we initialize a sample $\vp_T$ from $\mathcal{N}(\mathbf{0}, \mathbf{I})$ (\ie, a sample from $q_R^T(\cdot)$). We then use a reverse diffusion transition kernel to construct a simple $\vp_{t-1}$ from distribution $q_R^{t-1}(\cdot)$, given a sample $p_t$ from $q_R^{t}(\cdot)$. This reverse diffusion kernel corresponds to
\begin{equation}
    \vp_{t-1} = B_t(\vp_t - C_t \noise_R(\vp_t, \vg, t) + D_t \mathbf{\xi} ),\; \mathbf{\xi} \sim \mathcal{N}(\mathbf{0}, \mathbf{I}),
    \label{eqn:reverse}
\end{equation}
where $B_t$, $C_t$, and $D_t$ are all constant terms and $\noise_R(\vp_t, \vg, t)$ is our learned denoising function. The final generated sample $\vp_0$ corresponds to a sample from $q_R^0(\cdot) = q_R(\cdot)$. 

\subsection{Composing Diffusion Models for Pose Generation}
\label{ssec:diffusion}

In a single diffusion model, we can generate new samples from the learned distribution by sampling an initial pose $\vp_T \sim \mathcal{N}(\mathbf{0}, \mathbf{I})$, and then repeatedly applying the learned transition kernel, sequentially sampling objects $q_R^{t-1}(\cdot)$ until we reach $\vp_0$, which is a sample from the desired distribution. 

However, in the composed diffusion model setting, our target distribution is defined by an entire factor graph, and we wish to sample from a sequence of product distributions $\{\prod_{R \in G}  q_R^t(\vp \mid \vg)\}_{t=0:T}$ starting from an initial sample $\vp_T$ drawn from $\mathcal{N}(\mathbf{0}, \mathbf{I})$. For brevity, we refer to $\prod_{R \in G} q_R^t(\vp \mid \vg)$ as $q_{\emph{prod}}^t(\vp \mid \vg)$\footnote{Here we used a simplified notation. Each relation $g \in G$ is actually a tuple of $\left(R, (o_1, \ldots, o_{k_R})\right)$. It selects a particular relation $R$ and s subset of objects to which to apply it. Thus, the corresponding distribution $q^g$ should be defined as $q_R(\vp^g \mid \vg^g) = q_R(p_1, \ldots, p_{k_R} \mid g_1, \ldots, g_{k_R})$. For brevity, we used $\prod_{R\in G} q^R(\vp \mid \vg)$ to denote the composite distribution.}. While it is tempting to use the reverse diffusion kernel in Equation~\ref{eqn:reverse} to transition to each distribution  $q_{\emph{prod}}^t(\vp \mid \vg)$, we do not have access to the score function for this distribution~\citep{du2023reduce}.

\begin{figure*}[tp]
    \centering
    \includegraphics[width=\linewidth]{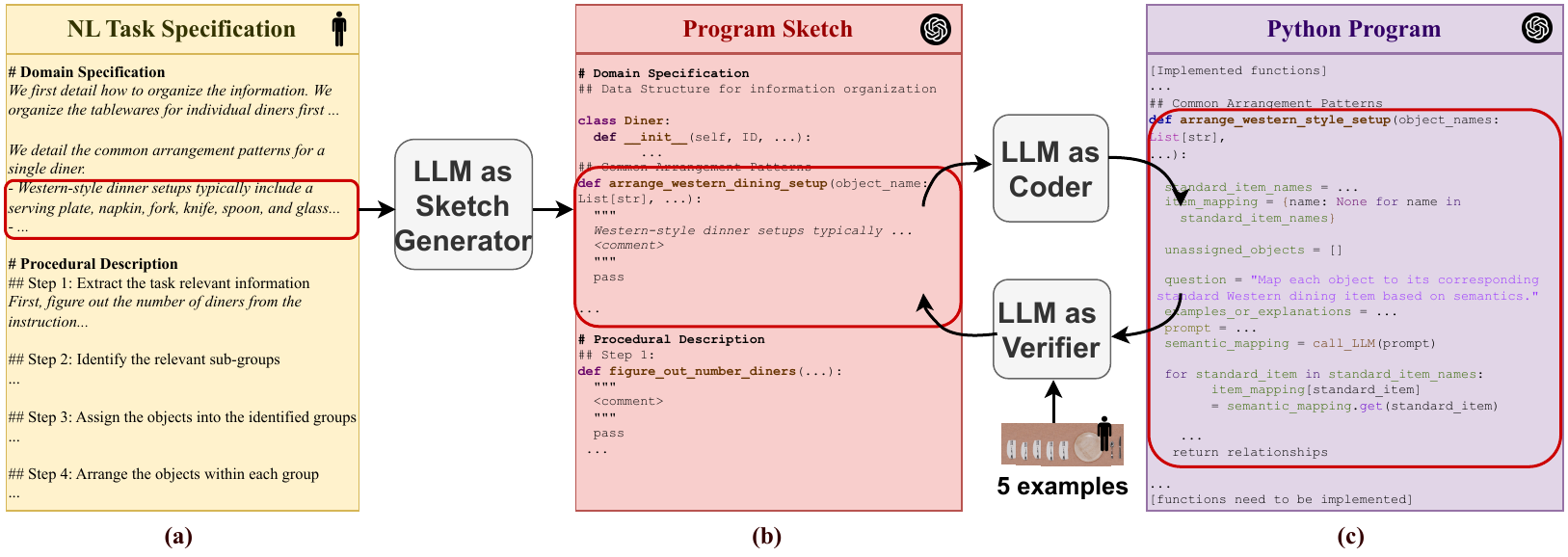}
    \caption{Compositional Program Induction using LLMs as sketch translator, coder, and verifier. For each task family, we provide a hierarchical natural language specification \emph{spec} of sub-tasks (as illustrated in (a)), along with 5 examples $\gD$. The process involves two steps: First, the LLM acts as a sketch generator, translating each sub-problem into a Python code skeleton-a program sketch specifying function names and signatures while retaining descriptions in code comments (see (b)). Next, the LLM functions as a coder and verifier, generating and verifying the code implementation one function at a time, based solely on the program sketch and the examples, as illustrated in (c).}
    \label{fig:LLM-rule-induction}
\end{figure*}

To address this, we follow the energy-based inference approach proposed by \citet{du2023reduce}, and transition between composite distributions using annealed MCMC. Specifically, we approximate the score of the product distribution $q_{\emph{prod}}^t(\vp \mid \vg)$ by summing the scores from each independently trained relation-specific diffusion model: $\sum_R \noise_R(\vp_t, \vg, t)$. This composite score enables us to perform approximate sampling to transit from $q_{\emph{prod}}^t(\vp \mid \vg)$ to $q_{\emph{prod}}^{t-1}(\vp \mid \vg)$, even though the exact reverse kernel for the product distribution is not available. Since each relation model is trained independently, their score magnitudes may differ. In practice, we address this by normalizing the score of each relation before composition. This helps ensure that no single relation dominates the composite energy due to scale differences. However, in settings where relations involve vastly different arities, it may be beneficial to further tune the composition weights to balance their relative influence. In our experiments, we found that normalization alone was sufficient to achieve effective and stable composition across diverse constraint graphs. 

\citet{du2023reduce} proposed a family of MCMC kernels for this setting, though many require careful tuning of hyperparameters for stable inference. We observe-and prove in Appendix~\ref{append:reverse_ula}-that a particularly simple variant of the Unadjusted Langevin Algorithm (ULA) can be used with minimal modification. Specifically, we reuse the same time-dependent parameters $C_t$ and $D_t$ from the original reverse diffusion kernel (Equation~\ref{eqn:reverse}), applying them to the composite score:
\begin{equation*}
    \vp_{t}' = \vp_t - C_t \sum_R \noise_R(\vp_t, \vg, t) + D_t \mathbf{\xi}, \quad \mathbf{\xi} \sim \mathcal{N}(\mathbf{0}, \mathbf{I}),
\end{equation*}
where $C_t$ and $D_t$ denote the step size and noise scale. Note that this is not the exact reverse kernel for $q_{\emph{prod}}^t(\vp \mid \vg)$-which would require the true score-but rather defines a valid MCMC step that approximately samples from it.

Our ULA-based approach enables efficient and scalable sampling from composed diffusion models. By computing a single composite score, we can perform MCMC updates at a fixed noise level, yielding a valid transition toward $q_{\emph{prod}}^t$. In practice, we replace each reverse diffusion step with $M$ MCMC updates (with $M=20$ in our experiments), all using the composite score. This corresponds to performing multiple gradient-based updates within each noise level, and is guaranteed to converge to an unbiased approximation of $q_{\emph{prod}}^{t-1}(\vp \mid \vg)$. We prove in Appendix~\ref{append:reverse_ula} that this procedure yields an unbiased approximation of $q_{\emph{prod}}^{t-1}(\vp \mid \vg)$ under mild regularity conditions. To generate a final sample from $q_{\emph{prod}}^0(\cdot)$, we begin from $\mathcal{N}(\mathbf{0}, \mathbf{I})$ and iteratively apply this MCMC-based update across all noise levels.

To illustrate this process concretely, we include a toy example in Figure~\ref{fig:composed_energy}. Here, the poses of a table, fork, and knife are fixed, and four spatial relations-\textit{central-column}, \textit{near-front-edge}, \textit{left-of}, and \textit{right-of}-are composed to determine a valid placement for a plate. We first visualize the individual energy fields associated with each relation, then show their normalized composition, and finally present a sample drawn from the composed field, illustrating a plate pose that satisfies all four relations.
\section{Grounding Graph Generation via Program Induction}
\label{ssec:llm}

Section~\ref{sec:ccsp} describes a module that takes a grounding graph description and generates the object poses. In this section, we aim to build a method that can generate this graph description $\gG$ from human instructions and examples. We achieve this by inducing a Python program from the natural language task specification \emph{spec} and very few examples $\gD$.

Interpreting under-specified instructions based on a small number of examples alone is challenging because much of the domain-specific prior knowledge required for the task is missing. As a result, LLMs perform poorly on functional object arrangement tasks when used in a few-shot in-context learning setup---where they are expected to fill in domain knowledge from the few examples alone. To address this challenge, we do not rely on few-shot conditioning over examples in $\gD$. Instead, we introduce a compositional program induction approach: we use an expert- or user-specified task description $\textit{spec}$ as an explicit specification of the domain knowledge, and translate this structured description into Python programs. This task specification only has to be specified once for every task family. These induced programs act as procedural generators, producing intermediate, verifiable outputs that guide the systematic solution of novel problems at test time. Furthermore, the examples $\mathcal{D}$ are used only during the verification phase to check the correctness of the induced program.

In our implementation, this hierarchical program is constructed compositionally through an LLM-powered modular pipeline, as illustrated in Figure \ref{fig:LLM-rule-induction}. By translating the domain knowledge from $\emph{spec}$ into a symbolic, programmatic form, and validating it against $\mathcal{D}$, we enable systematic generalization to novel instructions and object configurations. At inference time, we execute this program to generate the grounding graph for new instructions and object sets.

\begin{figure}[tp]
 \centering
\includegraphics[width=\columnwidth, page=1]{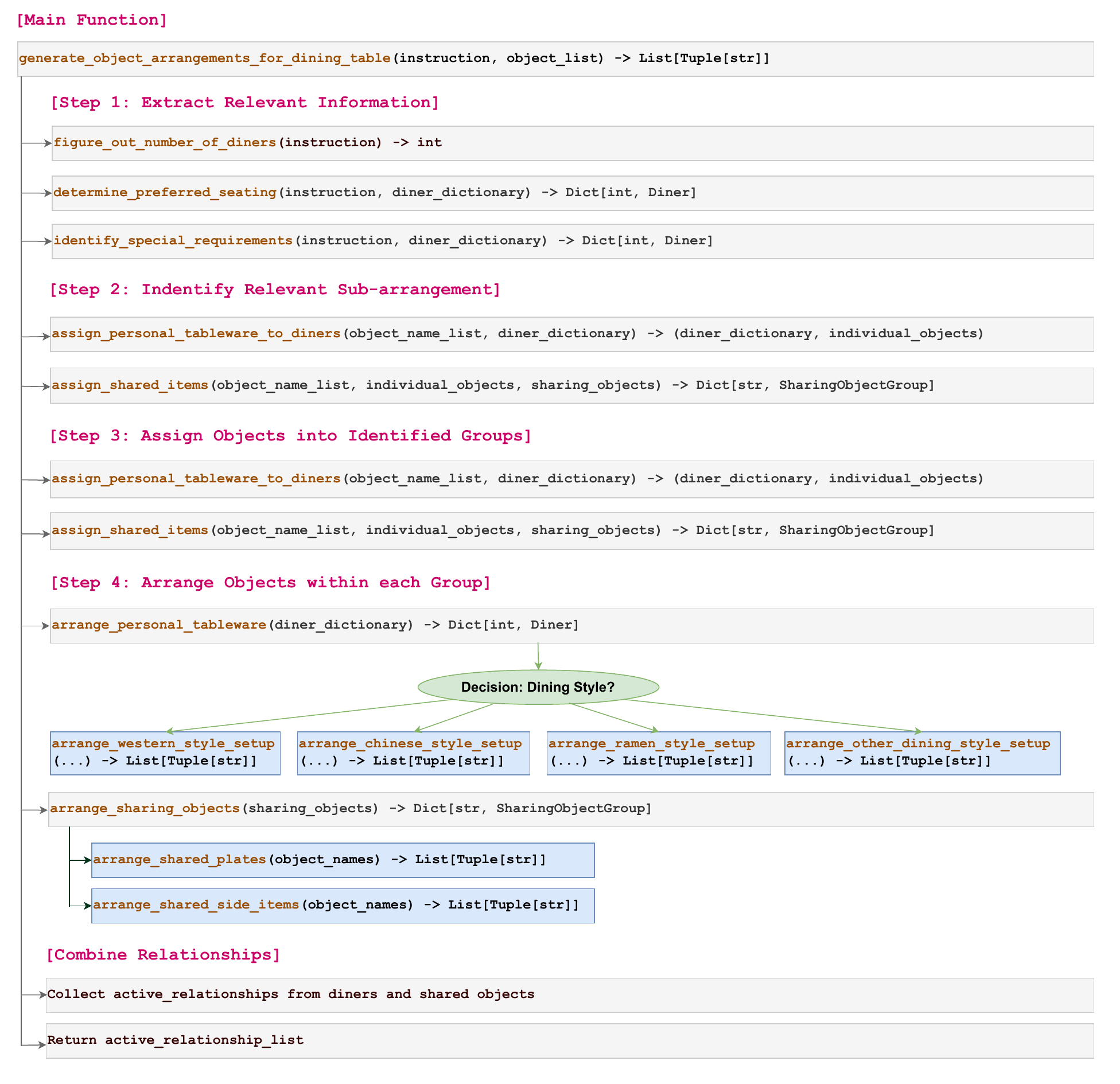}
\caption{A visualization of the generated program sketch for arranging objects on a dining table. Its computation graph is mostly sequential, with some conditional calls for common arrangement patterns.}
\label{fig:computation_graph_example}
\end{figure}

Inspired by \cite{zelikman2023parsel} and how humans can hierarchically decompose complex problems into simpler sub-problems, our program induction pipeline translates the task specification into a Python program in two steps. First, we use the LLM to convert the structured natural language specification for the task family into a {\em Program Sketch} (see Figure~\ref{fig:LLM-rule-induction}a and b). The program sketch is represented as Python code which includes function names, signatures (input and output types), and natural language descriptions of the function. They define a hierarchical reasoning procedure underlying the reasoning process, in particular, the functionality of individual modules and how they relate to each other. Figure~\ref{fig:computation_graph_example} illustrates the program sketch for setting a dining table.
Next, given the program sketch, we prompt the LLM to implement and verify each module in the sketch, based on the training examples for the given task family, as illustrated in Figure~\ref{fig:LLM-rule-induction}c. This allows us to decompose the learning problem into learning a set of smaller functional modules. Importantly, these modules can be individually generated and verified against the training examples, thereby improving the overall correctness of the final program.

\subsection{Hierarchical Natural Language Task Specification}
\label{sec:hierarchical_task_specification}

Instead of specifying tasks with programs, we use structured natural language to outline general strategies. This format preserves the expressiveness of natural language while imposing a clear organizational structure---such as decomposing subtasks and articulating object relationships. Importantly, it remains free of formal syntax, code, or schema. This is a deliberate design choice: it avoids the ambiguity of free-form instructions while eliminating the technical barrier of programming. Structured natural language offers a practical middle ground---making the system usable by non-programmers while enabling reliable program induction. This offers two key benefits: First, it reduces mental workload since describing strategies in natural language is easier than writing formal programs. Second, it allows us to leverage large language models' common-sense reasoning and completion capabilities to infer missing information. \model assumes that the task specification is composed of two main components: the {\em domain specification} and the {\em procedural description}.

\textbf{Domain Specification} outlines how to organize the intermediate results, and common recurring patterns specific to an arrangement task family. This specification can be further decomposed into two components: (i) descriptions of what information to track and how to organize and update it, and (ii) enumerations of common patterns within the task family, such as Western, Chinese, or Japanese tableware setups for a single diner.

\textbf{Procedural Description} describes how to generate the object arrangement in a hierarchical, step-by-step manner.
Overall, the procedural description follows four steps common to all task families: (i) extract relevant information from the natural language instruction, (ii) categorize objects into a set of identified groups, (iii) propose object arrangement relations within each group, and (iv) propose object arrangement relations among groups.

All task families share the same template as stated above. Each of them differs in the realization of individual steps. For some tasks, these high-level steps can further decompose them into smaller sub-steps. This decomposition continues until the subroutines are simple enough that either their solutions can be directly inferred by querying an LLM, such as directly extracting a number from the instruction, or we can articulate a fixed programmatic strategy for them. We provide an example task specification for dining table arrangement in Appendix~\ref{append:NL_task_specification}.

\subsection{Compositional Program Induction}
\label{ssec:compositional_program_induction}
Based on the natural language task specification, we adopt an LLM-empowered compositional program induction framework, as illustrated in Figure \ref{fig:LLM-rule-induction}. It has three main components:

\begin{enumerate}
    \item \textbf{Sketch Generator}: converts the structured natural language specification into a program sketch, providing the skeleton of the Python program.
    \item \textbf{Coder}: generates code for each function in the program sketch, implementing them one function at a time.
    \item \textbf{Verifier}: verifies the implemented functions against the five examples, updating the program sketch for re-generation if inconsistencies are found.
\end{enumerate}

Here, we provide a formal description of the process. The natural language task specification $\emph{spec}$ is hierarchically structured into multiple levels of simpler sub-problems, each represented by a title and its corresponding description, i.e., $(t_i, l_i)$. For example, a sub-problem might involve inferring the number of diners from the instruction, with the instruction as input and an integer as output. Our objective is to induce a Python program $P$ comprising a set of functions $\mathcal{F} = \{f_1, f_2, \ldots, f_n, f_{\text{main}}\}$, where each function $f_i$ corresponds to a section in the specification, and the main function $f_{\text{main}}$ calls these functions according to the hierarchy specified by the multi-level titles.

\myparagraph{Generating program sketch from NL Task \emph{spec}.} 
First, we use an LLM as a syntactic translator to generate a program sketch $S = \{s_i \}_{i=1}^n$ from the natural language specification $\emph{spec} = \{(t_i, l_i)\}_{i=1}^n$. This module translates natural language descriptions into code skeletons with semantically meaningful function names, signatures, and descriptions, but without implementation. The program sketch serves as an intermediate representation bridging the natural language description and the Python program. By abstracting hierarchical structures and organizing descriptions by sub-problems, it focuses on translating complex natural language into a code skeleton centered on high-level structure and function connectivity, rather than implementing functions simultaneously. This approach captures the hierarchical structure of the language specification more faithfully. Furthermore, by decomposing the task into sub-tasks with explicit input-output types, each function can be implemented and verified independently, which improves the overall reliability of the final program generation.

\paragraph{Sketch generator.} The program sketch consists of two components: a set of function definitions and the main function. We tailor \emph{SketchGenerator} to each component using different prompt templates, as detailed below. 

\begin{figure}[tp]
 \centering
    \includegraphics[width=\columnwidth, page=1]{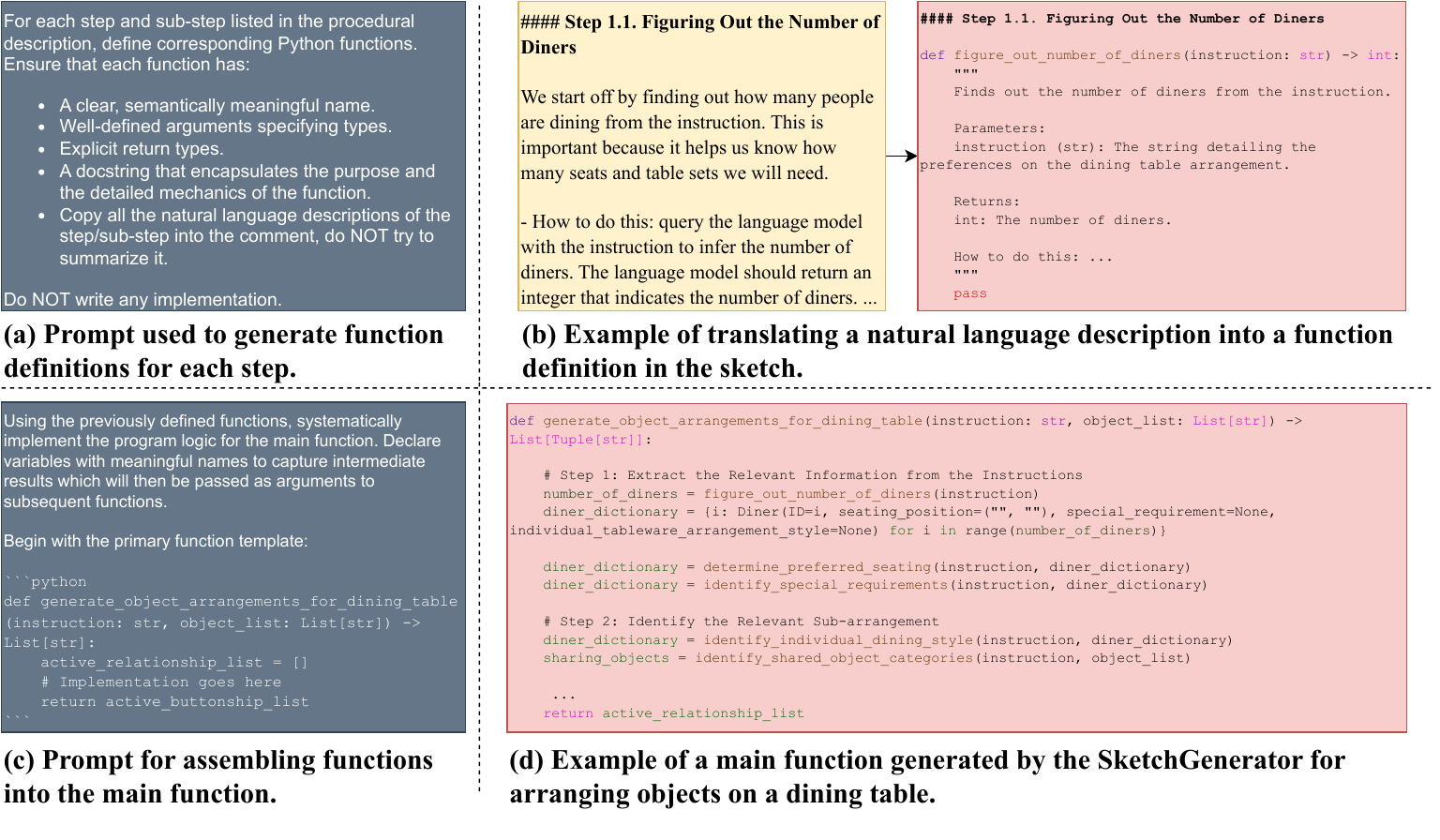}
\caption{\textit{SketchGenerator}'s prompt templates and example outputs.}
\label{fig:sketch_generator}
\end{figure}

To generate a function skeleton $s_i$ for each sub-section \((t_i, l_i) \in \emph{spec}\), we use the \emph{SketchGenerator} with prompts tailored to the function types:
\[
s_i = \emph{SketchGenerator}_{i}(t_i, l_i),
\]
where each \( s_i \) includes the function name, signature, and a description derived from the text. We design different prompt templates for translating (i) data structures, (ii) common patterns, and (iii) steps in procedural descriptions. The full prompts are listed in \ref{append:prompts_for_sketch_generator}. Figure~\ref{fig:sketch_generator}a shows the prompt template for procedural descriptions. Using this template, the \emph{SketchGenerator} extracts function names and signatures and saves all descriptions into code comments, as illustrated in Figure~\ref{fig:sketch_generator}b.

Moreover, the main function is generated by
\[
    s_{\text{main}} = \emph{SketchGenerator}_{\emph{main}}(t_1, t_2, \ldots, t_n),
\]
which takes the multi-level titles as input and assembles them into a function that resembles the hierarchy of the modules defined in the natural language \emph{spec}. Figure~\ref{fig:sketch_generator}c shows the prompt for assembling the function names into the main function, and Figure~\ref{fig:sketch_generator}d displays an example of the generated main function for setting a dining table.

\begin{figure}[tp]
 \centering
    \includegraphics[width=\columnwidth, page=1]{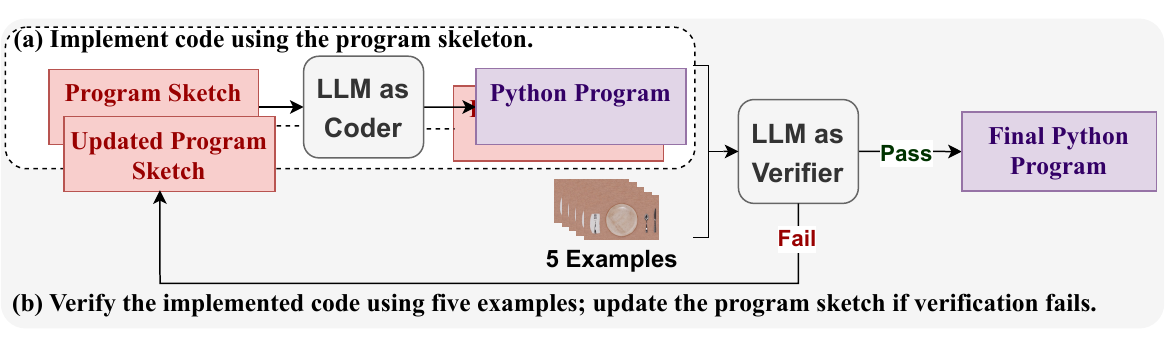}
\caption{Iterative \emph{Coder} and \emph{Verifier}. Given a program sketch \( S = \{s_1, \ldots, s_n, s_{\text{main}}\} \), we generate code one function at a time. For each function \( f_i \), \emph{Coder} takes its code skeleton \( s_i \) from the sketch-including the function signature and code comments-and generates the Python code \( c_i \) for that function (illustrated in (a)). \emph{Verifier} then checks the generated code \( c_i \) against the examples \( \mathcal{D}_i \), using the code skeleton \( s_i \), to verify consistency with the provided examples. If the code passes verification, we proceed to the next function \( f_{i+1} \); otherwise, \emph{Verifier} updates the code skeleton to \( s'_i \), and \emph{Coder} re-generates the implementation (illustrated in (b)).}
\label{fig:synthesizer_verifier}
\end{figure}

\myparagraph{Iterative coder and verifier.} Once \emph{SketchGenerator} generates the program sketch $S$, which decomposes the task into functions with names, input-output types, and descriptions, the next step is to implement and verify each function, as illustrated in Figure~\ref{fig:synthesizer_verifier}. We do this by iteratively prompting LLMs to generate and verify the code.

\paragraph{Coder.} For each function skeleton $s_i \in S$, we use an LLM as the \emph{Coder} to generate the code $c_i$ for function $f_i$ based on the sketch $s_i$:
\begin{equation*}
    c_i = \emph{Coder}(s_i).
\end{equation*}
We designed \emph{Coder} to write Python code based solely on the function name, signature, and the natural language description written as doc-string in \( s_i \) by prompting GPT-4. To effectively implement code from natural language comments, we first identify different types of descriptions. Next, we prompt \emph{Coder} on how to identify and design appropriate implementations via prompts.

There are two types of natural language descriptions: (1) fixed strategies, such as enumerating all objects and assigning them to groups, and (2) strategies requiring on-the-fly inference, such as inferring the number of diners from instructions. The first corresponds to imperative programs, where solutions are directly described, while the second corresponds to semantic reasoning which requires large language models at test time.

\begin{figure}[tp]
 \centering
    \includegraphics[width=\columnwidth, page=1]{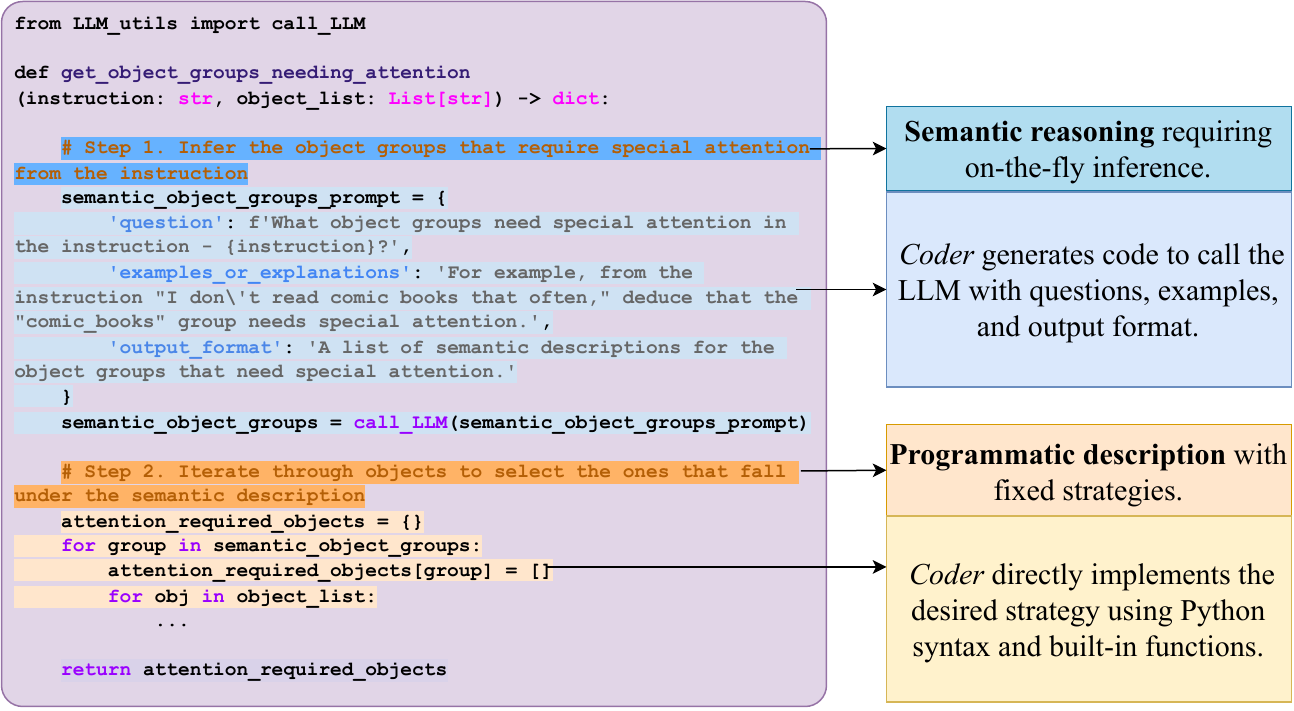}
    \caption{Python program generated by \textit{Coder}, including both programmatic strategies directly implemented in Python and on-the-fly semantic reasoning by querying LLM with specific questions.}
\label{fig:python_program_example}
\end{figure}

To handle these two types, we use two prompts to inform the LLM on how to identify and implement the descriptions. Specifically, if the description is a fixed strategy, the LLM implements the code by directly translating it to arithmetic operators in NumPy and control structures like if-else statements and loops. If the description implies inference, the LLM uses pre-defined APIs, such as \texttt{call\_llm(.)}, to query the LLM on the fly with questions, examples, and desired output formats. We include the detailed prompts in Appendix~\ref{append:coder_prompts}, and they use the following primitives: (i) LLM utility API calls such as \texttt{call\_LLM(.)} to handle semantic reasoning; (ii) the NumPy library, Python in-built functions, if-else statements and loops for controlling logic. An example of the generated function is shown in Figure~\ref{fig:python_program_example}. It includes both programmatic strategies using a for-loop and semantic reasoning implemented by calling \texttt{call\_LLM(.)} with specific questions.

\paragraph{Verifier.} We then verify $c_i$ against a set of examples $\gD_i = \{ d_{i1}, d_{i2}, d_{i3}, d_{i4}, d_{i5} \}$ using an LLM as the \emph{Verifier} with the tailor prompt documented in \ref{append:verifier_prompt}:
\begin{equation*}
    r_i,\ s_i' = \emph{Verifier}(c_i, \gD_i, s_i).
\end{equation*}
Here, $r_i \in \{\text{Pass},\ \text{Fail}\}$ indicates the result, and $s_i'$ is the updated sketch if necessary. If $r_i = \text{Fail}$, we update the sketch by setting $s_i \leftarrow s_i'$ and repeat the process for function $f_i$. If $r_i = \text{Pass}$, we proceed to the next function $f_{i+1}$. This iterative process continues until the function passes the verification or the maximum number of attempts is reached. 

The verifier updates the program by revising the code skeleton $s_i$---specifically, the code comments or task logic descriptions---rather than modifying the code $c_i$ directly. This is a deliberate design choice: discrepancies typically involve missing or commonsense details that are easier to express in descriptive language. The coder then regenerates the function based on the revised skeleton. This abstraction-level loop supports reliable correction while preserving modularity between reasoning and code generation. 

Empirically, the verifier-coder loop converges quickly, requiring only 1.2 iterations per function on average (maximum 3). Most initial generations succeed due to the specificity of the task specification. Additional iterations are usually triggered by under-specified commonsense details. For example, if the instruction says ‘place the seasoning on the side of the table,’ but all examples place it on the right, the verifier updates the skeleton to specify ‘on the right side’ to ensure alignment with the intended layout. While using fewer examples (e.g., 1 or 2) can limit the verifier’s ability to disambiguate such cases, we found that the dominant logic of most procedural tasks remains intact. In practice, the task description defines the core computational structure of the program, while the verifier and examples serve a corrective role-injecting user preferences or resolving overlooked commonsense details that are not explicitly conveyed in the task specification. 
 
\myparagraph{Variable Binding and Program Execution.}
At inference time, given a new set of objects $\gO_d$, task description $\emph{desc}$, we first use the LLM to bind these variables to the induced program. Then, we execute the program to return the set of functional abstract spatial relations.

\section{Experiments}

\begin{figure}[tp] 
 \centering
    \includegraphics[width=\linewidth, page=1]{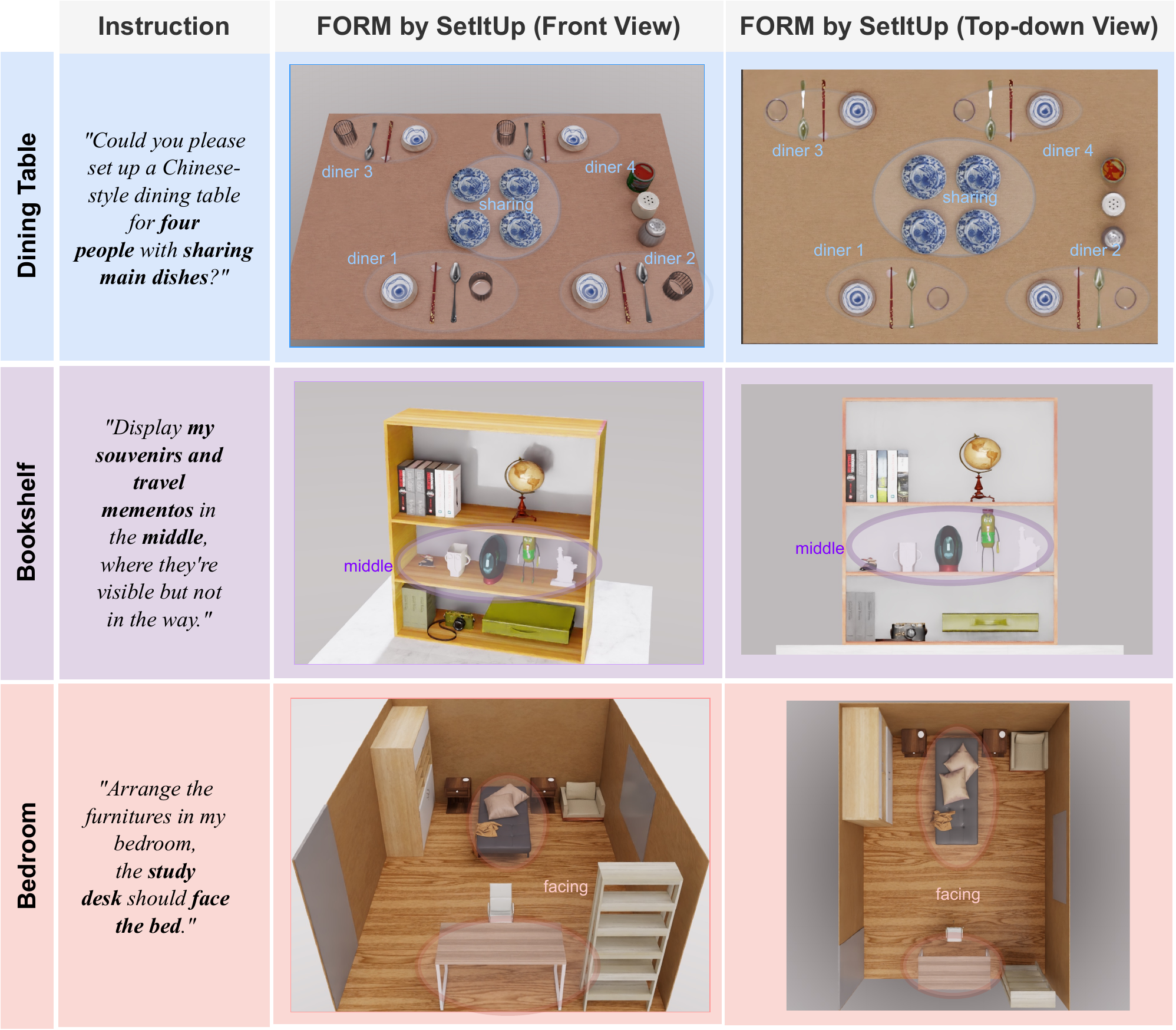}
\caption{Illustration of the Three Task Families. Functional object arrangements (FORM) generated by \model based on the instructions below.}
\label{fig:experiment-intro}
\end{figure}

We evaluated our method on three task families: dining tables, bookshelves, and bedroom layouts. These scenes involve different organizational principles, object types, aesthetic patterns, and human requirements (e.g., formal vs.\ casual dining, bookshelves for a bookworm vs.\ a cooking enthusiast). Hence, each task family presents unique challenges. 
Figure~\ref{fig:experiment-intro} illustrates examples of each task family and the functional object arrangements generated by \model.

For each task family, we generate 20 distinct scenes---5 for training and 15 for evaluation---using typical objects for that domain. Evaluation scenes contain novel combinations of instructions and object sets not seen during training, testing \model’s ability to generalize across unseen configurations. All task examples are detailed in Appendix~\ref{append:task_family}.

\begin{table*}[ht]
\centering
\caption{Comparison of methods and their use of intermediate graphs, semantic inference, and geometric inference.}
\resizebox{\textwidth}{!}{%
\begin{tabular}{l c c c}
\hline
\textbf{Method} & \textbf{Intermediate Graph} & \textbf{Semantic Inference} & \textbf{Geometric Reasoning} \\
\hline
Direct LLM Prediction (w/o task \emph{spec}) & No & \multicolumn{2}{c}{LLM, as commonsense knowledge repository} \\
Direct LLM Prediction (with task \emph{spec}) & No & \multicolumn{2}{c}{LLM, guided by task \emph{spec}} \\
End-to-end Diffusion Model & No & \multicolumn{2}{c}{CLIP embedding + Single diffusion model} \\
\hline
LLM (graph) + Comp DMs & Yes & LLM predicts the graph directly & Compositional DMs \\
LLM (program) + Comp DMs & Yes & LLM predicts the program directly & Compositional DMs \\
\textbf{Set It Up (Ours)} & \textbf{Yes} & \textbf{Compositional program induction} & \textbf{Compositional DMs} \\
\hline
\end{tabular}%
}
\label{tab:method_comparison}
\end{table*}

Figure~\ref{fig:result-showcase} presents qualitative results from \model under various instructions, while Table~\ref{tab:combined_comparison} reports quantitative comparisons across all baselines and metrics. Figure~\ref{fig:visual-comparison} visualizes side-by-side comparisons, providing insight into how compositional abstraction enhances performance relative to purely neural or LLM-based alternatives. Our results are organized by task family in Sections~\ref{sssec:dining_table}–\ref{sssec:bedroom}, where we highlight the distinct requirements of each domain, analyze five diverse outputs from \model, and compare quantitative and qualitative performance across methods.

To strengthen the connection to real-world robotics, Section~\ref{sec:real_robot_reachability} extends \model by incorporating robot reachability constraints and demonstrates its effectiveness on a dual-arm robotic platform across three dining scenarios. In Section~\ref{sec:real_robot_complex}, we validate \model in a significantly more complex buffet-style setup involving over 30 objects. Crucially, the same Python program induced from the original dining task generalizes to this scene without any prompt or template modification. These experiments highlight \model’s ability to integrate robot feasibility constraints and scale to realistic, physically constrained environments. 

Lastly, Section~\ref{sec:failure_cases} analyzes failure cases-especially in dense or spatially constrained scenes-and discusses limitations in the current geometric grounding process. 

\subsection{Baselines} \label{ssec:baselines}
We have implemented three baselines and two ablation variations of our model, as illustrated in Table~\ref{tab:method_comparison}. The implementation details are included in the Appendix~\ref{append:baseline_implementation}.

{\noindent \bf End-to-End diffusion model}: We implement an end-to-end diffusion model for language-conditioned object arrangement, inspired by StructDiffusion \citep{liu2022structdiffusion}. We trained the model using a combination of the same synthetic data and 15 \tasknameabbr training scenes (5 per task family) as our model. At test time, it directly generates object poses based on the language instructions and object shapes.

{\noindent \bf Direct LLM Prediction (w/o task \emph{spec})}: This baseline uses a large language model (LLM) to predict test object poses, inspired by Tidybot \citep{wu2023tidybot}. The model is conditioned on 15 \tasknameabbr training scenes (5 per task family), along with object categories and shapes. Without providing a natural language task specification, the arrangements rely solely on the LLM's commonsense knowledge.

{\noindent \bf Direct LLM Prediction (with task \emph{spec})}: This baseline extends Tidybot \citep{wu2023tidybot} by prompting the LLM with the same structured natural language task description as the one used for \model{} to guide the generation process. This task specification serves as an additional source of commonsense and domain knowledge for this method. The model generates object poses conditioned on 15 \tasknameabbr training scenes (5 per task family), as well as object categories and shapes.

{\noindent \bf LLM(Graph)-Diffusion}: This simplified variant of our approach omits program induction. Similar to our method, it uses an intermediate symbolic grounding graph to decompose tasks into semantic inference and geometric reasoning. Instead of inducing a Python program, it employs an LLM to directly predict the graph via few-shot learning, using structured natural language task specifications and 15 \tasknameabbr training scenes (5 per task family). The compositional diffusion model framework is then used for grounding.

{\noindent \bf LLM(Program)-Diffusion}: This variant closely follows our approach, using a simpler method for program induction. It employs an intermediate symbolic grounding graph to decompose tasks into semantic inference and geometric reasoning. For semantic inference, it leverages an LLM to generate a Python program via few-shot learning, using structured task specifications and 15 \tasknameabbr training scenes (5 per task family). The program is executed to produce an abstract grounding graph, which is grounded using the same compositional diffusion model framework.

\subsection{Evaluation Metrics}
A {\it \taskname object arrangement} must meet several criteria: physical feasibility (i.e., being collision-free and resting safely within the container), functionality (i.e., serving the intended purpose as per human instructions), and aesthetic appeal (i.e., neat arrangement with alignments and symmetries). While physical feasibility and functionality can be objectively evaluated using rule-based procedures, the subjective nature of convenience and aesthetic appeal precludes such fixed evaluations. Therefore, we employ both rule-based auto-grading and human surveys to evaluate the methods.

\begin{table}[tp]
\centering\small
\caption{Grading guidelines for human evaluation, detailing the criteria and point allocations for functionality and aesthetic appeal. Functionality is scored from 1 to 3, assessing how well the arrangement serves the task's purpose and adheres to user instructions. Aesthetic appeal is scored from 1 to 2, evaluating the visual harmony, alignment, and symmetry of the arrangement.}
\begin{tabular}{c m{7cm}}
\toprule
\multicolumn{2}{c}{\bf Functionality (1-3)} \\ \midrule
{\bf Points} & {\bf Description} \\ \midrule

{\bf 1} & \textbf{Non-functional}. The arrangement fails to serve the functional purpose of the task (e.g., dining) and does not meet the preferences or requirements specified in the instructions (e.g., seating preferences). \\ \midrule

{\bf 2} & \textbf{Partially functional}. The arrangement fulfills either the functional purpose of the task or the specified preferences, but not both. Alternatively, it inadequately satisfies both due to improper placement or lack of accessibility, failing to conform to user habits or social norms (e.g., a study desk is placed against the bed as requested but blocks the door). \\ \midrule

{\bf 3} & \textbf{Fully functional}. The arrangement properly serves both the functional purpose of the task and the preferences or requirements indicated in the instructions. \\ \midrule

\multicolumn{2}{c}{\bf Aesthetic Appeal (1-2)} \\ \midrule
{\bf Points} & {\bf Description} \\ \midrule

{\bf 1} & \textbf{Visually unappealing}. The arrangement lacks visual harmony, alignment, or symmetry. \\ \midrule

{\bf 2} & \textbf{Visually appealing}. The arrangement is aesthetically pleasing, exhibiting proper alignment, symmetry, and overall visual harmony. \\ \bottomrule

\end{tabular}
\label{tab:grading_guidelines}
\end{table}

\myparagraph{Physical feasibility:} We assess the physical feasibility score of an arrangement as the proportion of objects that are both collision-free and within their designated containers in the final layout, using a 2D collision detector. The containers are defined as follows: for dining tables, the table surface; for bookshelves, the compartments; and for bedrooms, the room itself.

\myparagraph{Basic functionality:} We measure the proportion of functional relations satisfied in the final scene configuration. For each scene, we define basic functional relations using manually specified rules. These include constraints such as ensuring key objects (\eg, utensils in a dining table setting) are within reach of the user (whose position is predetermined), arranging objects so they can serve their individual purposes (\eg, doors and windows are not obscured by furniture), and fulfilling preferences or requirements indicated in the instructions (\eg, placing all greenery to receive more sunlight).

\myparagraph{Human Evaluation of Aesthetics and Functionality:} We conducted a human evaluation with 15 graduate students. Each participant evaluated 45 scenes generated by our method or the baselines, assigning scores based on the criteria in Table~\ref{tab:grading_guidelines}. For each scene, they awarded up to 3 points for functionality and up to 2 points for aesthetic appeal, for a maximum total score of 5. We report the average scores for each method across all test cases. To facilitate evaluation, we rendered the corresponding 3D scene from 2D object poses using NVIDIA Omniverse~\citep{nvidia_omniverse}, a photorealistic simulator. Importantly, the 3D models of the objects were not used in any of the methods for proposing object poses.

\begin{figure*}[tp]
    \centering
    \includegraphics[width=\linewidth, page=1]{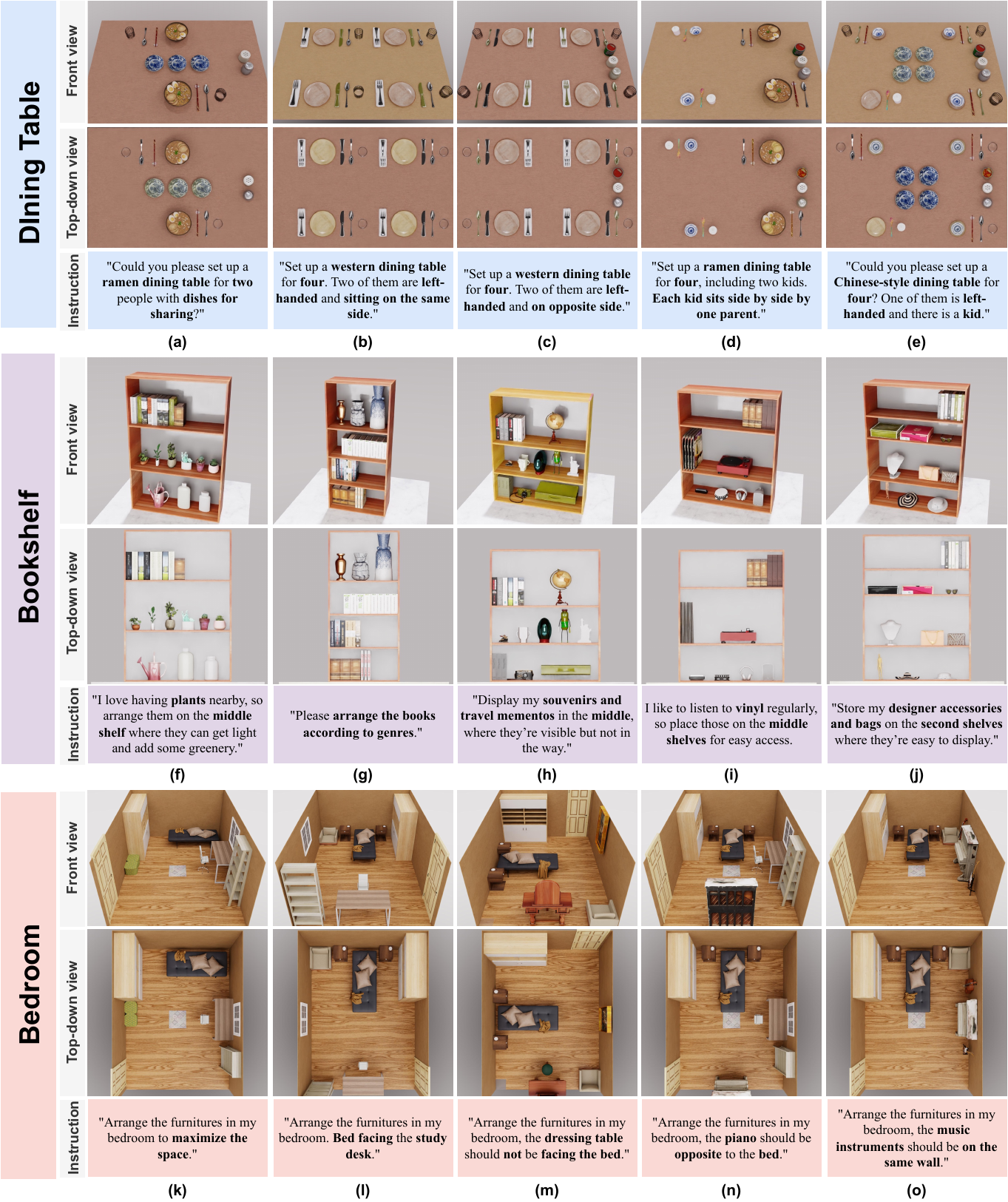}
    \caption{Examples of functional object arrangements generated by \model for the three task families under various instructions. \model produces arrangements that satisfy the instructions while being physically plausible and aesthetically pleasing. To facilitate human evaluation, we rendered 3D scenes from the proposed 2D object poses using NVIDIA Omniverse~\citep{nvidia_omniverse}.}
    \label{fig:result-showcase}
\end{figure*}

\subsection{Dining Table Arrangement} 
\label{sssec:dining_table}

\subsubsection{Challenges.} 

The dining table arrangement task presents two significant challenges. First, it involves arranging numerous objects within a hierarchical structure-setting up individual place settings for each diner while ensuring that the overall table arrangement meets group dining requirements. Second, there is a non-trivial mapping from the instructions to the final arrangement, as multiple preferences and relations must be jointly considered. For example, arranging a table may require simultaneously accommodating left-handed diners, seating preferences (such as sitting on the same side), and shared dishes, all of which must be reflected accurately in the final setup.

\subsubsection{\model on Multi-level Hierarchical Arrangement Tasks.} \model manages complex instruction-to-arrangement mappings in multi-level tasks by generating a hierarchical Python program. This program breaks down the arrangement tasks into sub-tasks and combines the local arrangements to form the complete arrangement. 

We showcase \model's ability to generate functional object arrangements for multiple diners (2–4) with novel combinations of requirements (Figures~\ref{fig:result-showcase}a--e), such as sharing a main table, accommodating left-handed diners, and seating on the same side. \model not only ensures basic functionalities-like correct utensil placement for diners sitting opposite each other (Figure~\ref{fig:result-showcase}a)-but also generalizes to accommodate more diners (Figures~\ref{fig:result-showcase}b--e), adapts to diners with different tableware setups (Figures~\ref{fig:result-showcase}d and e, where some diners are children), adjusts for left-handedness (Figures~\ref{fig:result-showcase}b, c, and e), and fulfills specific seating preferences (Figures~\ref{fig:result-showcase}b--d). 

Moreover, \model adeptly handles multiple requirements simultaneously. For example, in Figure~\ref{fig:result-showcase}b, it arranges a setup for two left-handed diners sitting on the same side (the back side) while incorporating shared dishes, demonstrating its ability to process complex instructions into coherent functional arrangements.

\subsubsection{Baseline Comparison.} 
Our method, \model, outperforms all baselines, especially the monolithic models like Direct LLM approaches and end-to-end diffusion models, by a large margin on all three criteria. This demonstrates the advantages of our neuro-symbolic framework with an intermediate representation.

\myparagraph{Hierarchically structured task specification is crucial for multi-level hierarchical arrangement tasks.} Methods that utilize hierarchically structured natural language task specifications-such as Direct LLM with specifications, our two ablation models, and \model{}-achieve significantly higher functionality scores than those that do not. This underscores the importance of incorporating structured domain knowledge for tasks that rely on cultural conventions. In particular, multi-level hierarchical arrangements like setting a dining table---where individual tableware must be arranged for each diner and globally aligned based on seating positions---require such domain knowledge to achieve correct functionality.

\myparagraph{Intermediate symbolic graphs abstract away geometric details to enable reasoning over functionality.}
While Direct LLM-based methods perform reasonably in terms of physical feasibility and aesthetic appeal, they score poorly in functionality. Without intermediate symbolic graphs to abstract geometric details, these methods greedily exploit numeric differences among the object poses to produce aligned layouts but often neglect functionalities that are not directly deducible from numeric poses. In contrast, methods using intermediate symbolic graphs-namely the two ablation methods and \model{}-enable reasoning over semantically meaningful symbolic representations, which more directly ground functionality.

\myparagraph{Compositional program induction ensures reliable intermediate graph prediction.} For the methods that utilize the intermediate graphs, \model outperforms the two ablations by fully exploiting the hierarchical structure through compositional program induction. By decomposing the arrangement task into simpler sub-tasks and solving and verifying each individually, \model enhances the reliability of its intermediate grounding graph predictions. In contrast, the ablation models merely provide the task specification as context to the LLM and predict the program or graph in a single step, often producing incomplete or conflicting graphs. This not only results in lower functionality scores but also leads to physically infeasible and less aesthetically appealing layouts, especially as the number of diners and requirements increase. Although these ablations use compositional diffusion models to ground the generated graphs, their physical feasibility scores are lower than those of the Direct LLM baselines. This counter-intuitive result stems from how the intermediate representations are generated: without verification or iterative correction, the symbolic abstractions often encode incorrect or overly constrained spatial logic. As a result, grounding through compositional diffusion is misdirected---demonstrating that the utility of intermediate representations critically depends on their accuracy.

\myparagraph{Compositional diffusion models excel at grounding extended symbolic graphs.} Both the end-to-end diffusion model and our library of diffusion models are trained on the same dataset, which includes 30,000 synthetically generated single-relation arrangements and 15 scenes of human-labeled tidy arrangements. The end-to-end diffusion model performs poorly because it trains a single diffusion model to generate the object poses, but the test arrangements are often out of distribution. Our compositional design allows our pose diffusion models to generalize robustly to scenes with more objects and relations by explicitly summing individually trained energy functions. This approach aligns with the findings of \citet{yang2023diffusion}, showing that compositional diffusion models generalize better than monolithic models.

\begin{table*}[tp]
\centering
\caption{Holistic evaluation of different methods across task families. 'Physical Feasibility' and 'Functionality' percentages are the mean over 15 diverse test scenes per task family; the best performance in each column is boldfaced. 'Human Judgments' are the mean scores with standard deviations from 15 participants evaluating the same 15 test scenes; results in bold indicate statistically significant improvement over other methods ($p < 0.05$).}
\label{tab:combined_comparison}
\setlength{\tabcolsep}{2pt}
\scriptsize
\resizebox{\textwidth}{!}{
\begin{tabular}{@{}lccccccccccccc@{}}
\toprule
\multirow{3}{*}{\textbf{Model}} & \multicolumn{4}{c}{\textbf{Dining Table}} & \multicolumn{4}{c}{\textbf{Bookshelf}} & \multicolumn{4}{c}{\textbf{Bedroom}} \\ \cmidrule(lr){2-5} \cmidrule(lr){6-9} \cmidrule(lr){10-13}
& Physical & Functionality & \multicolumn{2}{c}{Human Judgment} & Physical & Functionality & \multicolumn{2}{c}{Human Judgment} & Physical & Functionality & \multicolumn{2}{c}{Human Judgment} \\ \cmidrule(lr){4-5} \cmidrule(lr){8-9} \cmidrule(lr){12-13}
& Feasibility (\%) & (\%) & Functionality (1-3) & Aesthetics (1-2) & Feasibility (\%) & (\%) & Functionality (1-3) & Aesthetics (1-2) & Feasibility (\%) & (\%) & Functionality (1-3) & Aesthetics (1-2) \\ \midrule

\textbf{Direct LLM (w/o spec)} & 63.1 & 40.7 & 1.83$_{\pm 0.76}$ & \textbf{1.89}$_{\pm 0.26}$ & 50.3 & 90.7 & 2.24$_{\pm 0.45}$ & 1.34$_{\pm 0.32}$ & 67.6 & 34.4  & 1.53$_{\pm 0.64}$ & 1.73$_{\pm 0.46}$\\

\textbf{Direct LLM (with spec)} & 82.1 & 67.7  & 2.12$_{\pm 0.70}$ & 1.75$_{\pm 0.25}$ & 53.5 & 90.2 & 2.23$_{\pm 0.56}$ & 1.43$_{\pm 0.32}$ & 68.3 & 37.4 & 1.60$_{\pm 0.64}$ & 1.83$_{\pm 0.47}$\\

\textbf{End-to-End Diffusion} & 22.7 & 21.8 & 1.13$_{\pm 0.35}$ & 1.20$_{\pm 0.41}$ & 86.8 & 31.2 & 1.32$_{\pm 0.54}$ & 1.43$_{\pm 0.32}$ & 17.0 & 25.5  & 1.27$_{\pm 0.43}$ & 1.26$_{\pm 0.45}$\\ \midrule

\textbf{LLM(Graph)-Comp DMs} & 45.2 & 67.3 & 2.43$_{\pm 0.52}$ & 1.23$_{\pm 0.49}$ & 87.0 & 93.2 & 2.87$_{\pm 0.34}$ & 1.63$_{\pm 0.43}$ & 57.7 & 60.3 & 2.07$_{\pm 0.35}$ & 1.53$_{\pm 0.52}$  \\

\textbf{LLM(Program)-Comp DMs} & 53.3 & 70.8 & 2.62$_{\pm 0.46}$ & 1.30$_{\pm 0.50}$ & 90.0 & 93.5 & \textbf{2.92}$_{\pm 0.25}$ & 1.65$_{\pm 0.65}$ & 45.6 & 78.2 & 2.27$_{\pm 0.59}$ & 1.40$_{\pm 0.51}$\\

\textbf{\model (Ours)} & \textbf{94.0} & \textbf{89.1} & \textbf{2.86}$_{\pm 0.35}$ & 1.87$_{\pm 0.34}$  & \textbf{98.1} & \textbf{94.4} & \textbf{2.92}$_{\pm 0.23}$ & \textbf{1.95}$_{\pm 0.21}$ & \textbf{93.8} & \textbf{91.5} & \textbf{2.87}$_{\pm 0.35}$ & \textbf{1.87}$_{\pm 0.34}$\\ \bottomrule

\end{tabular}
}
\end{table*}

\begin{figure*}[tp]
    \centering
    \includegraphics[width=\linewidth, page=1]{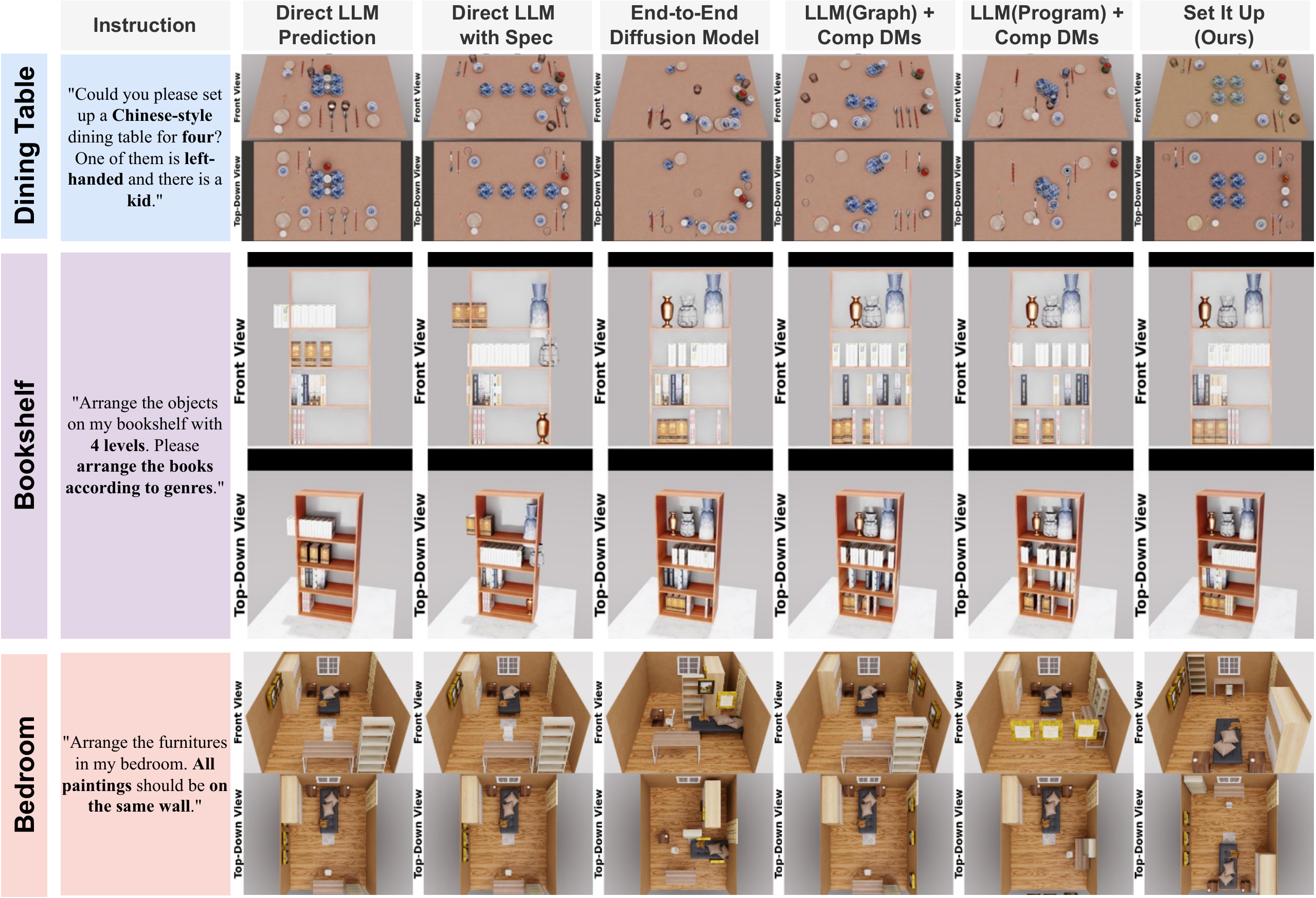}
    \caption{Illustrations of the final arrangements generated by our method and the baselines. Our model consistently generates more physically plausible, functional, and aesthetically pleasing object arrangements. We rendered 3D scenes from the proposed 2D object poses using NVIDIA Omniverse~\citep{nvidia_omniverse} to facilitate human evaluation.}
    \label{fig:visual-comparison}
\end{figure*}

\subsection{Bookshelf Arrangement}
\label{sssec:bookshelf}

\subsubsection{Challenges.} 
The primary challenge in this task family is the enormous variety of objects that need to be organized. These objects range from common items such as books, plants, and vases to highly specialized items reflecting individual preferences, such as small instruments for music lovers, souvenirs for global travelers, and designer boxes and accessories for fashion enthusiasts. It is impractical to cover all these object types-or even similar types-in the limited number of training examples. Moreover, there are no one-size-fits-all organizing principles due to the diverse compositions of items and personal preferences.

\subsubsection{\model on Tasks That Require On-The-Fly Semantic Inference.}
\model is capable of organizing diverse objects using adaptable, context-aware grouping strategies. This is achieved by leveraging LLMs to understand the semantic requirements of instructions and to infer grouping strategies on the fly.

To illustrate how \model deals with these challenges, we present five examples (Figure~\ref{fig:result-showcase}f--j) that showcase its ability to interpret diverse instructions and organize objects accordingly. In the first two examples, even though they both have a set of books, the organizing principles differ based on the contexts and preferences. In Figure~\ref{fig:result-showcase}f, to arrange for a nature lover, the plants are positioned in the middle compartments to receive more light base on the user preference, and the books are placed on the top shelf. In contrast, in the case of a bookworm (Figure~\ref{fig:result-showcase}g), the books are grouped by genre and placed in different compartments accordingly. These examples highlight that even with the same type of objects, \model can apply different organizing principles tailored to the users' contexts and instructions.

The subsequent examples demonstrate the diverse scenarios \model can handle. Figure~\ref{fig:result-showcase}h showcases a global traveler, Figure~\ref{fig:result-showcase}i a music enthusiast, and Figure~\ref{fig:result-showcase}j a fashion lover. These examples illustrate \model's ability to accommodate a wide variety of objects-none of which overlap with or are similar to those in the training examples-and to apply unique organizing principles across diverse contexts, resulting in personalized and coherent bookshelf arrangements.

\subsubsection{Baseline Comparison.}
\model achieves nearly perfect performance across all evaluation metrics: functionality, physical feasibility, and aesthetic appeal. Baseline methods that use LLMs for inference attain comparable performance in functionality, while methods utilizing diffusion models for geometric reasoning score higher in physical feasibility and aesthetic appeal.

\myparagraph{Querying LLMs enables on-the-fly semantic inference.} Methods that employ LLMs for on-the-fly semantic inference-including the two variants that directly prompt the LLM for object poses, our two ablations that query the LLM to generate abstract grounding graphs, and our method-significantly outperform methods incapable of querying the LLM for semantic reasoning (e.g., the end-to-end diffusion model) in terms of functionality. This is because the primary challenge in this task family is generating context-aware and personalized groupings of diverse objects and assigning them to the appropriate shelf levels. Such open-ended yet straightforward reasoning can be effectively performed by querying LLMs. Consequently, hierarchical task specifications or compositional program induction do not result in substantial performance differences.

\myparagraph{Explicit optimization is crucial for geometric reasoning in varying environmental contexts.} Methods employing explicit optimization techniques-specifically diffusion models, both compositional and end-to-end-achieve significantly higher physical feasibility and aesthetic appeal scores than methods that rely on LLMs to directly predict object poses. This is because, in tasks with varying environmental contexts (e.g., shelves of different dimensions), LLMs may struggle to interpret these contexts, resulting in physically infeasible object placements. In contrast, explicit optimization methods are specifically designed to perform such geometric reasoning. For example, in Figure~\ref{fig:visual-comparison}, Direct LLM methods propose object poses that are outside the bookshelf and floating in mid-air, whereas methods using diffusion models place all items within the bookshelf compartments.

Furthermore, in this task family, methods using compositional diffusion models perform on par with the monolithic end-to-end diffusion model because the abstract relation graphs for bookshelf arrangements are simple; they involve only a few symbolic relations, which does not fully exploit the extensibility of the compositional design.

\subsection{Bedroom Layout} 
\label{sssec:bedroom}

\subsubsection{Challenges.} The main challenge lies in generating interdependent constraints that are all valid in a compact environment. In small spaces, furniture relations are highly entangled; placing one piece affects others. Each decision must consider all constraints to ensure the arrangement is feasible. For example, positioning a bed in a certain spot consumes space and influences where wardrobes or desks can go.

Additionally, the room's environmental context-the locations of windows and doors-adds further constraints. Furniture must not block doors or windows, and placing a bed under a window might be undesirable due to humidity.

Specific requirements in the instructions further compound these challenges by adding more constraints. Together, these factors make predicting a feasible arrangement difficult. Unlike arranging a dining table, we cannot break down the task into independent sub-tasks; in a compact space, all decisions are closely intertwined.

\subsubsection{\model on Tasks with Intricately Entangled Constraints.}  \model can generate valid arrangements despite intricately entangled constraints in compact spaces. This is achieved by using an abstract grounding graph as an intermediate representation. On the one hand, this graph unifies all constraints from multiple sources-functional requirements, instructions, and environmental context; on the other hand, it allows for easy and efficient editing and checking of intermediate arrangements.

To demonstrate how \model handles these challenges, we present Figure~\ref{fig:result-showcase}k--o that highlight its ability to generate physically feasible and functional furniture arrangements under various constraints and preferences. In Figure~\ref{fig:result-showcase}k, \model addresses a global preference of ``maximizing open space'' by strategically arranging essential furniture to create a sense of spaciousness without compromising basic functionality. It accommodates specific positive relations among furniture, such as positioning one piece to face another (Figure~\ref{fig:result-showcase}l and n), and specific negative relations, like ensuring one piece does not face another (Figure~\ref{fig:result-showcase}m), all while respecting spatial constraints and maintaining functionality. Additionally, \model manages relative preferences defined over a class of objects with preferred locations-for instance, arranging all paintings on the same wall.
These examples collectively showcase \model's proficiency in generating valid arrangements that adhere to the instructions and respect the environmental context, including the locations of doors and windows. 

\subsubsection{Baseline Comparison.}

\model achieves outstanding performance, scoring over 90\% in functionality, physical feasibility, and aesthetic appeal. Two of our ablation models that utilize symbolic relations as intermediate representations also outperform other baselines in functionality. On the other hand, methods that directly use LLMs to predict object poses score better in physical feasibility and aesthetic appeal. The end-to-end diffusion model, which neither employs intermediate representations nor leverages LLMs, ranks last across all evaluation metrics.

\myparagraph{Grounding graph unifies complex constraints and eases their management.} Methods that use intermediate symbolic graphs-our two ablation models and \model-achieve higher functionality scores than monolithic models like direct LLM prompting and the end-to-end diffusion model. This highlights the importance of using an intermediate symbolic representation: it unifies constraints from multiple sources, enabling efficient and accurate checking and updating of object relations.

\myparagraph{Intermediate representation mitigates overfitting.} Although direct LLM methods rank second after \model in physical feasibility and aesthetic appeal, this is not due to superior geometric reasoning but because they overfit to the training examples. Due to similar furniture categories in both training and testing sets, these methods-using few-shot prompting with limited data-tend to replicate layouts from the training data. As illustrated in Figure~\ref{fig:visual-comparison}, the bedroom layouts generated by the LLM methods are nearly identical. This overfitting yields high aesthetic and physical feasibility scores but neglects specific functional requirements in the instructions and changing environmental contexts. This highlights the importance of symbolic abstraction in facilitating functional reasoning and preventing naive copying of object poses from training examples.

\myparagraph{Compositional diffusion models flexibly ground minor variations in similar symbolic graphs.} Because the furniture set remains largely consistent across test cases-typically including beds, wardrobes, and tables-the main challenge is grounding similar symbolic graphs into different object poses within varying environmental contexts, such as varying door and window locations. Methods using compositional diffusion models for geometric grounding significantly outperform end-to-end diffusion models. This is because \model composes diffusion models strictly according to the symbolic graphs, therefore explicitly capturing slight variations in the composed energy landscape. In contrast, a monolithic diffusion model directly maps instructions to object poses, making it less clear whether such variations are adequately accounted for in the resulting energy function. This highlights the importance of using compositional diffusion models to flexibly ground slight variations in symbolic graphs.

\subsection{Real-Robot Experiment: Incorporating Reachability Constraints}\label{sec:real_robot_reachability}

\begin{figure*}[tp]
    \centering
\includegraphics[width=\linewidth]{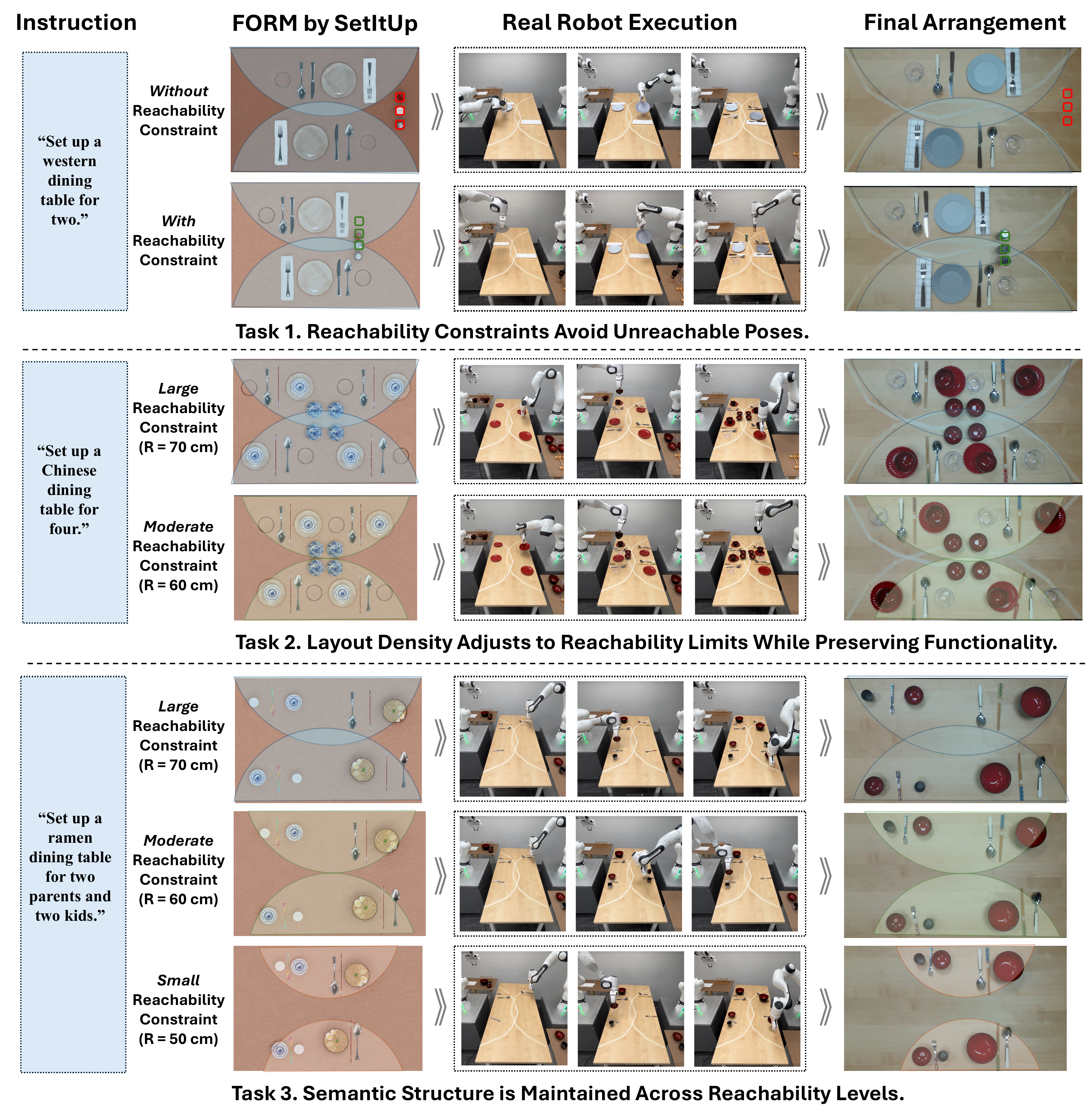}
    \vspace{-2em}
    \caption{SetItUp adapting to reachability constraints across three dining scenarios. In the western dining setup (Task 1), reachability-aware execution repositions items to avoid unreachable placements (highlighted in red). In the Chinese dining setup (Task 2), layouts vary in density as the reachability radius changes between 60cm and 70cm. In the ramen dining setup (Task 3), layout density increases progressively as the reachability constraint tightens from 50cm to 70cm. Across all scenarios, SetItUp preserves functional grouping and layout intent, demonstrating generalization to physical feasibility constraints in real dual-arm robot deployments.}
    \label{fig:reachability}
    \vspace{-1em}
\end{figure*}

While the primary focus of \model is to generate functional object arrangements that fulfill user-specified goals, ensuring that these arrangements are also feasible for robotic execution is critical for real-world deployment. So far, \model serves as a goal-level specification system, producing object configurations that can be used by downstream planning and control modules. It does not generate manipulation trajectories or support online replanning. However, our system is modular and extensible: we demonstrate how \model can be extended to incorporate robot-specific feasibility constraints, focusing here on \emph{reachability}, and validate this capability on a real dual-arm robotic platform.

To achieve this, we introduce a new abstract relation, $\text{\em reachable}_R(o_i, a_j)$, indicating that object $o_i$ must lie within a reachability radius $R$ of robot arm $a_j$. We consider three levels of reachability: large (70 cm), moderate (60 cm), and small (50 cm). Abstract reachability relations are established based on object positions relative to the robot arms: in single-arm setups, all objects must be reachable by the arm, while in dual-arm setups, objects are assigned to the nearest arm, with centrally located objects distributed evenly. These reachability constraints are combined with the spatial and functional relations already modeled by \model, forming a unified constraint graph. We train additional diffusion models conditioned on robot base positions, object geometry, and reachability constraints. Training data is generated by relabeling the synthetic dataset used for \model, annotating whether each object is within a specified radius of the robot base.

\begin{figure*}[tp]
    \centering
\includegraphics[width=\linewidth]{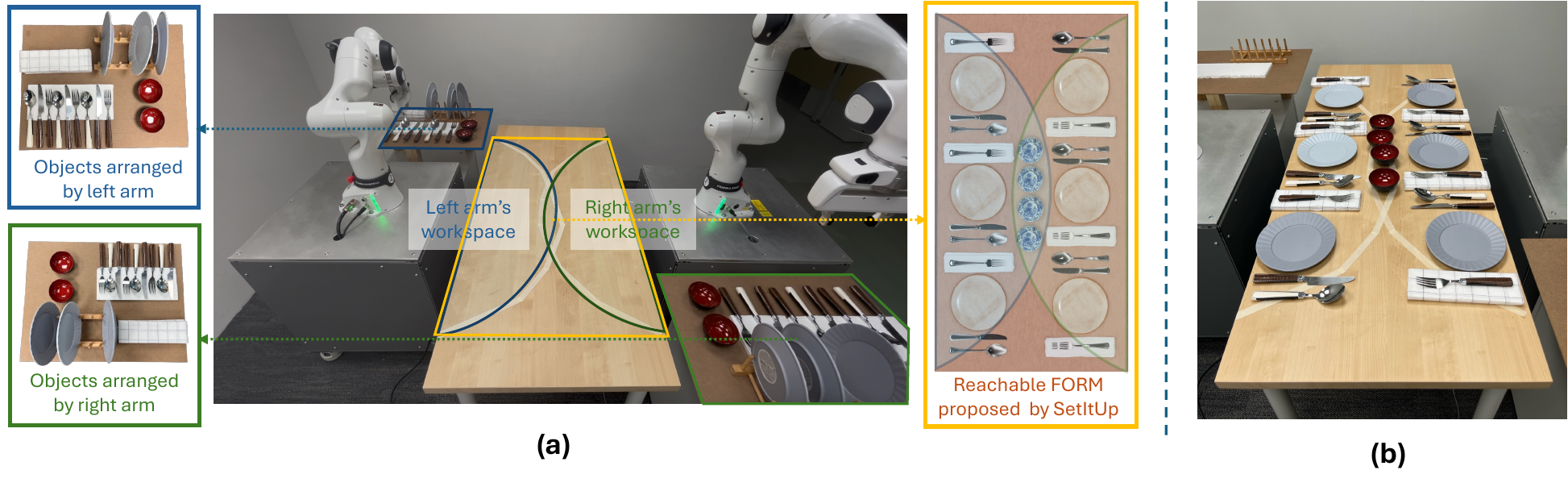}
    \vspace{-2em}
    \caption{SetItUp evaluated in a complex buffet-style dining scene for six diners, with over 30 objects densely arranged across a long table. (a) Two Franka Research 3 (FR3) robotic arms are assigned objects based on symbolic reachability constraints. (b) The system successfully arranges the table, demonstrating SetItUp’s ability to generalize to realistic, physically constrained environments without additional prompt tuning.}
    \label{fig:complex_scene_real_robot}
    \vspace{-1em}
\end{figure*}

We validate the reachability-aware extension of \model on a dual-arm robot system, where two fixed-base manipulators are mounted on opposite sides of a dining table, as illustrated in Figure ~\ref{fig:complex_scene_real_robot} (a). Each arm has limited reach, and their combined workspaces only partially cover the table. We evaluate \model under identical task instructions and object sets, comparing results generated with and without reachability constraints, and across different reachability levels. Qualitative results are shown in Figure~\ref{fig:reachability}.

In the western dining setup (Figure~\ref{fig:reachability}, Task 1), \model without reachability places seasonings at the table’s edges-unreachable by either arm. With constraints (R = 70 cm), these items are repositioned closer to the center, ensuring accessibility while preserving functional layout.

In the Chinese dining setup (Figure~\ref{fig:reachability}, Task 2), \model generates a spacious layout at R = 70 cm, while the layout becomes denser at R = 60 cm, adapting to workspace constraints while preserving task semantics.

In the ramen dining setup (Figure~\ref{fig:reachability}, Task 3), decreasing reachability results in progressively denser layouts. Nonetheless, adult-child groupings and utensil placement remain intact, showing that \model systematically adapts layout density in response to physical constraints.

These results demonstrate that \model can integrate robot-specific feasibility constraints to generate object configurations that are both functionally meaningful and physically feasible. By validating the system on real hardware, we illustrate its potential as a bridge between abstract goal generation and practical robotic execution in complex, multi-object environments.

\subsection{Real-Robot Experiment: Generalization to Complex Scene}
\label{sec:real_robot_complex}

The \taskname object arrangements proposed by \model can be seamlessly integrated into standard robotic manipulation pipelines, enabling real-world execution of high-level instructions without manual specification of object layouts. To demonstrate this, we deploy \model in a challenging, realistic setting: a buffet-style dining scene for six diners, involving over 30 objects densely arranged across a long table, executed using two robotic arms (Figure~\ref{fig:complex_scene_real_robot} (a)).

\model outputs exact 2D object poses that serve as goal configurations for downstream perception, planning, and control. In our experiment, we assume to know the objects' initial poses and types; in general, one can detect them using a camera. Given the instruction “set up a western-style buffet table for six diners,” \model synthesizes a symbolic program that determines the functional arrangement logic, and grounds this layout into concrete object poses via compositional diffusion models. These target poses are executed by two Franka Research 3 (FR3) robotic arms using standard motion planning.

This scenario presents substantially higher complexity than prior evaluation in terms of object count, spatial tightness, and functional groupings. Crucially, \model handles this task using the same task-specific Python program and prompt templates induced from the original dining table specification---demonstrating that no additional prompt tuning or template modification is needed to adapt to the novel configuration. For example, the template for assigning seating arrangements remains fixed; only the input variable (i.e., 6 diners) is updated dynamically during inference. The symbolic program logic, previously induced from the natural language specification, invokes the correct LLM query automatically based on the instruction argument.

This scene also introduces physical feasibility challenges: with over 30 objects densely packed across a long table, no single robotic arm can feasibly reach all placements. To address this, we extend \model to support robot-specific reachability constraints, as described in Section \ref{sec:real_robot_reachability}. The reachability constraints are assigned by associating each object with its nearest robotic arm under a predefined reachability radius. In this case, we use a dual-arm system with manipulators mounted on opposing table sides, as illustrated in Figure~\ref{fig:complex_scene_real_robot} (a). Central objects are divided between arms based on proximity, and the abstract grounding graph is augmented to reflect these arm-specific reachability assignments. These relations are then grounded using additional diffusion models trained to generate poses within the reachable workspace of each robot arm.

As shown in Figure~\ref{fig:complex_scene_real_robot} (b), the robots successfully execute the layout, satisfying both functional semantics and manipulation feasibility. This experiment confirms that \model’s goal specifications are compatible with real-world execution pipelines, even in large, physically constrained, multi-arm environments. Implementation details are provided in Appendix~\ref{ssec:robot}.

\subsection{Failure Case Analysis}
\label{sec:failure_cases}

While \model achieves the highest physical feasibility scores across all task families (94.0\% for dining tables, 98.1\% for bookshelves, and 93.8\% for bedrooms; see Table~\ref{tab:combined_comparison}), some failure cases remain. These failures are most common in compact scenes---particularly in dining table and bedroom arrangements---where objects must satisfy multiple spatial constraints within tight spatial bounds. Failures typically involve slight object-object collisions or marginal placement outside designated containers.

The root cause lies in the compositional nature of the spatial reasoning process: \model uses separate diffusion models to satisfy multiple abstract spatial relations predicted in the grounding graph. When a single object is subject to several overlapping or competing constraints, the joint sampling space can become narrow or conflicting. Although the composed energy functions guide sampling toward low-energy regions, the resulting poses may only approximately satisfy all constraints, occasionally violating physical feasibility. These issues highlight a limitation of the current independent grounding setup. Addressing this may require joint training of relation-specific diffusion models or more sophisticated sampling methods (e.g., constrained MCMC) to better explore feasible joint regions.
\section{Conclusion}

We present \model, a neuro-symbolic model that enables humans to specify goals for functional object arrangement through natural language instructions. \model interprets underspecified instructions and generates precise object poses, resulting in arrangements that fulfill the intended functional purpose while being physically feasible and aesthetically pleasing. Instead of requiring a large amount of human-annotated data, \model needs only two pieces of information: a hierarchically structured natural language task description that outlines the general strategy for solving the arrangement task for a task family, and five examples of instruction-demonstration pairs.

At the core of \model is the use of a symbolic grounding graph as an intermediate representation, decomposing the task into two sub-problems: semantic inference to interpret instructions and predict the grounding graph, and geometric reasoning to ground this graph into exact object poses. We propose a hierarchical program induction pipeline that translates the task description into a Python program-verified by the five examples-that reliably generates the grounding graph for novel instructions and object sets. This graph is then grounded by flexibly composing individual diffusion-based generative models corresponding to single abstract relations.

By leveraging its multi-level compositional design, \model reliably generates functional object arrangements for dining tables, bookshelves, and bedroom furniture-each task family posing unique challenges. Rule-based and human evaluations show that our model outperforms baselines-especially pure LLM-based and end-to-end diffusion models-by producing object placements that are more physically feasible, functional, and aesthetic.

\model opens up a new paradigm for generating functional object arrangements, but also comes with limitations that point toward valuable future directions.
First, although structured natural language specifications are far more accessible than formal code, writing them still demands effort and domain expertise. To reduce this burden, we envision interactive authoring interfaces where the system engages users through suggestions, clarifying questions, or partial completions. This would shift specification from a static document to a collaborative dialogue, moving \model further along the spectrum toward intuitive, conversational task design. Second, our current coder-verifier pipeline assumes that the task specification is logically consistent. While the verifier is effective at recovering missing details, it does not yet address ambiguities or contradictions in user input task specifications. Future work could leverage advances in robust program synthesis-e.g., self-debugging, multi-turn generation, and verification-aware language models-to enhance the reliability of code generation from natural instructions ~\citep{bavarian2022codex, kulal2022spider, qin2023codegen, hong2024codeagent}. Third, extending \model to 3D scenes introduces challenges such as reasoning about support, elevation, and physical stability, along with the need for richer 3D training data. Prior work has demonstrated the feasibility of compositional diffusion in 3D ~\citep{yang2023diffusion}, and our symbolic abstraction layer provides a strong foundation. We are currently extending \model to generate multi-level, physically stable 3D arrangements. 


\onecolumn
\appendix
\section{Discussion of Differences with the Conference Version} \label{append:open_discussion}

This manuscript is an extension of our previous work presented at the Robotics: Science, and Systems (RSS) conference, titled “Set It Up: Functional Object Arrangement with Compositional Generative Models”~\citep{Xu-RSS-24}. We have significantly expanded upon the conference version in two primary aspects:

\begin{enumerate}
    \item \textbf{Introduction of a new program induction pipeline that accepts structured natural language task specification as input}, improving usability and reliability (Section~\ref{ssec:llm}).
    \item \textbf{Expansion of experimental evaluations to diverse task families}, demonstrating the generality and robustness of our framework (Section~\ref{sssec:bookshelf} and ~\ref{sssec:bedroom}).
\end{enumerate}

\subsection*{1. Introduction of a New Program Induction Pipeline with Structured Natural Language Input}

In the conference version, the system required task specifications in the form of a Python code skeleton, which was less intuitive and user-friendly. In this manuscript, we have developed a novel program induction pipeline that accepts \emph{structured natural language} task specifications as input, as presented in Section \ref{ssec:llm}. This enhancement significantly improves user accessibility and usability by allowing users to specify tasks in natural language.

We frame semantic inference as a compositional program induction problem (Section~\ref{ssec:program_induction_overview}) using hierarchical natural language task specifications (Section~\ref{sec:hierarchical_task_specification}) and a small number of examples. We propose a hierarchical, modular framework that utilizes LLMs as syntactic translators, coders, and verifiers (Section~\ref{ssec:compositional_program_induction}). This represents a substantial advancement over previous few-shot prompting methods. The hierarchical design systematically decomposes the induction process, enhancing the reliability and accuracy of program generation.

We have conducted additional experimental evaluations with corresponding baselines to demonstrate the effectiveness of this hierarchical design (details in Section~\ref{ssec:baselines}). The results highlight the improved reliability in translating structured natural language task specifications into hierarchical Python programs for abstract grounding graph generation, validating the significance of our approach.

\subsection*{2. Expanded Experimental Evaluation across Diverse Task Families}

Beyond the tabletop environment evaluated in the conference version, we have extended our experimental evaluations to include two more complex task families.

\begin{itemize}
    \item \emph{Personalized Bookshelf Arrangement (Section~\ref{sssec:bookshelf}):} We introduce tasks that involve the arrangement of open-set objects on bookshelves with varying compartment configurations. This setting poses unique challenges due to the vast array of objects and the need for on-the-fly semantic inference and personalization.
    \item \emph{Compact Furniture Arrangement in a Bedroom (Section~\ref{sssec:bedroom}):} We address the complex task of arranging furniture in a bedroom where constraints from functionality and instructions are intricately intertwined without a clear hierarchical structure. This scenario tests the adaptability and robustness of our framework in handling tasks with convoluted and entangled constraints.
\end{itemize}

These new task families demonstrate the generality of our proposed compositional framework. They present unique and significant challenges not addressed in the original tabletop environment, such as handling large and diverse object sets, intricate functional constraints, and the absence of clear hierarchical structures.

By introducing a new program induction pipeline with structured natural language input and expanding the experimental evaluations to more complex and diverse task families, this manuscript offers substantial extensions over the conference paper.
\section{Spatial Relation Library}
\label{apped:abstract_rule}
Our spatial relation library consists of the following 48 relations.
We use these relations as the intermediate representation for generating the abstract object arrangements using the LLM and grounding these abstract spatial relations using the compositional diffusion models. In the following section, we describe the geometric concepts of these relations and the implementation of their classifiers $h_R$.

\begin{longtable}[ht]{>{\raggedright\arraybackslash}p{3.5cm}>{\raggedright\arraybackslash}p{2cm}>{\raggedright\arraybackslash}p{4.5cm}>{\raggedright\arraybackslash}p{4.5cm}}
\toprule
\textbf{Relation} & \textbf{Arity} & \textbf{Description} & \textbf{Implementation of $h_R$} \\
\midrule
central-column(Obj\_A) & Unary & The centroid of obj\_A is positioned within the central column of the container, which runs vertically through the container's midpoint, spanning half the container's width. & Check if centroid\_A is within the central column bounds. \\
\midrule
central-row(Obj\_A) & Unary & The centroid of obj\_A is positioned within the central row of the container, which runs horizontally through the container's midpoint, spanning half the container's length. & Check if centroid\_A is within the central row bounds. \\
\midrule
at-centered(Obj\_A) & Unary & The centroid of Obj\_A coincides with center of the container. & Check if the distance between the centroid\_A and the container's center is below a threshold. \\
\midrule
left-half(Obj\_A) & Unary & Obj\_A entirely resides within the left half of the container, occupying the left portion as divided vertically along the container's center axis. & Check if the rightmost point of the bounding box is on the left of the container's vertical centerline. \\
\midrule
right-half(Obj\_A) & Unary & Obj\_A entirely resides within the left half of the container, occupying the right portion as divided vertically along the container's center axis. & Check if the leftmost point of the bounding box is on the right of the container's vertical centerline.  \\
\midrule
front-half(Obj\_A) & Unary & Obj\_A entirely resides within the front half of the container, occupying the front portion as divided horizontally along the container's center axis. & Check if the highest point of the bounding box is below the container's horizontal centerline. \\
\midrule
back-half(Obj\_A) & Unary & Obj\_A entirely resides within the back half of the container, occupying the back portion as divided horizontally along the container's center axis. & Check if the lowest point of the bounding box is above the container's horizontal centerline. \\
\midrule
near-front-edge(Obj\_A) & Unary & Obj\_A is positioned near the container's front edge, which appears at the bottom in the top-down camera view and closest to the camera in the front view. & The shortest distance from any point on the bounding box to the front edge falls below a specified threshold. \\
\midrule
near-back-edge(Obj\_A) & Unary &  Obj\_A is near the back edge of the container (i.e. the edge opposite the front edge).  & The shortest distance from any point on the bounding box to the back edge falls below a specified threshold.\\
\midrule
near-left-edge(Obj\_A) & Unary & Obj\_A is near the left edge of the container, which is on the left in both top-down and front views. & The shortest distance from any point on the bounding box to the left edge falls below a specified threshold. \\
\midrule
near-right-edge(Obj\_A) & Unary & Obj\_A is near the right edge of the container, which is on the right in both top-down and front views. & The shortest distance from any point on the bounding box to the right edge falls below a specified threshold. \\
\midrule
facing-front(Obj\_A) & Unary & Obj\_A is oriented to face the front of the container, aligned with the container's front edge. & Check if the orientation of Obj\_A's bounding box is 0 radians (facing forward). \\
\midrule
facing-back(Obj\_A) & Unary & Obj\_A is oriented to face the back of the container, opposite to the container's front edge. & Check if the orientation of Obj\_A's bounding box is $\pi$ radians (facing backward). \\
\midrule
against-right-wall(Obj\_A) & Unary & Obj\_A is positioned against the right wall of the container, facing left toward the center from the right wall. & Verify that the rightmost point of Obj\_A's bounding box touches the container's right boundary, and the orientation is $\frac{\pi}{2}$ radians (facing left). \\
\midrule
against-left-wall(Obj\_A) & Unary & Obj\_A is positioned against the left wall of the container, facing right toward the center from the left wall. & Verify that the leftmost point of Obj\_A's bounding box touches the container's left boundary, and the orientation is $-\frac{\pi}{2}$ radians (facing right). \\
\midrule
against-front-wall(Obj\_A) & Unary & Obj\_A is positioned against the front wall of the container, facing back toward the center from the front wall. & Verify that the frontmost point of Obj\_A's bounding box touches the container's front boundary, and the orientation is $\pi$ radians (facing back). \\
\midrule
against-back-wall(Obj\_A) & Unary & Obj\_A is positioned against the back wall of the container, facing front toward the center from the back wall. & Verify that the backmost point of Obj\_A's bounding box touches the container's back boundary, and the orientation is $0$ radians (facing front). \\
\midrule
right-of-wall(Obj\_A) & Unary & Obj\_A is on the right side of the wall it is placed against, viewed from the room's center facing that wall. & Determine the wall against which Obj\_A is placed, then verify that Obj\_A's position along that wall is within the right half when facing the wall from the room's center. \\
\midrule
left-of-wall(Obj\_A) & Unary & Obj\_A is on the left side of the wall it is placed against, viewed from the room's center facing that wall. & Determine the wall against which Obj\_A is placed, then verify that Obj\_A's position along that wall is within the left half when facing the wall from the room's center. \\
\midrule
center-of-wall(Obj\_A) & Unary & Obj\_A is at the center of the wall it is placed against, viewed from the room's center facing that wall. & Determine the wall against which Obj\_A is placed, then verify that Obj\_A's position along that wall falls within the central third when facing the wall from the room's center. \\
\midrule
at-front-left-corner(Obj\_A) & Unary & Obj\_A is placed at the front-left corner of the container, facing the back wall. & Verify that Obj\_A's position is within a defined threshold of the front-left corner of the container, and the orientation is $\pi$ radians (facing the back wall). \\
\midrule
at-front-right-corner(Obj\_A) & Unary & Obj\_A is placed at the front-right corner of the container, facing the back wall. & Verify that Obj\_A's position is within a defined threshold of the front-right corner of the container, and the orientation is $\pi$ radians (facing the back wall). \\
\midrule
at-back-left-corner(Obj\_A) & Unary & Obj\_A is placed at the back-left corner of the container, facing the front wall. & Verify that Obj\_A's position is within a defined threshold of the back-left corner of the container, and the orientation is $0$ radians (facing the front wall). \\
\midrule
at-back-right-corner(Obj\_A) & Unary & Obj\_A is placed at the back-right corner of the container, facing the front wall. & Verify that Obj\_A's position is within a defined threshold of the back-right corner of the container, and the orientation is $0$ radians (facing the front wall). \\
\midrule
horizontally-aligned-bottom(Obj\_A, Obj\_B) & Binary & Obj\_A and Obj\_B are aligned horizontally at their bases. & Check if $|\textit{BB}_A.\textit{bottom} - \textit{BB}_B.\textit{bottom}|< \textit{threshold}$ and  $|\textit{BB}_A.\theta - \textit{BB}_B.\theta|< \textit{threshold}$ \\
\midrule
horizontally-aligned-centroid(Obj\_A, Obj\_B) & Binary & Obj\_A and Obj\_B are aligned horizontally at their centroids. & Check if $|\textit{centroid}_A.x - \textit{centroid}_B.x| < \textit{threshold}$ and  $|\textit{BB}_A.\theta - \textit{BB}_B.\theta| < \textit{threshold}$ \\
\midrule
vertically-aligned-centroid(Obj\_A, Obj\_B) & Binary & Obj\_A and Obj\_B are aligned vertically at their centroids. & Check if $|\textit{centroid}_A.y - \textit{centroid}_B.y| < \textit{threshold}$ and  $|\textit{BB}_A.\theta - \textit{BB}_B.\theta| < \textit{threshold}$ \\
\midrule
left-of(Obj\_A, Obj\_B) & Binary & Obj\_A is immediately to the left side of Obj\_B viewed from the canonical camera frame. & Check if 1) $|\textit{BB}_A.\textit{right} - \textit{BB}_B.\textit{left}| < \textit{threshold}$, and 2) $\textit{BB}_A.y$ and $\textit{BB}_B.y$ overlap substantially.\\
\midrule
right-of(Obj\_A, Obj\_B) & Binary & Obj\_A is immediately to the right side of Obj\_B viewed from the canonical camera frame. & Check if 1) $|\textit{BB}_A.\textit{left} - \textit{BB}_B.\textit{right}| < \textit{threshold}$, and 2) $\textit{BB}_A.y$ and $\textit{BB}_B.y$ overlap substantially.\\
\midrule
in-front-of(Obj\_A, Obj\_B) & Binary & Obj\_A is immediately in front of Obj\_B, both facing the same wall. Obj\_A's orientation aligns with Obj\_B's. & Check if: 1) $|\textit{BB}_A.\textit{back} - \textit{BB}_B.\textit{front}| < \textit{threshold}$; 2) $\textit{BB}_A.x$ and $\textit{BB}_B.x$ overlap substantially; and 3) $|\textit{orientation}_A - \textit{orientation}_B| < \delta$ (orientations are aligned). \\
\midrule
on-top-of(Obj\_A, Obj\_B) & Binary & Obj\_A is placed on top of Obj\_B. & Check if 1) $\textit{BB}_A$ completely overlaps with $\textit{BB}_B$, and 2) the area of $\textit{BB}_B$ is equal or larger than $\textit{BB}_A$.\\
\midrule
centered(Obj\_A, Obj\_B) & Binary & Center of Obj\_A coincides with center of Obj\_B. & Check if $|\textit{centroid}_A.y - \textit{centroid}_B.y| < \textit{threshold}$. \\
\midrule
left-touching(Obj\_A, Obj\_B) & Binary & Obj\_A is directly to the left of Obj\_B, touching it, both facing the same wall from the room's center. & Confirm orientations are aligned within \(\delta\); determine axis based on facing wall; check Obj\_A is left of Obj\_B along that axis; verify adjacent sides are within touching threshold. \\
\midrule
right-touching(Obj\_A, Obj\_B) & Binary & Obj\_A is directly to the right of Obj\_B, touching it, both facing the same wall from the room's center. & Confirm orientations are aligned within \(\delta\); determine axis based on facing wall; check Obj\_A is right of Obj\_B along that axis; verify adjacent sides are within touching threshold. \\
\midrule
under-window(Obj\_A, Obj\_B) & Binary & Obj\_A is placed directly under the window (Obj\_B), facing the window. & Confirm Obj\_A is horizontally aligned with Obj\_B within a threshold; positioned directly below Obj\_B; orientation matches window's wall (facing the window). \\
\midrule
horizontal-symmetry-about-axis-obj
(Axis\_obj, Obj\_A, Obj\_B)  & Ternary & Obj\_A and Obj\_B are symmetrical to each other with respect to the horizontal axis of the Axis\_obj, meaning they are mirror images along that axis. & Check mirrored centroids along Axis\_obj's horizontal line \\
\midrule
vertical-symmetry-about-axis-obj
(Axis\_obj, Obj\_A, Obj\_B) & Ternary & Obj\_A and Obj\_B are symmetrical to each other with respect to a vertical axis of the Axis\_obj, meaning they are mirror images along that axis. & Check mirrored centroids along Axis\_obj's vertical line \\
\midrule
aligned-in-horizontal-line-bottom(Obj\_1, ..., Obj\_N) & Variable (N) & Objects aligned horizontally with equal spacing at their bases. & Check for pairwise $|\textit{BB}_i.\textit{bottom} - \textit{BB}_j.\textit{bottom}|< \textit{threshold}$ and  $|\textit{BB}_i.\theta - \textit{BB}_j.\theta|< \textit{threshold}$, and equal spacing horizontally between the adjacent objects. \\
\midrule
aligned-in-horizontal-line-centroid(Obj\_1, ..., Obj\_N) & Variable (N) & Objects aligned horizontally with equal spacing at their centroids. & Check for pairwise $|\textit{centroid}_i.\textit{x} - \textit{centroid}_j.\textit{x}|< \textit{threshold}$ and  $|\textit{BB}_i.\theta - \textit{BB}_j.\theta|< \textit{threshold}$, and equal spacing vertically between the adjacent objects. \\
\midrule
aligned-in-vertical-line-centroid(Obj\_1, ..., Obj\_N) & Variable (N) & Objects aligned vertically with equal spacing at their centroids. & Check for pairwise $|\textit{centroid}_i.\textit{y} - \textit{centroid}_j.\textit{y}|< \textit{threshold}$ and  $|\textit{BB}_i.\theta - \textit{BB}_j.\theta|< \textit{threshold}$, and equal spacing vertically between the adjacent objects. \\
\midrule
regular-grid(Obj\_1, ..., Obj\_N) & Variable (N) & Arranged in a grid or matrix pattern. & Verify if the centroids of objects in the same row or column are aligned and ensure the distances between rows and columns are constant.\\ 
\midrule
\midrule
contiguously-aligned(Obj\_1, ..., Obj\_N) & Variable (N) & Obj\_1 through Obj\_N are aligned horizontally in a straight line at their bases, each directly contacting its neighbors. &
- Confirm orientations are similar: For all \(i, j\), \(|\theta_i - \theta_j| < \textit{threshold}\).
- Check bases are aligned: For all \(i, j\), \(|y^{\text{bottom}}_i - y^{\text{bottom}}_j| < \textit{threshold}\).
- Verify objects are ordered along the horizontal axis:
  - For \(i = 1\) to \(N-1\), confirm \(x^{\text{right}}_i \approx x^{\text{left}}_{i+1}\) within a touching threshold.
\\
\midrule
height-sorted-ascending(Obj\_1, ..., Obj\_N) & Variable (N) & Objects are aligned from left to right at their bases, sorted in ascending order of height. &
- Confirm bases are aligned: For all \(i, j\), \(|y^{\text{bottom}}_i - y^{\text{bottom}}_j| < \textit{threshold}\).
- Verify left-to-right order: For \(i = 1\) to \(N-1\), ensure \(x_i < x_{i+1}\).
- Check heights are ascending: For \(i = 1\) to \(N-1\), confirm \(\text{height}_i < \text{height}_{i+1}\).
\\
\midrule
height-sorted-descending(Obj\_1, ..., Obj\_N) & Variable (N) & Objects are aligned from left to right at their bases, sorted in descending order of height. &
- Confirm bases are aligned: \(|y^{\text{bottom}}_i - y^{\text{bottom}}_j| < \textit{threshold}\) for all \(i, j\).
- Verify left-to-right order: For \(i = 1\) to \(N-1\), ensure \(x_i < x_{i+1}\).
- Check heights are descending: For \(i = 1\) to \(N-1\), confirm \(\text{height}_i > \text{height}_{i+1}\).
\\
\midrule
width-sorted-ascending(Obj\_1, ..., Obj\_N) & Variable (N) & Objects are aligned from left to right at their bases, sorted in ascending order of width. &
- Confirm bases are aligned: \(|y^{\text{bottom}}_i - y^{\text{bottom}}_j| < \textit{threshold}\) for all \(i, j\).
- Verify left-to-right order: For \(i = 1\) to \(N-1\), ensure \(x_i < x_{i+1}\).
- Check widths are ascending: For \(i = 1\) to \(N-1\), confirm \(\text{width}_i < \text{width}_{i+1}\).
\\
\midrule
width-sorted-descending(Obj\_1, ..., Obj\_N) & Variable (N) & Objects are aligned from left to right at their bases, sorted in descending order of width. &
- Confirm bases are aligned: \(|y^{\text{bottom}}_i - y^{\text{bottom}}_j| < \textit{threshold}\) for all \(i, j\).
- Verify left-to-right order: For \(i = 1\) to \(N-1\), ensure \(x_i < x_{i+1}\).
- Check widths are descending: For \(i = 1\) to \(N-1\), confirm \(\text{width}_i > \text{width}_{i+1}\).
\\
\bottomrule
\label{tab:relationship_dictionary}
\end{longtable}

\section{Reverse Sampling and ULA Sampling}
\label{append:reverse_ula}

In this section, we illustrate how a step of reverse sampling at a fixed noise level in a diffusion model is equivalent to ULA sampling at the same fixed noise level. The reverse sampling step on an input $x_t$  at a fixed noise level at timestep $t$ is given by by a Gaussian with a mean 
\begin{equation*}
        \mu_\theta(x_t, t) = x_t - \frac{\beta_t}{\sqrt{1 - \bar{\alpha}_t}}\epsilon_\theta(x_t, t).
\end{equation*}
with the variance of $\beta_t$ (using the variance small noise schedule in~\cite{ho2020denoising}). This corresponds to a sampling update,
\begin{equation*}
        x_{t+1} = x_t - \frac{\beta_t}{\sqrt{1 - \bar{\alpha}_t}}\epsilon_\theta(x_t, t) + \beta_t \xi, \quad \xi \sim \mathcal{N}(0, 1).
\end{equation*}

Note that the expression $\frac{\epsilon_\theta(x_t, t)}{\sqrt{1 - \bar{\alpha}_t}}$ corresponds to the score $\nabla_x p_t(x)$, through the denoising score matching objective~\citep{vincent2011connection}, where $p_t(x)$ corresponds to the data distribution perturbed with $t$ steps of noise. The reverse sampling step can be equivalently written as
\begin{equation}
        \label{eqn:reverse2}
        x_{t+1} = x_t - \beta_t \nabla_x p_t(x) + \beta_t \xi, \quad \xi \sim \mathcal{N}(0, 1).
\end{equation}

The ULA sampler draws a MCMC sample from the probability distribution $p_t(x)$ using the expression
\begin{equation}
        \label{eqn:ula}
        x_{t+1} = x_t - \eta \nabla_x p_t(x) + \sqrt{2} \eta \xi, \quad \xi \sim \mathcal{N}(0, 1),
\end{equation}
where $\eta$ is the step size of sampling.

By substituting $\eta=\beta_t$ in the ULA sampler, the sampler becomes
\begin{equation}
        \label{eqn:ula_beta}
        x_{t+1} = x_t - \beta_t \nabla_x p_t(x) + \sqrt{2} \beta_t \xi, \quad \xi \sim \mathcal{N}(0, 1).
\end{equation}
Note that the sampling expression for ULA sampling in Eqn~\ref{eqn:ula_beta} is very similar to the sampling procedure in the standard diffusion reverse process in Eqn~\ref{eqn:reverse2}, where there is a factor of $\sqrt{2}$ scaling in the ULA sampling procedures. Thus, we can implement ULA sampling equivalently by running the standard reverse process, but by scaling the variance of the small noise schedule by a factor of 2 (essentially multiplying the noise added in each timestep by a factor of $\sqrt{2}$). Alternatively, if we directly use the variance of the small noise schedule, we are running ULA on a tempered variant of $p_t(x)$ with temperature $\frac{1}{\sqrt{2}}$ (corresponding to less stochastic samples).

\section{Diffusion Model Implementation Details}
\label{append:diffusion}

Each individual diffusion constraint solver is trained to predict the noise required for reconstructing object poses, thereby satisfying a specified spatial relation. At its core, the model operates by taking normalized shapes and poses of objects, represented as 2D bounding boxes, and outputs the predicted noise for reconstructing these poses. The normalization process uses the dimensions of a reference table region to ensure consistent scaling.

\noindent
\textbf{Encoder Implementation}: The model features three primary encoders - a shape encoder, a pose encoder, and a time encoder, each tailored for its specific type of input. The shape and pose encoders leverage a two-layer neural network with SiLU activations, scaling the input dimensions to a predefined hidden dimension size 256. Specifically, the shape encoder transforms object geometry information, while the pose encoder processes the 2D bounding boxes' positions and sizes:
\begin{itemize}
    \item Shape Encoder: Processes geometry inputs using a sequential neural network architecture, first mapping from input dimensions to half the hidden dimension, followed by a mapping to the full hidden dimension.
    \item Pose Encoder and Pose Decoder: The pose encoder follows a structure similar to the shape encoder, focusing on the pose information of objects. Correspondingly, the pose decoder translates the hidden representation back to the pose dimensions, facilitating the reconstruction of object poses.
    \item Time Encoder: Comprises a sinusoidal positional embedding followed by a linear scaling to four times the hidden dimension, processed with a Mish activation, and then downscaled back to the hidden dimension.
\end{itemize}

\noindent
\textbf{Backbone Architecture}: Based on the arity of spatial relations, our model employs two different backbones. A Multi-Layer Perceptron (MLP) serves as the backbone for relations with fixed arity, utilizing a sequential architecture to process concatenated shape and pose features. For relations of variable arity, a transformer backbone is employed, capable of handling varying sequence lengths and providing a flexible means to process an arbitrary number of objects:

\begin{itemize}
    \item MLP Backbone: Utilizes a linear layer followed by a SiLU activation, designed to intake normalized and concatenated object features for fixed arity relations.
    \item Transformer Backbone: Processes variable-arity relations by dynamically adjusting to the number of objects. The transformer is configured with a maximum capacity of 16 objects, determined by its sequence length limit of 16. It includes preprocessing steps for positional encodings, attention mask preparation, and padding to accommodate sequences of different lengths. The transformer's output is then post-processed and decoded into the pose dimensions.
\end{itemize}

\noindent
\textbf{Training and Inference}: The entire model, including encoders, backbone, and the pose decoder, is trained end-to-end with an L2 loss, aiming to minimize the difference between predicted and actual noise values used for pose reconstruction. During inference, our model utilizes a cosine beta schedule over 1500 diffusion steps to iteratively refine the noise predictions. For complex scenes with multiple spatial relations, the model aggregates and averages (or weights) the noise features from relevant relations before final decoding, enhancing pose prediction accuracy.

\section{Prompts}
\label{append:prompt}

\subsection{Natural Language Task Specification}
\label{append:NL_task_specification}

We provide the full structured natural language task description for arranging the tablewares on the dining tables below.

\begin{lstlisting}[caption={Natural Language Task Specification for Dining Tables}]
# Program Synthesis for Dining Table

## Description

This guide details how to arrange tableware and cutlery on a dining table according to specified instructions and preferences. It consists of two main components: 

1) **Domain Specification**: Defines the customized storage structure for organizing information and describes common arrangement rules and patterns for recurring sub-structures.
2) **Procedural Description**: Provides a step-by-step process to propose the object arrangement layout.

## Domain Specification

### Step 1: **Create Data Structure for Information Handling**

To ensure consistency in our arrangement, we create data structures to help organize information as we procedurally generate abstract object relations. The object arrangement layout involves two main components:
1. Tableware arrangement for individual diners.
2. Shared tableware arrangement.

Based on this structure, we develop two data structures to handle relevant information and abstract relations for each component.

**Component 1: Individual Diner Tableware Arrangement**

We create a Python dictionary named `diner_dictionary`, indexed by diner ID. Each entry stores the details of an individual diner's tableware arrangement using a `Diner` class. The `Diner` class includes the following attributes:
- `ID`: A unique identification number.
- `special_requirement`: Specific needs or preferences (default is none).
- `individual_tableware_arrangement_style`: A semantic name indicating the dining style, e.g., "Chinese dining setup" or "Western dining setup".
- `seating_position`: A tuple describing the diner's seating position at the table. The first element indicates the table side ("near_front_edge" or "near_back_edge"), and the second element specifies the relative location on that side ("left_half," "central_column," or "right_half").
- `object_list`: A list of object names assigned to the diner.
- `active_relations`: A list of abstract object relations describing the layout for the diner.

**Component 2: Shared Tableware Arrangement**

We create a Python dictionary named `sharing_objects`, indexed by the semantic names of sharing object groups (e.g., "sharing dishes" or "sharing seasonings"). Each entry uses a `SharingObjectGroup` class to organize the information, which includes the following attributes:
- `object_list`: A list of object names belonging to the sharing object group.
- `abstract_relations`: A list of proposed active abstract relations for the shared objects.

By organizing our data structures in this way, we can efficiently manage both individual and shared tableware arrangements, ensuring a consistent and coherent layout.


### Step 2: **Establish General Setup Guidelines**

Different dining styles (e.g., Western vs. Chinese) and occasions (e.g., setups for kids or left-handed diners) require specific tableware arrangements. We follow common guidelines to arrange these setups based on available tableware and provided instructions.

There are recurring basic arrangement units for specific setups, including individual dining-style arrangements, sharing dishes, and sharing utensils. We describe the common rules for these arrangements below, allowing us to select, combine, and adapt them as needed.

#### Step 2.1. **Arranging Western-style Tableware for an individual Diner**

Western-style dinner setups typically include a serving plate, napkin, fork, knife, spoon, and glass. Let's discuss a standard setup for a right-handed diner, whose `seating_position` is at the "central column" "near the front edge" of the table.

- **Serving Plate**: Positioned according to the diner's `seating position`: ("central_column", "serving_plate") and ("near_front_edge", "serving_plate").
- **Napkin**: Placed to the left of the serving plate ("left_of", "napkin", "serving_plate").
- **Fork**: Placed on top of the napkin ("on_top_of", "fork", "napkin").
- **Knife and Spoon**: Positioned to the right of the serving plate, with the knife first and then the spoon ("right_of", "knife", "serving_plate") and ("right_of", "spoon", "knife").
- **Glass**: Positioned to the right of the spoon ("right_of", "glass", "spoon").

Adjustments may be needed based on:
1. Available tableware.
2. Diner's seating position.
3. Diner's personal preferences.

Given a list of object names, the diner's `seating_position` as a tuple, and a string indicating their `personal_preference`, here is the adjustment process:

1. Query the language model to associate the object names with standard setup items.
2. Collect unassigned object names in a separate list.
3. Propose the placement for the object designated as the "serving_plate" based on the `seating_position`.
4. For matched objects, follow the standard arrangement rules.
5. For unassigned objects, query the language model with:
   - Dining style.
   - Current object list.
   - Current arrangement.
   This will help determine their abstract relations with the rest of the setup.
6. Adjust the arrangement based on `personal preference` and the table edge in the `seating_position`. If the diner is left-handed xor facing the back edge, reverse the "left_of" and "right_of" relations.


#### Step 2.2. **Arranging Chinese-style Tablewares for an individual diner.**

Typical items used in Chinese-style table setups include a small plate, rice bowl, chopsticks, spoon, and a glass. Let's discuss a common setup for a right-handed diner whose `seating_position` is at the "central column" "near the front edge" of the table.

- **Small Plate**: Positioned according to the diner's `seating_position`: such as "central_column" and "near_front_edge".
- **Rice Bowl**: Placed on top of the small plate.
- **Chopsticks, Spoon, and Glass**: Positioned sequentially from left to right on the right-hand side of the small plate.

Adjustments may be needed based on:
1. Available tableware.
2. Diner's seating position.
3. Diner's personal preferences.

Given a list of object names, the diner's `seating_position` as a tuple, and a string indicating their `personal_preference`, here is the adjustment process:

1. Query the language model to associate the object names with standard setup items.
2. Collect unassigned object names in a separate list.
3. Propose the placement for the object designated as the "small_plate" based on the `seating_position`.
4. For matched objects, follow the standard arrangement rules.
5. For unassigned objects, query the language model with:
   - Dining style.
   - Current object list.
   - Current arrangement.
   This will help determine their abstract relations with the rest of the setup.
6. Adjust the arrangement based on `personal preference` and the table edge in the `seating_position`. If the diner is left-handed xor facing the back edge, reverse the "left_of" and "right_of" relations.


#### Step 2.3. **Arranging Ramen-style Tablewares for an individual diner.**

A standard Japanese Ramen table setup commonly includes a ramen bowl, chopsticks, spoon, and glass. Let's discuss a basic arrangement for a right-handed diner, whose `seating_position` is at the "central column" "near the front edge" of the table.

- **Ramen Bowl**: Positioned according to the diner's `seating_position`, such as "central_column" and "near_front_edge".
- **Chopsticks**: Placed to the right of the ramen bowl.
- **Spoon**: Positioned to the right of the chopsticks.
- **Glass**: Positioned to the right of the spoon.

Adjustments may be needed based on:
1. Available tableware.
2. Diner's seating position.
3. Diner's personal preferences.

Given a list of object names, the diner's `seating_position` as a tuple, and a string indicating their `personal_preference`, here is the adjustment process:
1. Query the language model to associate the object names with standard setup items.
2. Collect unassigned object names in a separate list.
3. Propose the placement for the object designated as the "ramen_bowl" based on the `seating_position`.
4. For matched objects, follow the standard arrangement rules.
5. For unassigned objects, query the language model with:
   - Dining style.
   - Current object list.
   - Current arrangement.
   This will help determine their abstract relations with the rest of the setup.
6. Adjust the arrangement based on `personal preference` and the table edge in the `seating_position`. If the diner is left-handed xor facing the back edge, reverse the "left_of" and "right_of" relations.


#### Step 2.4. **Arranging other-dining style for an individual diner.**

Given a list of object names, a tuple indicating the diner's `seating_position`, a string indicating the diner's `personal_preference`, and a string specifying the `individual_tableware_arrangement_style`, we will propose a tableware arrangement for a single diner by following these steps:

1. **Identify the Main Dish Container**: Determine the "main_dish_container" essential for the dining style. For example, the "main_serving_plate" is the "main_dish_container" for a western dining setup. Query the language model using the `individual_tableware_arrangement_style` and the object name list to identify the relevant object name.
2. **Determine Placement for the Main Dish**: Position the "main_dish_container" based on the diner's `seating_position`. For example, if the `seating_position` is ("central_column", "near_front_edge"), the proposed placements are ("central_column", "main_dish_container") and ("near_front_edge", "main_dish_container").
3. **Arrange Related Objects**: Query the language model to propose the abstract object relations for other objects relative to the "main_dish_container" and among themselves, according to the `individual_tableware_arrangement_style` and the object name list. The language model should consider the dining conventions for a right-handed diner. 
4. **Adjust for Preferences**: Modify the arrangement based on the diner's `personal_preference` and exact `seating_position`. For a left-handed diner or one facing the back edge, reverse the "left_of" and "right_of" relations.

#### Step 2.5. **Arranging Shared Plates.**

Given a list of object names for shared dishes at a meal, the arrangement should be neat and centrally located for easy access by all diners. The arrangement varies based on the number of objects:

1. **Neat Arrangement**:
   - **One Object**: Simply center the object.
   - **Two Objects**: Arrange them symmetrically along the table's vertical axis.
   - **Three or Four Objects**: Arrange them in a single horizontal line.
   - **More Than Four Objects**: Arrange them in a regular grid.

2. **Central Arrangement**:
   - **One Object**: Place it at the exact center of the table.
   - **Multiple Objects**: Place each of them on central row of the table.


#### Step 2.6. **Arranging Shared Side Items**

Given a list of object names for shared side items, such as seasonings or bottles of water, follow these steps to ensure easy access and a neat arrangement:

1. **Vertical Alignment**: Align the side items vertically.
2. **Central Row Placement**: Place each item along the central row of the table.
3. **Edge Positioning**: Decide on either the left or right edge of the table, then place each item near that edge.


## Procedural Description

### Step 1: **Extract the Relevant Information from the Instructions**

#### Step 1.1. **Figuring Out the Number of Diners**
We start off by finding out how many people are dining from the instruction. This is important because it helps us know how many seats and table sets we will need. 
- How to do this: query the language model with the instruction to infer the number of diners. The language model should return an integer that indicates the number of diners. Based on the number of diners, create a diners dictionary to store the information for all the diners.


#### Step 1.2. **Determining Where Everyone Wants to Sit**
To determine where everyone wants to sit at the table, follow these steps. Our table has two main sides (front and back edge) and three seating positions on each side (left, center, and right). We will update the `seating_position` for each diner. The `seating_position` is a tuple where the first element indicates either "near_front_edge" or "near_back_edge" of the table, and the second element indicates whether the seat is on the "left_half", "central_column", or "right_half" of the assigned table edge.

Given `instruction`, `diner_dictionary`, you need to make sure that each diner in the dictionary has their `seating_position` assigned as a tuple. 
How to do this:
1. **Determine Table Side**:
    - Query the language model to infer from the instruction if diners want to sit on the same side or opposite sides of the table (default is opposite sides).
      - If opposite sides: Evenly distribute diners between the front_edge and back_edge.
      - If same side: Allocate all diners to the front_edge.

2. **Determine Preferred Side**:
    - Query the language model to infer if each diner has a preferred seating area.
      - If preferences are indicated: Assign seats according to each diner's preference.
      - If no preferences: Evenly distribute the diners on each edge by assigning left_half, central_column, or right_half positions.


#### Step 1.3. **Identifying the Diners with Special Requirements**
Lastly, we need to check if any diner has special requirements, such as "needing a baby seat" or being "left-handed". We will update the `personal_preference` field for each diner.

Given `instruction`, `diner_dictionary`, make sure that each diner in the dictionary has their `personal_preference` field updated with the inferred special requirements.


How to do this: 

1. **Query Special Requirements**:
    - Query the language model using the instruction and the number_of_diners to infer a list of strings indicating special requirements for each diner. The default is `None`.

2. **Assign Requirements**:
    - Assign each element of the inferred list to the `personal_preference` field for each diner.


### Step 2: Identify the Relevant Sub-arrangement

#### Step 2.1 **Identify the individual Dining Style**
Next we need to identify the relevant dining style for arranging the tablewares for each diner from the instruction, such as "western dining style" or "Chinese dining style". We update the `individual_tableware_arrangement_style` field for each diner.

How to do this:
- we query the language model to infer the style of dining based on the instruction as a single string,
- iterate through the diners, assign the inferred dining_style to the `individual_tableware_arrangement_style` for each diner.

Given instruction, diner_dictionary, assign the `individual_tableware_arrangement_style` for each diner in the diner_dictionary. 

#### Step 2.2. **Identify the categories for Shared Objects**

Next, we will identify the categories for the shared objects among the diners. There are two possible categories: "shared_dishes" for the main dishes and "shared_side_objects" for items like condiments or drinks.

How to do this: 

1. **Query the Categories**:
    - Using the instruction and the object name list, query the language model to determine the categories of the shared objects. There are two possible categories: "shared_dishes" for the main dishes and "shared_side_objects" for items like condiments or drinks.
    - Return a list of the identified category names.

2. **Initialize the `sharing_objects` Dictionary**:
    - Create a dictionary with the identified category names as keys.
    - Assign `SharingObjectGroup` as the placeholder value for each category.

Given instruction, object_name_list, output the `sharing_objects` dictionary. The keys of the dictionary are the identified category names, while the values are `SharingObjectGroup` for now. 


### Step 3: Assigning the Objects Into the Identified Groups

#### Step 3.1. **Assigning Personal Tableware to Diners**

Given the full `object_name_list` and the `diners` dictionary, we will assign a list of personal tableware items to the `object_list` for each diner.


How to do this:

1. **Identify Individual Usage Objects**:
    - Query the language model with the `object_name_list`, number of diners, and `individual_tableware_arrangement_style` to select a list of object names intended for individual use.
    - Ensure the selected list includes main serving tableware and the required utensils for each diner.
    - Save these object names in a list called `individual_objects`.

2. **Assign Objects to Diners**:
    - Iterate through each diner in the `diners` dictionary.
    - For each diner, query the language model to select a subset of object names based on the diner's `personal_preference` and `individual_tableware_arrangement_style` from the `individual_objects` list.
    - Assign this subset list to the `object_list` for each diner.

#### Step 3.2. **Assigning the Shared Items**

Given the full `object_name_list`, the previously identified `individual_objects`, and the `sharing_objects` dictionary, we will assign a list of shared object names to each shared object category.

How to do this: 
1. **Identify Shared Objects**:
    - Create a list called `shared_objects_list` by selecting all objects in `object_name_list` that are not in the `individual_objects` list.

2. **Assign Objects to Categories**:
    - Iterate through the keys of the `sharing_objects` dictionary (each key representing a category).
    - For each category, query the language model to select a subset of object names based on the `category_name` (the key) and the `shared_objects_list`.
    - Assign this subset of object names to the `object_list` of `ShareObjectGroup` of the corresponding category name.


### Step 4: **Arranging the Objects within Each Group**

#### Step 4.1 **Arranging the Personal Tableware For Each Diner**

We propose the object arrangement for each diner based on their `individual_tableware_arrangement_style`, `personal_preference`, `seating_position`, and `object_list`.

Given `diners` dictionary, we update it with the `active_relations` for each diner assigned a list of abstract relations.

How to do this:

1. **Iterate Through Diners**:
    - For each diner in the `diners` dictionary, based on its `individual_tableware_arrangement_style`, call the relevant arrangement function with the necessary arguments. We have the following arrangement functions for individual diner:
        - `western_dining_arrangement(object_list, seating_position, personal_preference)`
        - `chinese_dining_arrangement(object_list, seating_position, personal_preference)`
        - `ramen_dining_arrangement(object_list, seating_position, personal_preference)`
        - `other_dining_arrangement(individual_tableware_arrangement_style, object_list, seating_position, personal_preference)`

2. **Assign Abstract relations**:
    - Each arrangement function returns a list of abstract object relations.
    - Assign this list to the `active_relations` for each diner.


#### Step 4.2 **Arranging the Sharing Objects**

We propose the arrangements for groups of shared objects based on their categories.

Given `sharing_objects` dictionary, we update it with the `active_relations` for each `SharedObjectGroup` assigned a list of abstract relations.

How to do this:
1. we iterate through the `sharing_objects` dictionary, and call the relevant function for arranging the objects based on the `category_name` (the key of the dictionary) and the associated `object_list`. We have the following arrangement functions for the shared objects:
    - `shared_dished_arrangement(object_list)`
    - `shared_side_objects_arrangement(object_list)`
2. each arrangement function returns a list of abstract object relations. Assign this this to the `active_relation` to each `SharedObjectGroup` indexed by the category name.
\end{lstlisting}

\subsection{Prompts for \textit{SketchGenerator}} \label{append:prompts_for_sketch_generator}

The natural language task description comprises two parts: 1) domain specification, and 2) procedure description.

The domain specification includes: 
1) Intermediate information storage structures and partial abstract relation assignments.
2) Common arrangement rules and patterns for recurring sub-structures.

The storage structures are analogous to Python data structures, and these common arrangement patterns correspond to helper functions that generate proposed abstract relations for sub-structures based on input objects and additional information.

The procedural description outlines the step-by-step process for generating an abstract grounding graph for functional object arrangement. Each step corresponds to a specific function that implements the step as described, potentially creating or updating data structures with relevant intermediate information, and/or using helper functions to apply common patterns/rules to sub-structures.

We design two different prompts for these two components.

\subsubsection{Prompt for Domain Specification}

We first translate the domain specification from the natural language description to a Python code skeleton. This translation involves two components: 1) creating data structure, and 2) specifying the function definition for each common patterns.

\textbf{Creating Data Structure}

Design and implement an appropriate data structure in Python based on the provided natural language description for organizing the information.

\textbf{Specifying the Function Definition for each Common Patterns}
\begin{enumerate}
    \item A clear, semantically meaningful name.
    \item Well-defined arguments specifying types.
    \item Explicit return types.
    \item A docstring that encapsulates the purpose and the detailed mechanics of the function. 
    \item Copy all the natural language descriptions of the step/sub-step into the comment, do NOT try to summarize it.
\end{enumerate}
Do NOT write any implementation.

\subsubsection{Prompt for Procedural Description}

Next, translate the detailed procedural description for proposing tableware arrangements on a dining table into a Python code skeleton. You do NOT need to implement the functions; your role is to organize the information by translating the natural language description for each step and sub-step into Python function skeletons. You have the following two tasks: 1) specifying the function definition for each step and integrating the functions for the main function.

\textbf{Specify the Function Definition for each Step}

For each step and **sub-step** listed in the natural language instructions, define corresponding Python functions. Ensure that each function has:
\begin{enumerate}
    \item A clear, semantically meaningful name.
    \item Well-defined arguments specifying types.
    \item Explicit return types.
    \item A docstring that encapsulates the purpose and the detailed mechanics of the function. 
    \item Copy all the natural language descriptions of the step/sub-step into the comment, do NOT try to summarize it.
\end{enumerate}
Do NOT write any implementation.

\textbf{Integrating the Functions for the Main Function}

Using the previously defined functions, systematically implement the program logic for the main function. Declare variables with meaningful names to capture intermediate results which will then be passed as arguments to subsequent functions.

Please design the functions such that they can be chained together to process the input and produce the desired output based on the stepwise procedure explained in the instructions. Begin with the primary function template:

\begin{lstlisting}
def generate_object_arrangements_for_dining_table(instruction: str, object_list: List[str]) -> List[str]:
    """
    Generate a list of abstract relations among objects on a dining table based on given instructions.

    :param instruction: A string detailing the preference on the dining table arrangement.
    :param object_list: A list of strings, each a name of an object to be arranged.
    :return: A list of tuples of strings describing the relation among the arranged objects.
    """
    active_relation_list = []
    # Implementation goes here
    return active_relation_list
\end{lstlisting}

Ensure all function signatures within the main function are properly defined and that their integration adheres to the logical steps provided in the instructions.

\subsection{Prompts for \textit{Coder}}
\label{append:coder_prompts}

You are tasked with synthesizing Python code based on a fully-documented function signature and an accompanying docstring. 

Below are the function signature and docstring:

Program sketch goes to here.

Your task is to carefully interpret the provided function signature and docstring, then implement the function in Python by following the natural language description step-by-step. If the description specifies a fixed strategy, implement it directly using Python code. If the description requires on-the-fly inference or adaptation, use the \texttt{call\_LLM()} function to query GPT-4 and \texttt{parse\_LLM()} during your coding process. We have already imported \texttt{call\_LLM()} and \texttt{parse\_LLM()} for you.

When using \texttt{call\_LLM()}, structure your prompt in three parts:
\begin{enumerate}
    \item Question: Clearly state the query related to your implementation. 
    \item Examples or explanations: Provide examples or further explanation to clarify the query. You can extract this information from the natural language description.
    \item Output Format: Describe the format in which you want the response.
\end{enumerate}

\subsection{Prompt for \textit{Verifier}}
\label{append:verifier_prompt}

You are a verifier tasked with ensuring that a function's implementation matches specified input-output examples.

For this task, you are provided with:

\begin{enumerate}
    \item Function signature and the docstring: the input to the synthesizer.
    \item Function Implementation: The actual code of the function implemented by the \textit{Coder} LLM.
    \item 5 demonstrations: each demonstration is an input-output pair:
    \begin{itemize}
        \item Input: The data that will be fed into the function.
        \item Expected Output: The anticipated result based on the input.
    \end{itemize}
\end{enumerate}

Your role is to determine if the function's implementation correctly produces the expected output for the given input. When using \texttt{call\_LLM()}, use your commonsense reasoning to emulate the function's output. If the function passes the test, return "Pass". If it fails, return "Fail", and revise the function signature and docstring to inform the synthesizer about the specific code issues and how to correct them.
\section{Task Family Overview}
\label{append:task_family}

This work encompasses three distinct task families: dining tables, bookshelves, and bedroom layouts. Each family represents a specific type of functional object arrangement tasks, defined by its high-level purposes and characterized by unique organizational principles. Moreover, each family encompasses a group of objects typically found in its respective setting. This section outlines the intended function of organizing tabletop environments within each task family and enumerates the associated objects included in our datasets. 

\subsection{Dining Table}

\begin{figure*}[ht]
    \centering
    \includegraphics[width=\linewidth]{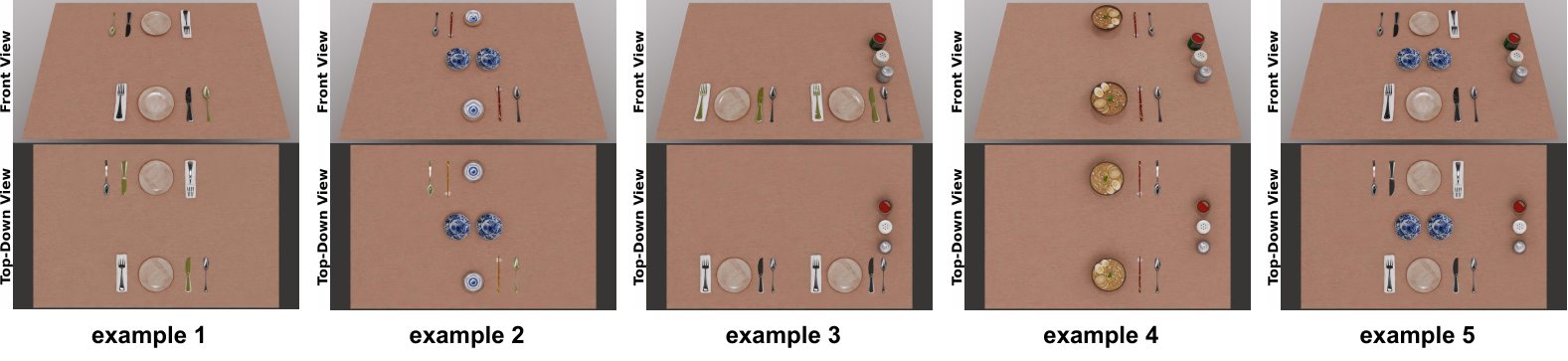}
    \caption{Five example FORM for Dining Table Arrangement.}
    \label{fig:dining_form}
\end{figure*}

\begin{itemize}
    \item \textbf{Purpose}: To prepare the dining table based on the specific dining style and the number of diners, accommodating any special requests (e.g., preferences for left-handed individuals). Utensils and tablewares are set according to conventional dining etiquette for the chosen dining style.
    \item \textbf{Objects}: serving plate, fork, napkin, spoon, knife, glass, ramen bowl, medium plate, chopsticks, small plate, rice bowl, baby bowl, baby plate, baby cup, baby spoon, baby fork, seasoning bottle.
    \item \textbf{Training \& Testing datasets}: Table \ref{tab:dining_table_training} and Figure \ref{fig:dining_form} contains the training examples, each consisting of user instructions, object lists, and FORM layouts. The testing dataset, which includes user instructions and object lists, is presented in Table \ref{tab:dining_table_testing}.
\end{itemize}

The task of arranging dining tables presents unique challenges due to the \textbf{large number of objects and relations} involved. For instance, setting up a table for four diners with diverse preferences can require coordinating up to 27 different tableware items. Minor changes in instructions, such as accommodating left-handed diners or arranging side-by-side seating, necessitate adjustments in the placement of these items. 

Despite the complexity, the task benefits from a clear and neat hierarchical structure. By decomposing the overall arrangement into subtasks---each focusing on a local arrangement---we can address each component independently. This modular approach ensures that solving one subtask does not adversely affect the others, allowing for efficient and accurate completion of the entire table setting.

\begin{table}[bp]
\centering
\caption{Training examples for arranging dining tables.}
\label{tab:dining_table_training}
\small
\begin{tabular}{c|p{6cm}|p{9cm}}
\hline
\textbf{Index} & \textbf{User Instruction} & \textbf{Object List} \\ 
\hline
1 & Could you please arrange a dining table for two people? & serving\_plate\_1, napkin\_1, fork\_1, knife\_1, spoon\_1, serving\_plate\_2, napkin\_2, fork\_2, knife\_2, spoon\_2  \\ 
\hline
2 & Please prepare a Chinese-style dining table for two guests. & medium\_plate\_1, medium\_plate\_2, small\_plate\_1, small\_plate\_2, rice\_bowl\_1, rice\_bowl\_2, chopsticks\_1, chopsticks\_2, spoon\_1, spoon\_2 \\ 
\hline
3 & Could you please arrange a dining table for two? We would like to sit side by side? & serving\_plate\_1, napkin\_1, fork\_1, knife\_1, spoon\_1, serving\_plate\_2, napkin\_2, fork\_2, knife\_2, spoon\_2, seasoning\_1, seasoning\_2, seasoning\_3 \\ 
\hline
4 & Could you please set up the ramen dining table for two, ensuring that the seating accommodates one left-handed diner? & ramen\_bowl\_1, chopsticks\_1, spoon\_1, ramen\_bowl\_2, chopsticks\_2, spoon\_2,  seasoning\_1, seasoning\_2, seasoning\_3  \\ 
\hline
5 & Instruction: Could you please set up a dining table for two, with the setup for sharing the main dishes? & medium\_plate\_1, medium\_plate\_2, serving\_plate\_1, napkin\_1, fork\_1, knife\_1, spoon\_1, serving\_plate\_2, napkin\_2, fork\_2, knife\_2, spoon\_2, seasoning\_1, seasoning\_2, seasoning\_3  \\ 
\hline
\end{tabular}
\end{table}

\begin{table}[hp]
    \centering
    \caption{Test cases for arranging dining tables.}
    \label{tab:dining_table_testing}
    \small
    \begin{tabular}{c|p{5cm}|p{7cm}|p{2cm}}
    \hline
    \textbf{Index} & \textbf{User Instruction} & \textbf{Object List} & \textbf{\# Objects} \\ 
    \hline
    1 & Could you please set up a western dining table for me with the seasoning on the side? & serving\_plate, napkin, fork, knife, spoon, glass, seasoning\_1, seasoning\_2, seasoning\_3 & 9 \\ 
    \hline
    2 & Could you please set up a Chinese-style dining table for me with the seasoning on the side? & medium\_plate\_1, medium\_plate\_2, medium\_plate\_3, small\_plate, rice\_bowl, chopsticks, spoon, glass, seasoning\_1, seasoning\_2, seasoning\_3 & 11 \\ 
    \hline
    3 & Could you please set up a ramen dining table for me with the seasoning on the side? & ramen\_bowl, chopsticks, spoon, glass, seasoning\_1, seasoning\_2, seasoning\_3 & 7 \\ 
    \hline
    4 & Could you please set up a dining table for two people with drinks on the side? & serving\_plate\_1, napkin\_1, fork\_1, knife\_1, glass\_1, spoon\_1, serving\_plate\_2, napkin\_2, fork\_2, knife\_2, spoon\_2, glass\_2, seasoning\_1, seasoning\_2, seasoning\_3 & 15 \\ 
    \hline
    5 & Could you please set up a western dining table for two people sitting side by side with drinks on the side? & serving\_plate\_1, napkin\_1, fork\_1, knife\_1, glass\_1, spoon\_1, serving\_plate\_2, napkin\_2, fork\_2, knife\_2, spoon\_2, glass\_2, seasoning\_1, seasoning\_2, seasoning\_3 & 15 \\ 
    \hline
    6 & Could you please set up a western dining table for a parent and a kid? They should sit side by side so the parent can help the kid. & serving\_plate\_1, napkin\_1, fork\_1, knife\_1, glass\_1, spoon\_1, baby\_plate, baby\_spoon, baby\_cup, seasoning\_1, seasoning\_2, seasoning\_3 & 12 \\ 
    \hline
    7 & Could you please set up a western dining table for two people with drinks on the side? Both of them are left-handed. & serving\_plate\_1, napkin\_1, fork\_1, knife\_1, spoon\_1, glass\_1, serving\_plate\_2, napkin\_2, fork\_2, knife\_2, spoon\_2, glass\_2, seasoning\_1, seasoning\_2, seasoning\_3 & 15 \\ 
    \hline
    8 & Could you please set up a ramen dining table for a parent and a kid? They should sit side by side so the parent can help the kid. & ramen\_bowl, chopsticks, spoon, baby\_bowl, baby\_spoon, baby\_cup, seasoning\_1, seasoning\_2, seasoning\_3 & 9 \\ 
    \hline
    9 & Could you please set up a Chinese-style dining table for two people with dishes for sharing? & medium\_plate\_1, medium\_plate\_2, medium\_plate\_3, medium\_plate\_4, small\_plate\_1, rice\_bowl\_1, chopsticks\_1, spoon\_1, glass\_1, small\_plate\_2, rice\_bowl\_2, chopsticks\_2, spoon\_2, glass\_2, seasoning\_1, seasoning\_2, seasoning\_3 & 17 \\ 
    \hline
    10 & Could you please set up a ramen dining table for two people with dishes for sharing? & medium\_plate\_1, medium\_plate\_2, medium\_plate\_3, ramen\_bowl\_1, chopsticks\_1, spoon\_1, glass\_1, ramen\_bowl\_2, chopsticks\_2, spoon\_2, glass\_2, seasoning\_1, seasoning\_2 & 13 \\ 
    \hline
    11 & Could you please set up a western dining table for four people? & serving\_plate\_1, napkin\_1, fork\_1, knife\_1, glass\_1, spoon\_1, serving\_plate\_2, napkin\_2, fork\_2, knife\_2, glass\_2, spoon\_2, serving\_plate\_3, napkin\_3, fork\_3, knife\_3, glass\_3, spoon\_3, serving\_plate\_4, napkin\_4, fork\_4, knife\_4, glass\_4, spoon\_4 & 24 \\ 
    \hline
    12 & Could you please set up a western dining table for four people? Two of them are left-handed and they should sit facing each other. & serving\_plate\_1, napkin\_1, fork\_1, knife\_1, glass\_1, spoon\_1, serving\_plate\_2, napkin\_2, fork\_2, knife\_2, glass\_2, spoon\_2, serving\_plate\_3, napkin\_3, fork\_3, knife\_3, glass\_3, spoon\_3, serving\_plate\_4, napkin\_4, fork\_4, knife\_4, glass\_4, spoon\_4, seasoning\_1, seasoning\_2, seasoning\_3 & 27 \\ 
    \hline
    13 & Could you please set up a Chinese-style dining table for four people with sharing main dishes? & medium\_plate\_1, medium\_plate\_2, medium\_plate\_3, medium\_plate\_4, small\_plate\_1, rice\_bowl\_1, chopsticks\_1, spoon\_1, glass\_1, small\_plate\_2, rice\_bowl\_2, chopsticks\_2, spoon\_2, glass\_2, small\_plate\_3, rice\_bowl\_3, chopsticks\_3, spoon\_3, glass\_3, small\_plate\_4, rice\_bowl\_4, chopsticks\_4, spoon\_4, glass\_4, seasoning\_1, seasoning\_2, seasoning\_3 & 27 \\ 
    \hline
    14 & Could you please set up a ramen dining table for four people with two kids? Each kid should sit side by side by one parent so the parent can help them. & ramen\_bowl\_1, chopsticks\_1, spoon\_1, ramen\_bowl\_2, chopsticks\_2, spoon\_2, baby\_bowl\_1, baby\_spoon\_1, baby\_cup\_1, baby\_bowl\_2, baby\_spoon\_2, baby\_cup\_2, seasoning\_1, seasoning\_2, seasoning\_3 & 15 \\ 
    \hline
    15 & Could you please set up a Chinese-style dining table for four? One of them is left-handed and there is a kid. & medium\_plate\_1, medium\_plate\_2, medium\_plate\_3, medium\_plate\_4, small\_plate\_1, rice\_bowl\_1, chopsticks\_1, spoon\_1, glass\_1, small\_plate\_2, rice\_bowl\_2, chopsticks\_2, spoon\_2, glass\_2, small\_plate\_3, rice\_bowl\_3, chopsticks\_3, spoon\_3, glass\_3, baby\_plate, baby\_spoon, baby\_cup, seasoning\_1, seasoning\_2, seasoning\_3 & 25 \\ 
    \hline
    \end{tabular}
\end{table}

\subsection{Bookshelf}

\begin{figure*}[ht]
    \centering
    \includegraphics[width=\linewidth]{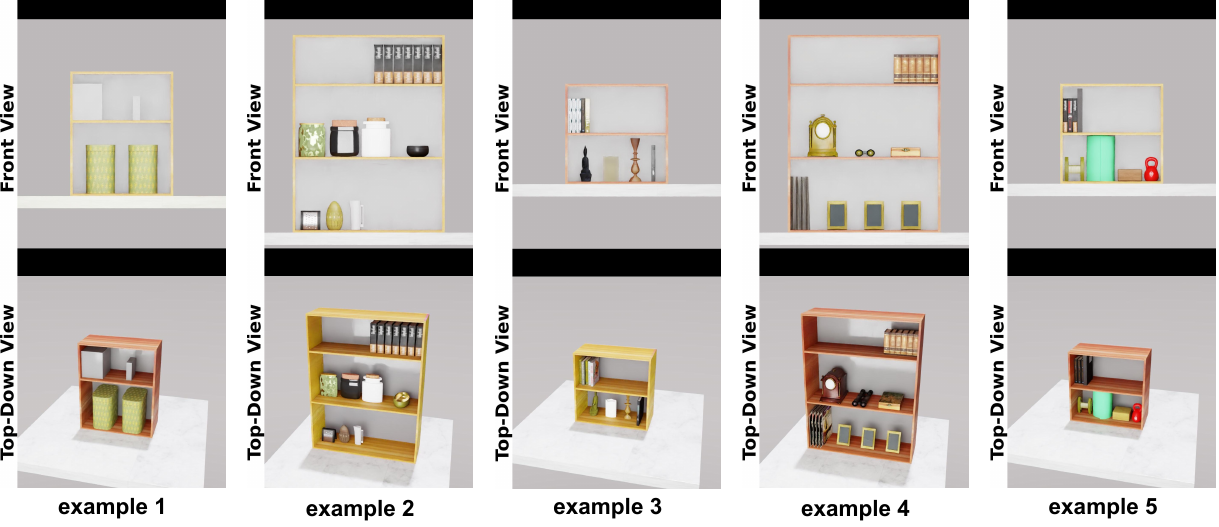}
    \caption{Five example FORM for Bookshelf Arrangement.}
    \label{fig:bookshelf_form}
\end{figure*}

\begin{itemize}
    \item \textbf{Purpose}: To arrange the objects on a bookshelf with different levels of compartments. The arrangement needs to be functional and satisfy the requirement or preference indicated in the instructions.
    \item \textbf{Objects}: binder, folder, laptop\_close, pen\_holder, hardcover\_book, small\_watering\_can, nutrition\_bottle, pesticide\_bottle, succulent\_1, succulent\_2, succulent\_3, potted\_plant\_1, potted\_plant\_2, potted\_plant\_3, environmental\_book, dictionary, comic\_book, novel, textbook, small\_vase, medium\_vase, large\_vase, photo\_album, vintage\_suitcase, souvenir\_1, souvenir\_2, souvenir\_3, souvenir\_4, souvenir\_5, travel\_guide, small\_globe, smart\_speaker, old\_computer\_component, vr\_headset, tech\_magazine, instruction\_manual, coding\_book, canvas, jar\_with\_paintbrushes, sketchbook, brush, pencils, paint\_tubes, framed\_painting\_1, framed\_painting\_2, framed\_painting\_3, framed\_painting\_4, art\_book, harmonica, tambourine, headphones, vinyl\_record, small\_turntable, music\_history\_book, biography\_of\_musician, basket\_of\_toys, children\_book, puzzles, kitchen\_gadget\_grater, kitchen\_gadget\_timer, kitchen\_gadget\_measuring\_cup, jar\_with\_dried\_herbs\_and\_spices\_1, jar\_with\_dried\_herbs\_and\_spices\_2, jar\_with\_dried\_herbs\_and\_spices\_3, decorative\_bowl, cookbook, small\_buddha\_statue, candle, incense\_holder, journal, self\_help\_book, antique\_old\_clock, antique\_binoculars, antique\_ornate\_box, old\_leather\_bound\_book, dumbbell, yoga\_mat, kettle\_bell, fitness\_book, workout\_log, collector\_catalogue, miniature\_car, miniature\_airplane, jar\_of\_collectible\_coin, decoration\_jar, stamp, tripod, light\_meter, lighting\_setup, camera\_lens\_1, camera\_lens\_2, camera, roll\_of\_film, photography\_book, photograph\_frame, gardening\_tool, repotting\_pot, bag\_of\_soil, plant\_spray\_bottle, botany\_book, mannequin\_torso, hat\_1, hat\_2, jewellery\_display\_stand, handbag\_1, handbag\_2, designer\_shoe\_box\_1, designer\_shoe\_box\_2, sunglasses, fashion\_book, craft\_project\_1, craft\_project\_2, extra\_materials\_bin, basket\_yarn, basket\_thread, basket\_fabric, scissors, sewing\_kit, craft\_book, old\_college\_textbook, notebook, reference\_book\_dictionary, reference\_book\_thesaurus, reference\_book\_encyclopedia, academic\_book, dvd, blu\_ray, movie\_memorabilia\_1, movie\_memorabilia\_2, action\_figure\_1, action\_figure\_2, funko\_pop\_toy\_1, funko\_pop\_toy\_2, book\_about\_films, book\_about\_tv\_shows, aromatherapy\_kit, essential\_oils, meditation\_cushion, yoga\_block, self\_care\_book.
    \item \textbf{Training \& Testing datasets}: Table \ref{tab:bookshelf_training} and Figure \ref{fig:bookshelf_form} contains the training examples, each consisting of user instructions, object lists, and FORM layouts. The testing dataset, which includes user instructions and object lists, is presented in Table \ref{tab:bookshelf_testing}.
\end{itemize}

\begin{table}[bp]
    \centering
    \caption{Training examples for arranging bookshelves.}
    \label{tab:bookshelf_training}
    \small
    \begin{tabular}{c|p{4cm}|p{1.5cm}|p{9cm}}
    \hline
    \textbf{Index} & \textbf{User Instruction} & \textbf{\# Compartments} & \textbf{Object List}  \\ 
    \hline
    1 & Children's books need to be easily accessible, so put them on the top shelf. & 2 & basket\_of\_toys\_1, basket\_of\_toys\_2, children\_book\_1, children\_book\_2, children\_book\_3, children\_book\_4, children\_book\_5, children\_book\_6, puzzles  \\ 
    \hline
    2 & I reference my cookbooks regularly, so put them on the top shelf. & 3 & kitchen\_gadget\_grater, kitchen\_gadget\_timer, kitchen\_gadget\_measuring\_cup, jar\_with\_dried\_herbs\_and\_spices\_1, jar\_with\_dried\_herbs\_and\_spices\_2, jar\_with\_dried\_herbs\_and\_spices\_3, decorative\_bowl, cookbook\_1, cookbook\_2, cookbook\_3, cookbook\_4, cookbook\_5, cookbook\_6, cookbook\_7  \\ 
    \hline
    3 & Keep my meditation items like candles and incense on the bottom shelf where I can easily access them during my sessions. & 2 & small\_buddha\_statue, candle, incense\_holder, journal, self\_help\_book\_1, self\_help\_book\_2, self\_help\_book\_3, self\_help\_book\_4, self\_help\_book\_5  \\ 
    \hline
    4 & Arrange antique items like clocks or keepsakes on the middle shelf. & 3 & vinyl\_record\_1, vinyl\_record\_2, vinyl\_record\_3, vinyl\_record\_4, photograph\_frame\_1, photograph\_frame\_2, photograph\_frame\_3, antique\_old\_clock, antique\_binoculars, antique\_ornate\_box, old\_leather\_bound\_book\_1, old\_leather\_bound\_book\_2, old\_leather\_bound\_book\_3, old\_leather\_bound\_book\_4, old\_leather\_bound\_book\_5, old\_leather\_bound\_book\_6  \\ 
    \hline
    5 & Keep my workout equipment on the bottom shelf where I can grab it easily when I'm ready to exercise. & 2 & dumbbell, yoga\_mat, yoga\_block, kettle\_bell, fitness\_book\_1, fitness\_book\_2, fitness\_book\_3, workout\_log\_1, workout\_log\_2 \\ 
    \hline
    \end{tabular}
\end{table}

\begin{table}[bp]
    \centering
    \caption{Test cases for arranging items on bookshelves.}
    \label{tab:bookshelf_testing}
    \footnotesize 
    \begin{tabular}{c|p{3.5cm}|c|p{7cm}|c}
    \hline
    \textbf{Index} & \textbf{User Instruction} & \textbf{\# Compartments} & \textbf{Object List} & \textbf{\# Objects} \\ 
    \hline
    1 & Keep my laptop and a few office supplies in the middle for quick access. & 3 & binder\_1, binder\_2, binder\_3, folder\_1, folder\_2, folder\_3, laptop\_close, pen\_holder, succulent\_1, hardcover\_book\_1, hardcover\_book\_2, hardcover\_book\_3, hardcover\_book\_4, hardcover\_book\_5, hardcover\_book\_6, hardcover\_book\_7, hardcover\_book\_8 & 17 \\ 
    \hline
    2 & I love having plants nearby, so arrange them on the middle shelf where they can get light and add some greenery. & 3 & small\_watering\_can, nutrition\_bottle, pesticide\_bottle, potted\_plant\_1, potted\_plant\_2, potted\_plant\_3, succulent\_1, succulent\_2, succulent\_3, environmental\_book\_1, environmental\_book\_2, environmental\_book\_3, environmental\_book\_4, environmental\_book\_5, environmental\_book\_6, environmental\_book\_7, environmental\_book\_8, dictionary & 18 \\ 
    \hline
    3 & Please arrange the books according to genres. & 4 & dictionary\_1, dictionary\_2, dictionary\_3, comic\_book\_1, comic\_book\_2, comic\_book\_3, novel\_1, novel\_2, novel\_3, novel\_4, novel\_5, novel\_6, textbook\_1, textbook\_2, textbook\_3, textbook\_4, textbook\_5, textbook\_6, textbook\_7, textbook\_8, small\_vase, medium\_vase, large\_vase & 23 \\ 
    \hline
    4 & Display my souvenirs and travel mementos in the middle, where they're visible but not in the way. & 3 & photo\_album\_1, photo\_album\_2, photo\_album\_3, camera, vintage\_suitcase, souvenir\_1, souvenir\_2, souvenir\_3, souvenir\_4, souvenir\_5, travel\_guide\_1, travel\_guide\_2, travel\_guide\_3, travel\_guide\_4, travel\_guide\_5, travel\_guide\_6, small\_globe & 17 \\ 
    \hline
    5 & Keep my gadgets and equipment on the bottom shelf where I can reach them easily. & 2 & smart\_speaker, old\_computer\_component, vr\_headset, tech\_magazine\_1, tech\_magazine\_2, tech\_magazine\_3, instruction\_manual\_1, instruction\_manual\_2, instruction\_manual\_3, coding\_book\_1, coding\_book\_2, coding\_book\_3 & 12 \\ 
    \hline
    6 & Keep my art supplies within reach on the second shelf so I can grab them while working. & 4 & canvas\_1, canvas\_2, canvas\_3, canvas\_4, jar\_with\_paintbrushes, sketchbook\_1, sketchbook\_2, sketchbook\_3, brush, pencils, paint\_tubes, framed\_painting\_1, framed\_painting\_2, framed\_painting\_3, framed\_painting\_4, art\_book\_1, art\_book\_2, art\_book\_3, art\_book\_4, art\_book\_5, art\_book\_6, art\_book\_7, art\_book\_8 & 23 \\ 
    \hline
    7 & I like to listen to vinyl regularly, so place those on the middle shelves for easy access. & 3 & harmonica, tambourine, headphones, smart\_speaker, vinyl\_record\_1, vinyl\_record\_2, vinyl\_record\_3, vinyl\_record\_4, small\_turntable, music\_history\_book\_1, music\_history\_book\_2, music\_history\_book\_3, biography\_of\_musician\_1, biography\_of\_musician\_2, biography\_of\_musician\_3 & 15 \\ 
    \hline
    8 & My rare books should be on display, so place them on the top shelf. & 4 & collector\_catalogue\_1, collector\_catalogue\_2, collector\_catalogue\_3, collector\_catalogue\_4, miniature\_car\_1, miniature\_airplane\_1, jar\_of\_collectible\_coin, decoration\_jar, stamp, old\_leather\_bound\_book\_1, old\_leather\_bound\_book\_2, old\_leather\_bound\_book\_3, old\_leather\_bound\_book\_4 & 13 \\ 
    \hline
    9 & Keep my camera gear and lenses on the middle shelf where they're easy to grab. & 3 & tripod, light\_meter, lighting\_setup, camera\_lens\_1, camera\_lens\_2, camera\_1, camera\_2, roll\_of\_film\_1, roll\_of\_film\_2, photography\_book\_1, photography\_book\_2, photography\_book\_3, photograph\_frame\_1, photograph\_frame\_2 & 14 \\ 
    \hline
    10 & Arrange my potted plants on the middle shelf where they get good light. & 3 & gardening\_tool, repotting\_pot\_1, repotting\_pot\_2, bag\_of\_soil\_1, potted\_plant\_1, potted\_plant\_2, potted\_plant\_3, plant\_spray\_bottle, botany\_book\_1, botany\_book\_2, botany\_book\_3, botany\_book\_4, botany\_book\_5, botany\_book\_6 & 14 \\ 
    \hline
    11 & Store my designer accessories and bags on the second shelf where they're easy to display. & 4 & mannequin\_torso, hat\_1, hat\_2, jewellery\_display\_stand, handbag\_1, handbag\_2, designer\_shoe\_box\_1, designer\_shoe\_box\_2, sunglasses, fashion\_book\_1, fashion\_book\_2, fashion\_book\_3, fashion\_book\_4, fashion\_book\_5, fashion\_book\_6, fashion\_book\_7, fashion\_book\_8 & 17 \\ 
    \hline
    12 & I need my craft books on hand while working, so they should go on the top shelf. & 3 & craft\_project\_1, craft\_project\_2, extra\_materials\_bin, basket\_yarn, basket\_thread, basket\_fabric, scissors, sewing\_kit, craft\_book\_1, craft\_book\_2, craft\_book\_3, craft\_book\_4 & 12 \\ 
    \hline
    13 & Journals and reference materials can go on the second shelf for easy access. & 4 & folder\_1, folder\_2, small\_globe, old\_college\_textbook\_1, old\_college\_textbook\_2, notebook\_1, notebook\_2, pen\_holder, reference\_book\_dictionary, reference\_book\_thesaurus, reference\_book\_encyclopedia, academic\_book\_1, academic\_book\_2, academic\_book\_3, academic\_book\_4 & 15 \\ 
    \hline
    14 & Display my memorabilia on the middle shelf where it's visible. & 3 & dvd\_1, dvd\_2, dvd\_3, dvd\_4, blu\_ray\_1, blu\_ray\_2, movie\_memorabilia\_1, movie\_memorabilia\_2, action\_figure\_1, action\_figure\_2, funko\_pop\_toy\_1, funko\_pop\_toy\_2, comic\_book\_1, comic\_book\_2, comic\_book\_3, book\_about\_films\_1, book\_about\_tv\_shows\_1 & 17 \\ 
    \hline
    15 & Keep my meditation tools on the bottom shelf where they're easily accessible during my wellness routines. & 2 & aromatherapy\_kit, essential\_oils, meditation\_cushion, yoga\_block, self\_care\_book\_1, self\_care\_book\_2, self\_care\_book\_3, self\_care\_book\_4, self\_care\_book\_5, self\_care\_book\_6, self\_care\_book\_7 & 11 \\ 
    \hline
    \end{tabular}
\end{table}

Organizing items on a bookshelf presents an \textbf{extremely open-ended semantic inference challenge} due to the huge amount of open-set objects. With 140 different object types tested across 20 distinct scenarios---such as "Minimalist Professional," "Nature Lover," "Bookworm Chaos," and "Tech Enthusiast"---the task requires interpreting diverse and nuanced user instructions. Each scenario introduces unique preferences and constraints, demanding an understanding of both the functional relations and aesthetic considerations for item placement. The open-ended nature of the task means that there are countless possible arrangements, making it essential to semantically infer the optimal organization that aligns with the user's intent while accommodating the physical limitations of the bookshelf's compartments.

\subsection{Bedroom}

\begin{figure*}[tp]
    \centering
    \includegraphics[width=\linewidth]{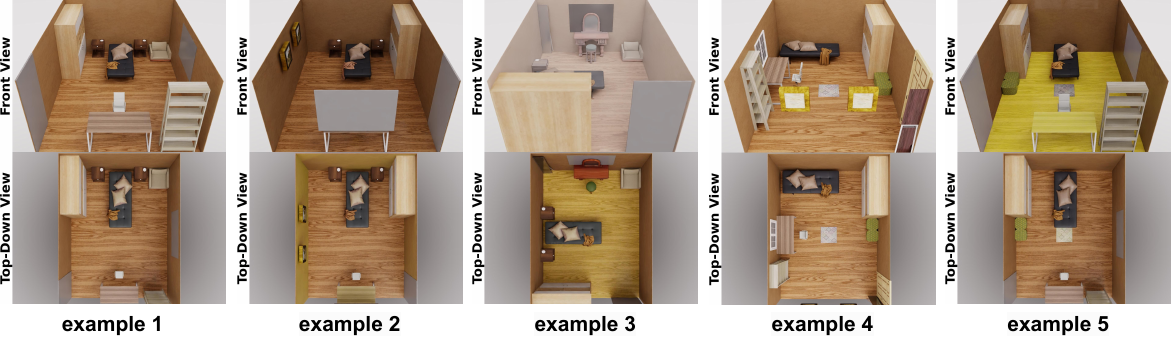}
    \caption{Five example FORM for Bedroom Arrangement.}
    \label{fig:bedroom_form}
\end{figure*}

\begin{itemize}
    \item \textbf{Purpose}: To arrange the furniture in a bedroom with the location of the window and door fixed. The arrangement has to 1) be functional, in the sense the arrangement satisfy the basic functionality of the bedroom, such as the furnitures does not block the door nor the window or people can conviently use the furnitures for it's intended purposes, and 2) adhere to the preference or requirement expressed in the instructions.
    \item \textbf{Objects}: bed, wardrobe, study\_desk, study\_chair, bookshelf, bedside\_table, piano, guitar, carpet, cello, painting\_s, painting\_m, painting\_l, painting\_xl, couch, mirror, dressing\_table, dressing\_chair, storage\_cabinet, Christmas\_tree, television
    \item \textbf{Training \& Testing datasets}: Table \ref{tab:bedroom_training} and Figure \ref{fig:bedroom_form} contains the training examples, each consisting of user instructions, object lists, and FORM layouts. The testing dataset, which includes user instructions and object lists, is presented in Table \ref{tab:bedroom_testing}.
\end{itemize}

Arranging furniture in a bedroom setting presents \textbf{unique challenges} that require \textbf{constraint-aware planning} and \textbf{feasibility checks during semantic inference}. The limited space inherently restricts object placement options, demanding careful consideration to ensure all items fit within the spatial boundaries. Abstract relations between furniture pieces must also respect these spatial limitations, making the task more complex than it might initially appear. As we arrange the objects one by another, they must be integrated sequentially, taking into account the existing layout to avoid conflicts and overlaps. This requires each new piece of furniture to be placed in a way that not only meets the specific instructions but also fulfills functional needs and adheres to physical constraints. 

Unlike tasks with a clear hierarchical structure where subtasks can be solved independently, the arrangements in a bedroom are intertwined. The placement of subsequent furniture items depends heavily on the positions of those already placed, posing a more challenging constraint optimization problem. This interconnectedness necessitates a holistic approach to symbolic object arrangement, ensuring that the final layout is both practical and harmonious.

\begin{table}[bp]
\centering
\caption{Training examples for arranging bedrooms.}
\label{tab:bedroom_training}
\small
\begin{tabular}{c|p{4cm}|p{4cm}|p{6cm}}
\hline
\textbf{Index} & \textbf{User Instruction} & \textbf{Room Layout} & \textbf{Object List} \\ 
\hline
1 & Arrange the furnitures in my bedroom. & The window is on the center of the left wall. The door is on the left of the left wall. & bed, bedside\_table\_1, bedside\_table\_2, wardrobe, study\_desk, study\_chair, bookshelf, couch \\ 
\hline
2 & Arrange the furnitures in my bedroom, the study desk should be facing the window. & The window is on the center of the front wall. The door is on the right of the right wall. & bed, bedside\_table\_1, bedside\_table\_2, wardrobe, study\_desk, study\_chair, painting\_s\_1, painting\_s\_2  \\ 
\hline
3 & Arrange the furnitures in my bedroom, the study desk should not be facing the bed. & The window is on the center of the back wall. The door is on the left of the front wall. & bed, bedside\_table\_1, bedside\_table\_2, wardrobe, dressing\_table, dressing\_chair, couch, mirror  \\ 
\hline
4 & Arrange the furnitures in my bedroom to maximize the space. 
& The window is on the center of the left wall. The door is on the right of the right wall. & bed, wardrobe, study\_desk, study\_chair, bookshelf, storage\_1, storage\_2, carpet, painting\_s\_1, painting\_s\_2, mirror \\ 
\hline
5 & Arrange the furnitures in my bedroom, the bookshelf should be on the left side of the study desk. & The window is on the center of the right wall. The door is on the left of the left wall. & bed, wardrobe, study\_desk, study\_chair, bookshelf, storage\_unit\_1, storage\_unit\_2, carpet \\ 
\hline
\end{tabular}
\end{table}

\begin{table}[bp]
\centering
\caption{Test cases for arranging furniture in a bedroom.}
\label{tab:bedroom_testing}
\small
\begin{tabular}{c|p{3.5cm}|p{4cm}|p{5cm}|p{1cm}}
\hline
\textbf{Index} & \textbf{User Instruction} & \textbf{Room Layout} & \textbf{Object List} & \textbf{\# Furniture} \\ 
\hline
1 & Arrange the furnitures in my bedroom. & The window is on the center of the left wall. The door is on the right of the right wall. & bed, bedside\_table\_1, bedside\_table\_2, wardrobe, study\_desk, study\_chair, bookshelf, couch & 8 \\ 
\hline
2 & Arrange the furnitures in my bedroom, the bed should be facing the TV. & The window is on the center of the right wall. The door is on the left of the left wall. & bed, wardrobe, dressing\_table, dressing\_chair, television, carpet & 6 \\ 
\hline
3 & Arrange the furnitures in my bedroom, the dressing table should not be facing the bed. & The window is on the center of the front wall. The door is on the right of the back wall. & bed, bedside\_table\_1, bedside\_table\_2, wardrobe, dressing\_table, dressing\_chair, couch, painting\_xl, Christmas\_tree & 9 \\ 
\hline
4 & Arrange the furnitures in my bedroom to maximize the space. & The window is on the center of the right wall. The door is on the left of the left wall. & bed, wardrobe, study\_desk, study\_chair, bookshelf, storage\_unit\_1, storage\_unit\_2, carpet & 8 \\ 
\hline
5 & Arrange the furnitures in my bedroom, the study desk should be facing the window. & The window is on the center of the back wall. The door is on the left of the right wall. & bed, bedside\_table\_1, bedside\_table\_2, wardrobe, study\_desk, study\_chair, bookshelf, carpet, painting\_s\_1, painting\_s\_2, painting\_s\_3 & 11 \\ 
\hline
6 & Arrange the furnitures in my bedroom, the bedside table should be on the right side of the bed. &  The window is on the center of the left wall. The door is on the left of the right wall. & bed, bedside\_table, wardrobe, dressing\_table, dressing\_chair, carpet, painting\_s\_1, painting\_s\_2 & 8 \\ 
\hline
7 & Arrange the furnitures in my bedroom, the study desk should be on the right side of the bookshelf. & The window is on the center of the left wall. The door is on the left of the right wall. & bed, bedside\_table\_1, bedside\_table\_2, wardrobe, study\_desk, study\_chair, bookshelf, storage\_unit\_1, storage\_unit\_2, carpet, guitar & 11 \\ 
\hline
8 & Arrange the furnitures in my bedroom, the bookshelf should not be close to the bed. & The window is on the center of the front wall. The door is on the left of the left wall. & bed, wardrobe, bedside\_table\_1, bedside\_table\_2, study\_desk, study\_chair, bookshelf, carpet & 8 \\ 
\hline
9 & Arrange the furnitures in my bedroom, the bookshelf should be close to the bed. & The window is on the center of the front wall. The door is on the left of the left wall. & bed, wardrobe, bedside\_table\_1, bedside\_table\_2, study\_desk, study\_chair, bookshelf, carpet & 8 \\ 
\hline
10 & Arrange the furnitures in my bedroom, the piano should be opposite to the bed. & The window is on the center of the right wall. The door is on the left of the left wall. & bed, wardrobe, bedside\_table\_1, bedside\_table\_2, study\_desk, study\_chair, bookshelf, carpet, piano & 9 \\ 
\hline
11 & Arrange the furnitures in my bedroom, the painting should be facing the bed. & The window is on the center of the back wall. The door is on the left of the front wall. & bed, wardrobe, bedside\_table\_1, bedside\_table\_2, study\_desk, study\_chair, bookshelf, carpet, couch & 9 \\ 
\hline
12 & Arrange the furnitures in my bedroom, the painting should not be facing the bed. & The window is on the center of the back wall. The door is on the left of the front wall. & bed, wardrobe, bedside\_table\_1, bedside\_table\_2, study\_desk, study\_chair, bookshelf, carpet, couch & 9 \\ 
\hline
13 & Arrange the furnitures in my bedroom, the piano should not be facing the bed. & The window is on the center of the back wall. The door is on the left of the right wall. & bed, wardrobe, bedside\_table\_1, bedside\_table\_2, study\_desk, study\_chair, carpet, piano, Christmas\_tree & 9 \\ 
\hline
14 & Arrange the furnitures in my bedroom, the paintings should be on the same wall. & The window is on the center of the back wall. The door is on the left of the left wall. & bed, wardrobe, bedside\_table\_1, bedside\_table\_2, study\_desk, study\_chair, bookshelf, carpet, painting\_s\_1, painting\_s\_2, painting\_s\_1, painting\_s\_2 & 12 \\ 
\hline
15 & Arrange the furnitures in my bedroom, the music instruments should be on the same wall. & The window is on the center of the front wall. The door is on the left of the left wall. & bed, wardrobe, bedside\_table\_1, bedside\_table\_2, couch, carpet, guitar, piano, cello & 9 \\ 
\hline
\end{tabular}
\end{table}

\section{Baselines}
\label{append:baseline_implementation}

For our study, we have implemented five baseline models: the End-to-end Diffusion Model; two Direct LLM Prediction methods---one with task specification and one without; and two ablation models that simplify the semantic inference component---LLM(graph) + DMs, where the LLM directly predicts the intermediate symbolic grounding graph, and LLM(program) + DMs, where the LLM directly predicts the Python program. Detailed descriptions of these models are provided below.

\vspace{1em}
\noindent
\subsection{End-to-end Diffusion Model:}
Drawing inspiration from StructDiffusion \cite{liu2022structdiffusion}, we developed a language-conditioned diffusion model. This model is designed to predict object poses in FORM arrangements. It has been trained using the same datasets as our main model. These datasets include a synthetic object relation dataset and 15 specific FORM examples.

\vspace{0.5em}
For synthetic objects, we employ a script that describes the geometric relations between scene objects as the language input. For FORM scenes, the language input comes from user instructions. We then encode these language inputs into language feature embedding using CLIP embedding. In our model, each object is represented by a combined token of its shape and pose embeddings. We have configured the transformer to handle up to 16 objects, which corresponds to its sequence length limit of 16. The model processes the language embeddings together with object sequence embeddings and time embeddings through a transformer encoder. This encoder uses four layers of residual attention blocks and is equipped with two heads. We also incorporate normalization layers both before and after the transformer layers. Finally, we utilize the same diffusion timesteps $t=1500$ in this model as in our \model.

\vspace{0.5em}
\myparagraph{Modification on StructDiffusion \cite{liu2022structdiffusion}}: Our adaptation maintains the core framework of StructDiffusion, which is a multimodal transformer that conditions on both the language embeddings from instructions and the objects' geometries. Key modifications are made for data format compatibility and architectural coherence.

\vspace{0.5em}

\begin{itemize}
    \item \textbf{Model Architecture}: We changed orginal 3D point cloud (Point Cloud Transformer) and a 6DOF pose encoder in StructDiffusion to the 2D geometry and pose encoder (same as \model) to fit the 2D bounding box object geometry and pose encoder used in \model. Both models utilize MLPs for diffusion encoding; our adaptation simply modifies the hidden layer parameters to align with the architecture of \model for consistent performance evaluation.
    \item \textbf{Different Training Data}: We use the training dataset for \model, which differs from the required training dataset for StructDiffusion. We changed from the orginal extensive dataset on broad paired instructions and scene-level object arrangements to the \model dataset, which only includes 1) 5 examples of the paired instructions and scene-level object arrangements, and 2) synthetic datasets labelled by their abstract relations.
\end{itemize}

\vspace{0.5em}

These necessary modifications ensure that while the adapted StructDiffusion framework aligns closely with the original’s multi-modal transformative capabilities, it also enhances compatibility with \model’s data formats and model architecture choices. 

\vspace{1em}
\noindent
\subsection{Direct LLM Prediction} 
Drawing on Tidybot's approach, we use an LLM-based method to directly propose the object poses for FORM arrangements. We implement two variants, one takes the natural language task specification through its prompt and the other does not. We detail the prompts for each of these methods below.

\paragraph{Prompts for Direct LLM Prediction without Task Specification}

Below, we detail the prompt \textbf{without} task specification used for Direct LLM Prediction on the dining table task family.

\begin{lstlisting}[caption={Prompt for Direct LLM Prediction w/o Task Specification}]
Your task is to generate precise object poses to arrange tableware objects on a dining table, ensuring the arrangement satisfies a given instruction.

The dining table is a rectangle measuring 3 units in length and 2 units in width. The coordinate system is centered at the front-left corner, with the origin (0, 0) at the center of the tabletop. Object positions, specified by coordinates (x, y, \theta), must fall within:
- Length (x-axis): -1.5 < x < 1.5
- Width (y-axis): -1 < y < 1
- Rotation angle \theta in radians

Here, x and y represent the centroid coordinates of an object, and \theta is the 2D rotation angle. The angle \theta corresponds to:
- \theta = 0: Object faces the front
- \theta = np.pi: Faces the back
- \theta = np.pi/2: Faces right
- \theta = -np.pi/2: Faces left
(All orientations are viewed from the front.) Each object's dimensions are given as a 2D bounding box (w, l).

For each task, you'll receive:
1. A natural language instruction indicating the desired functional arrangement of the objects.
2. A list of objects, including their names and 2D bounding boxes.

You need to predict the object poses that satisfy the instruction, formatted as:

```json
{
  "serving_plate": {
    "dimension": [0.48, 0.48],
    "pose": [0.0, -0.66, 0],
    "name": "serving_plate"
  },
  "napkin": {
    "dimension": [0.15, 0.52],
    "pose": [-0.465, -0.64, 0],
    "name": "napkin"
  },
  ...
}
```

To assist you, you'll be provided with five example arrangements, each consisting of an instruction and the corresponding functional arrangement in JSON format. These examples will be given one by one. Please begin the arrangement task only after you have received all five examples and the test instruction with the input object list.

<5 instruction-example pairs>

\end{lstlisting}

\paragraph{Prompts for Direct LLM Prediction with Task Specification}

Below, we detail the prompt \textbf{with} task specification used for Direct LLM Prediction on the dining table task family.

\begin{lstlisting}[caption={Prompt for Direct LLM Prediction w/o Task Specification}]

Your task is to generate precise object poses to arrange tableware objects on a dining table, ensuring the arrangement satisfies a given instruction.

The dining table is a rectangle measuring 3 units in length and 2 units in width. The coordinate system is centered at the front-left corner, with the origin (0, 0) at the center of the tabletop. Object positions, specified by coordinates (x, y, \theta), must fall within:
- Length (x-axis): -1.5 < x < 1.5
- Width (y-axis): -1 < y < 1
- Rotation angle \theta in radians

Here, x and y represent the centroid coordinates of an object, and \theta is the 2D rotation angle. The angle \theta corresponds to:
- \theta = 0: Object faces the front
- \theta = np.pi: Faces the back
- \theta = np.pi/2: Faces right
- \theta = -np.pi/2: Faces left
(All orientations are viewed from the front.) Each object's dimensions are given as a 2D bounding box (w, l).

For each task, you'll receive:
1. A natural language instruction indicating the desired functional arrangement of the objects.
2. A list of objects, including their names and 2D bounding boxes.

You need to predict the object poses that satisfy the instruction, formatted as:

```json
{
  "serving_plate": {
    "dimension": [0.48, 0.48],
    "pose": [0.0, -0.66, 0],
    "name": "serving_plate"
  },
  "napkin": {
    "dimension": [0.15, 0.52],
    "pose": [-0.465, -0.64, 0],
    "name": "napkin"
  },
  ...
}
```

To assist you, you will be provided with a heirarchical task description and five examples.

Here's the natural language description on how to arrangement the objects for a dining table setup:

<Hierarchical Natural Language Task Specification>

You are also provided with five example arrangements, each consisting of an instruction and the corresponding functional arrangement in JSON format. These examples will be given one by one. Please begin the arrangement task only after you have received all five examples and the test instruction with the input object list.

<5 instruction-example pairs>

\end{lstlisting}

The two methods only differ whether to include the hierarchical natural language task specification as the input to the LLMs. The rest of the design remains the same.
Specifically, we feed LLMs the same task instructions, object sets, and the few-shot examples. Additionally, we provide details of the table layout and include the object shapes in the few-shot training examples. It's important to note that, unlike our model, which only processes abstract relations, the LLM is given actual shapes from the ground truth. We then instruct the LLM to identify common patterns within these examples, following the methodology proposed by Tidybot. For inference, we supply the user instructions, a list of objects, shapes of these objects to the LLM. Then, we ask the LLM to predict the poses for objects in FORM arrangements.

\vspace{0.5em}
\textbf{TidyBot Adaptation}: We use TidyBot's core strategy of using the summarization capabilities of the LLM to generalize from a limited number of examples to novel scenarios. Our adaptation adjust the summarization and pose inference mechanisms to better suit the specific needs of \model.

\vspace{0.5em}

\begin{itemize}
    \item \textbf{Summarization}: While TidyBot focuses on placing objects into designated receptacles, our objectives involve prescribing more complex and joint object poses. This necessitates richer and more intricate rules. To handle this complexity, we input additional information (so TidyBot-variant has the same information as \model), including the task instruction, sketch, abstract relation library, and the CoT prompt.
    \item \textbf{Pose Inference}: Tidybot requires the LLM to output the abstract action primitives and the receptables for ``tidy arrangement", yet we need the object poses. Therefore, while the tasks in the Tidybot project requires no geometric understanding from LLM, FORM requires the LLM to be able to understand the geometries so as to propose the object poses. To help LLM with this, we provide the table layout and the object dimensions for few-shot learning.
\end{itemize}

\vspace{0.5em}

These adjustments ensure our adaptation of TidyBot not only retains the advantageous summarization for learning from minimal data but also meets the enhanced requirements of \model for precise and context-aware object placement.

\vspace{1em}
\noindent
\subsection{LLM + Diffusion Variants}

To illustrate the importance of our compositional program induction, we implement two ablations of our \model. Both methods follow the same neuro-symbolic framework as \model, using the grounding graph as an intermediate prediction task to decompose the goal specification into semantic inference and geometric reasoning. They use the same compositional diffusion models as \model to ground this abstract grounding graph to exact object poses. The key difference lies in the semantic inference components, where they simplify the grounding graph prediction in different ways. Both methods take the same hierarchical natural language task specification and the five instruction-example pairs as input, identical to \model. They differ only in how they generate the abstract spatial relation graphs.

\paragraph{LLM(Graph) + DMs}

In the first variant, LLM(Graph) + DMs, we implement few-shot learning. The LLM takes the task specification as background context and includes the five demonstrations as examples in the system prompt. It is then prompted to output the grounding graph directly, given new instructions and object sets. We detail the prompts used for LLM(Graph) + DMs below.

\begin{lstlisting}[caption={Prompt for LLM(Graph) + DMs}]
Your task is to create a set of abstract relations for arranging tableware on a dining table, ensuring the arrangement complies with given instructions. The objects are represented as 2D bounding boxes viewed from above, with all orientations considered from the front.

**Use only the relations defined in the following library.** Each abstract relation is a tuple of strings: the first element is the relation's name, followed by the associated object names.

**relation Library:**

- **("near-front-edge", obj_A):** obj_A is near the table's front edge (bottom in the top-down view).
- **("near-back-edge", obj_A):** obj_A is near the table's back edge (top in the top-down view).
- **("near-left-edge", obj_A):** obj_A is near the table's left edge.
- **("near-right-edge", obj_A):** obj_A is near the table's right edge.
- **("facing-front", obj_A):** obj_A is facing the front (from the table's front edge).
- **("facing-back", obj_A):** obj_A is facing the back (from the table's front edge).
- **("front-half", obj_A):** obj_A is entirely within the front half of the table (horizontally divided along the center).
- **("back-half", obj_A):** obj_A is entirely within the back half of the table.
- **("left-half", obj_A):** obj_A is entirely within the left half of the table (vertically divided along the center).
- **("right-half", obj_A):** obj_A is entirely within the right half of the table.
- **("central-column", obj_A):** The centroid of obj_A is within the central vertical column of the table (middle half of the table's width).
- **("central-row", obj_A):** The centroid of obj_A is within the central horizontal row of the table (middle half of the table's length).
- **("centered-table", obj_A):** The centroid of obj_A coincides with the table's center.
- **("horizontally-aligned-bottom", obj_A, obj_B):** obj_A and obj_B are horizontally aligned at their bases.
- **("horizontally-aligned-centroid", obj_A, obj_B):** obj_A and obj_B are horizontally aligned at their centroids.
- **("vertically-aligned-centroid", obj_A, obj_B):** obj_A and obj_B are vertically aligned at their centroids.
- **("left-of", obj_A, obj_B):** obj_A is immediately to the left of obj_B (from the canonical camera frame).
- **("right-of", obj_A, obj_B):** obj_A is immediately to the right of obj_B.
- **("front-of", obj_A, obj_B):** obj_A is immediately in front of obj_B.
- **("back-of", obj_A, obj_B):** obj_A is immediately behind obj_B.
- **("on-top-of", obj_A, obj_B):** obj_A is placed on top of obj_B.
- **("centered", obj_A, obj_B):** The center of obj_A coincides with the center of obj_B.
- **("vertical-symmetry-on-table", obj_A, obj_B):** obj_A and obj_B are symmetrically placed across the table's vertical axis.
- **("horizontal-symmetry-on-table", obj_A, obj_B):** obj_A and obj_B are symmetrically placed across the table's horizontal axis.
- **("vertical-line-symmetry", axis_obj, obj_A, obj_B):** obj_A and obj_B are symmetrical about the vertical axis of axis_obj.
- **("horizontal-line-symmetry", axis_obj, obj_A, obj_B):** obj_A and obj_B are symmetrical about the horizontal axis of axis_obj.
- **("aligned-in-horizontal-line-bottom", obj_1, ..., obj_N):** Objects are horizontally aligned at their bases with equal spacing.
- **("aligned-in-horizontal-line-centroid", obj_1, ..., obj_N):** Objects are horizontally aligned at their centroids with equal spacing.
- **("aligned-in-vertical-line-centroid", obj_1, ..., obj_N):** Objects are vertically aligned at their centroids with equal spacing.
- **("regular-grid", obj_1, ..., obj_N):** Objects are arranged in a grid or matrix pattern.

For each task, you will receive:
1. A natural language instruction indicating the desired arrangement of the objects.
2. A list of objects, each with a unique name.

You need to predict the set of abstract relations (constraints) that satisfy the instruction, formatted as:

```json
{
  "constraints":
    [
        ["facing-front", "serving_plate"],
        ["near-front-edge", "serving_plate"],
        ...
    ]
}
```

Each abstract relation is written as a constraint: the first element is the relation name, followed by the object names bound to it. Ensure all relations are from the library, all object names are provided in the input, and the number of arguments matches the specification.

To assist you to generate the abstract relations, you will be provided with a heirarchical task description and five examples.

Here's the natural language description on how to arrangement the objects for a dining table setup:

<hierarchical natural language task specification>

You are also provided with five example arrangements, each consisting of an instruction and the corresponding functional arrangement in JSON format. These examples will be given one by one. Please begin the arrangement task only after you have received all five examples and the test instruction with the input object list.

<5 instruction-example pairs>

\end{lstlisting}

\paragraph{LLM(Program) + DMs}

The second variant, LLM(Program) + DMs, takes the same task specification and examples as input, and is prompted to generate a Python program that outputs the intermediate symbolic graphs. During inference, given new instructions and object sets, we use another LLM to interpret the generated Python program to produce the grounding graph. This variant generates the program directly from the natural language task specification and five examples, without decomposition or verification. It uses the task specification and examples purely as background knowledge and can thus be understood as naive zero-shot program induction. We detail its prompt below.

\begin{lstlisting}[caption={Prompt for LLM(Program) + DMs}]
Your task is to generate a Python program that proposes a functional arrangement of tableware objects on a dining table. The program accepts:

1. **A natural language instruction** specifying the desired table arrangement.
2. **A list of available tableware objects**, each represented by a unique name.

The output is a **list of abstract relations** among the objects that represent the desired dining table arrangement according to the given instruction.

- Each abstract relation is a tuple of strings:
  - The **first element** is the semantic name of the relation.
  - The **subsequent elements** are the object names associated with it.

The objects are represented as **2D bounding boxes viewed from above**, with all orientations considered from the front.

**Use only the abstract relations from the following library:**

- **("near-front-edge", obj_A):** obj_A is near the table's front edge (bottom in the top-down view).
- **("near-back-edge", obj_A):** obj_A is near the table's back edge (top in the top-down view).
- **("near-left-edge", obj_A):** obj_A is near the table's left edge.
- **("near-right-edge", obj_A):** obj_A is near the table's right edge.
- **("facing-front", obj_A):** obj_A is facing the front (from the table's front edge).
- **("facing-back", obj_A):** obj_A is facing the back (from the table's front edge).
- **("front-half", obj_A):** obj_A is entirely within the front half of the table (horizontally divided along the center).
- **("back-half", obj_A):** obj_A is entirely within the back half of the table.
- **("left-half", obj_A):** obj_A is entirely within the left half of the table (vertically divided along the center).
- **("right-half", obj_A):** obj_A is entirely within the right half of the table.
- **("central-column", obj_A):** The centroid of obj_A is within the central vertical column of the table (middle half of the table's width).
- **("central-row", obj_A):** The centroid of obj_A is within the central horizontal row of the table (middle half of the table's length).
- **("centered-table", obj_A):** The centroid of obj_A coincides with the table's center.
- **("horizontally-aligned-bottom", obj_A, obj_B):** obj_A and obj_B are horizontally aligned at their bases.
- **("horizontally-aligned-centroid", obj_A, obj_B):** obj_A and obj_B are horizontally aligned at their centroids.
- **("vertically-aligned-centroid", obj_A, obj_B):** obj_A and obj_B are vertically aligned at their centroids.
- **("left-of", obj_A, obj_B):** obj_A is immediately to the left of obj_B (from the canonical camera frame).
- **("right-of", obj_A, obj_B):** obj_A is immediately to the right of obj_B.
- **("front-of", obj_A, obj_B):** obj_A is immediately in front of obj_B.
- **("back-of", obj_A, obj_B):** obj_A is immediately behind obj_B.
- **("on-top-of", obj_A, obj_B):** obj_A is placed on top of obj_B.
- **("centered", obj_A, obj_B):** The center of obj_A coincides with the center of obj_B.
- **("vertical-symmetry-on-table", obj_A, obj_B):** obj_A and obj_B are symmetrically placed across the table's vertical axis.
- **("horizontal-symmetry-on-table", obj_A, obj_B):** obj_A and obj_B are symmetrically placed across the table's horizontal axis.
- **("vertical-line-symmetry", axis_obj, obj_A, obj_B):** obj_A and obj_B are symmetrical about the vertical axis of axis_obj.
- **("horizontal-line-symmetry", axis_obj, obj_A, obj_B):** obj_A and obj_B are symmetrical about the horizontal axis of axis_obj.
- **("aligned-in-horizontal-line-bottom", obj_1, ..., obj_N):** Objects are horizontally aligned at their bases with equal spacing.
- **("aligned-in-horizontal-line-centroid", obj_1, ..., obj_N):** Objects are horizontally aligned at their centroids with equal spacing.
- **("aligned-in-vertical-line-centroid", obj_1, ..., obj_N):** Objects are vertically aligned at their centroids with equal spacing.
- **("regular-grid", obj_1, ..., obj_N):** Objects are arranged in a grid or matrix pattern.

The Python program takes in the following input:
1. A natural language instruction indicating the desired arrangement of the objects.
2. A list of objects, each with a unique name.

The program should then output the set of abstract relations (constraints) that satisfy the instruction, formatted as:

```json
{
  "constraints":
    [
        ["facing-front", "serving_plate"],
        ["near-front-edge", "serving_plate"],
        ...
    ]
}
```

Each abstract relation is written as a constraint: the first element is the relation name, followed by the object names bound to it. Ensure all relations are from the library, all object names are provided in the input, and the number of arguments matches the specification.

To assist you to generate the python program, you will be provided with a heirarchical task description and five examples.

Here's the natural language description on how to arrangement the objects for a dining table setup:

<hierarchical natural language task specification>

You are also provided with five example arrangements, each consisting of an instruction and the corresponding functional arrangement in JSON format. These examples will be given one by one. Please only generate the python program for arrangement after you have received all five examples and the test instruction with the input object list.

<5 instruct-example pairs>

\end{lstlisting}
\section{Robot Setup}
\label{ssec:robot}

We conduct our real-robot experiments using two Franka Research 3 (FR3) robotic arms, mounted on opposing sides of a dining table. Each robotic arm is equipped with a Franka Hand parallel gripper. The robots perform motion planning to reach target gripper poses while avoiding collisions. We assume to know the poses from the dining table to the robot bases.

Given the set of objects $\objectSet = \{o_0, \dots, o_{N-1}\}$ and the instruction $desc$, \model outputs the corresponding 2D goal poses $\gP = \{ p_0, \dots, p_{N-1}\}$. We convert each 2D goal pose $p_i= (x_i, y_i, \theta_i)$ to the corresponding 3D goal pose $\goalPose_i \in \mathrm{SE}(3)$.

We assume the initial pose $\initPose_i \in \mathrm{SE}(3)$ of each object $o_i$ is known. In our setup, we put the objects on two calibrated side tables. In general, vision-based methods can be used to estimate object poses.

For each type of object, we define a relative grasp pose $\graspPose_{\tau'}$ between the gripper frame and the object frame. We also specify a sequence of approach poses $\setTfMatrixPick_{\tau'}$ for picking (\eg, moving the gripper above the object, then descending to align with it). To pick up an object $o_i$, the robot executes the sequence of poses $\initPose_i \graspPose_{\tau_i} \tfMatrix$ for each $\tfMatrix \in \setTfMatrixPick_{\tau_i}$, followed by closing the gripper. For placement, we define pre-placement and post-placement pose sequences, $\setTfMatrixPrePlace_{\tau'}$ and $\setTfMatrixPostPlace$, respectively, which are executed before and after opening the gripper.

We sort the arrangement order based on the object types. Once the arrangement order is given, we execute the pick-and-place routines in sequence.

\end{document}